\documentclass[10pt,twocolumn,letterpaper]{article}

\usepackage[pagenumbers]{cvpr}

\usepackage{tabularx}
\usepackage[table]{xcolor}
\usepackage{multirow}
\usepackage{enumitem}
\usepackage{pifont}
\usepackage{bbm}
\usepackage{overpic}
\usepackage{caption}
\usepackage{subcaption}
\usepackage{amsfonts}
\usepackage{booktabs}
\usepackage{arydshln}
\usepackage{nicefrac}
\usepackage{wrapfig}
\usepackage{microtype}
\usepackage{xcolor}
\usepackage[many]{tcolorbox}
\usepackage{floatflt}
\usepackage{textcomp}
\usepackage[export]{adjustbox}
\usepackage{amsmath}
\usepackage{scalerel}
\usepackage{makecell}

\usepackage{algorithm}
\usepackage{algpseudocode}
\usepackage{listings}

\newcommand\doublecheckmark{\textcolor{black}{\checkmark\kern-0.45em\checkmark}}
\newcommand{\inlineColorbox}[2]{\begingroup\setlength{\fboxsep}{1pt}\colorbox{#1}{\hspace*{2pt}\vphantom{Ay}#2\hspace*{2pt}}\endgroup}
\definecolor{DrawioBlue}{RGB}{218,232,252}
\definecolor{DrawioOrange}{RGB}{255,230,204}
\definecolor{DrawioGreen1}{RGB}{204,255,204}
\definecolor{DrawioGreen}{RGB}{213,232,212}
\definecolor{DrawioPurple}{RGB}{225,213,231}
\definecolor{DrawioRed}{RGB}{248,206,204}
\definecolor{DrawioYellow}{RGB}{255, 242, 207}
\definecolor{lightgray}{RGB}{225,225,225}
\definecolor{DrawioDarkPurple}{HTML}{9673A6}
\definecolor{DrawioDarkRed}{HTML}{B85450}
\definecolor{DrawioDarkGreen}{HTML}{009900}
\definecolor{DrawioDarkBlue}{HTML}{6C8EBF}
\definecolor{skyblue}{RGB}{135, 206, 235}
\definecolor{lightcoral}{RGB}{240, 128, 128}
\definecolor{ultraviolet}{HTML}{8365BA}

\newcommand{\ccc}[1]{%
  \ifnum\fpeval{#1 < -8}=1
    \cellcolor{DrawioRed!150}%
  \else\ifnum\fpeval{#1 < -5}=1
    \cellcolor{DrawioRed!100}%
  \else\ifnum\fpeval{#1 < 0}=1
    \cellcolor{DrawioRed!50}%
  \else\ifnum\fpeval{#1 > 20}=1
    \cellcolor{DrawioGreen!200}%
  \else\ifnum\fpeval{#1 > 15}=1
    \cellcolor{DrawioGreen!150}%
  \else\ifnum\fpeval{#1 > 10}=1
    \cellcolor{DrawioGreen!100}%
  \else\ifnum\fpeval{#1 > 5}=1
    \cellcolor{DrawioGreen!100}%
  \else\ifnum\fpeval{#1 > 2.5}=1
    \cellcolor{DrawioGreen!80}%
  \else\ifnum\fpeval{#1 > 0}=1
    \cellcolor{DrawioGreen!50}%
  \else
    \cellcolor{white}%
  \fi\fi\fi\fi\fi\fi\fi\fi\fi
  #1%
}

\lstdefinestyle{Pytorch}{
    language         = Python,
    backgroundcolor  = \color{white},
    basicstyle = \fontsize{8.0pt}{9pt}\selectfont\ttfamily\bfseries,
    columns          = fullflexible,
    breaklines       = true,
    captionpos       = b,
    commentstyle     = \fontsize{4pt}{4pt}\color{codeblue},
    keywordstyle     = \fontsize{4pt}{4pt}\color{codekw},
    morekeywords     = {augment, softmax, confidence\_filter, torch, argmax},
}

\definecolor{codeblue}{rgb}{0.25, 0.5, 0.5}
\definecolor{codekw}{rgb}{0.35, 0.35, 0.75}

\newcommand{\ourslong}{\textit{\textbf{C}IRCLE \textbf{I}teratively \textbf{R}efines \textbf{C}ontextual \textbf{L}earning \textbf{E}xamples}\xspace}
\newcommand{\ours}{CIRCLE\xspace}

\newcommand{\myparagraph}[1]{\vspace{2pt}\noindent \textbf{#1}}

\newcommand{\suppmat}{{\emph{Supp. Mat.}}}

\definecolor{cvprblue}{rgb}{0.21,0.49,0.74}
\usepackage[pagebackref,breaklinks,colorlinks,allcolors=cvprblue]{hyperref}

\title{Large Multimodal Models as General In-Context Classifiers}

\author{Marco Garosi\textsuperscript{1}\quad
Matteo Farina\textsuperscript{1}\quad
Alessandro Conti\textsuperscript{1}\quad
Massimiliano Mancini\textsuperscript{1}\quad
Elisa Ricci\textsuperscript{1,2}\quad \\
{\small \textsuperscript{1}University of Trento\quad\textsuperscript{2}Fondazione Bruno Kessler} \\
{\small \url{https://circle-lmm.github.io}}
}

\begin{document}

\maketitle

\begin{abstract}
Which multimodal model should we use for classification? Previous studies suggest that the answer lies in CLIP-like contrastive Vision-Language Models (VLMs), due to their remarkable performance in zero-shot classification. 
In contrast, Large Multimodal Models (LMM) are more suitable for complex tasks. 
In this work, we argue that this answer overlooks an important capability of LMMs: in-context learning.
We benchmark state-of-the-art LMMs on diverse datasets for closed-world classification and find that, although their zero-shot performance is lower than CLIP's, LMMs with a few in-context examples can match or even surpass contrastive VLMs with cache-based adapters, their ``in-context'' equivalent.
We extend this analysis to the open-world setting, where the generative nature of LMMs makes them more suitable for the task. 
In this challenging scenario, LMMs struggle whenever provided with imperfect context information. 
To address this issue, we propose \ours, a simple training-free method that assigns pseudo-labels to in-context examples, iteratively refining them with the available context itself.
Through extensive experiments, we show that \ours establishes a robust baseline for open-world classification, surpassing VLM counterparts and highlighting the potential of LMMs to serve as unified classifiers, and a flexible alternative to specialized models. 
\end{abstract}
\section{Introduction}
\label{sec:intro}

\begin{figure}[t!]
\centering
\includegraphics[width=\columnwidth]{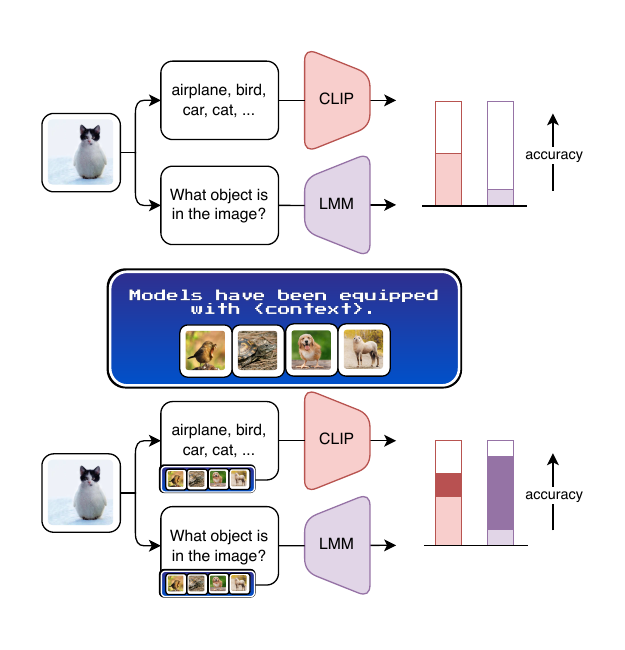}
\caption{
    \textbf{The role of context in  classification.}
    CLIP-like models outperform Large Multimodal Models (LMMs) in closed-world classification.
    However, we show that context dramatically unlocks LMM performance in closed-world classification while allowing them to surpass VLMs on open-world classification.
}
\label{fig:teaser}
\end{figure}

Recent advances in large-scale contrastive vision-language models (VLMs)~\cite{radford2021learning,zhai2023sigmoid} reshaped the landscape of image classification.
In particular, these models can effectively tackle arbitrary classification tasks by measuring the similarity between a list of text labels and the input image in their multimodal representation space.
This ability is closely tied to their strong zero-shot transfer performance~\cite{radford2021learning}, unlocked by their large pre-trained datasets.
While VLMs strive for classification, the same cannot be affirmed for Large Multimodal Models (LMMs)~\cite{li2024llava-ov,bai2023qwen,wang2024qwen2,abdin2024phi,chen2024internvl}.
 
A few studies~\cite{liu2024revisiting,conti2025large,yue2024object,zhang2024visually} have analyzed how well LMMs can recognize objects under both closed- and open-world settings. 
In the closed-world case, models are required to select a label from a predefined set of categories, whereas in open-world~\cite{conti2025large}, they must answer open-ended queries such as \textit{``What is the object in the image?''}.  
These investigations typically compare LMMs to VLMs~\cite{radford2021learning,zhai2023sigmoid}, and consistently find VLMs as highly competitive against their generative counterparts~\cite{conti2025large,zhang2024visually}.
This is surprising, as LMMs are 
usually benchmarked in much more complex scenarios~\cite{fu2023mme,liu2024mmbench,yue2023mmmu,zhang2025lmms}, and should find classification a much easier task. 
Thus, this finding prompts a fundamental question: \emph{are LMMs worse than VLMs at classification, or are they not properly conditioned for the task?}

In this work, we address this question through the lens of \emph{In-Context Learning}~(ICL), investigating whether examples provided at inference time can enhance the performance of LMMs on image classification tasks.
ICL enables models to perform new tasks without parameter updates by conditioning generation on a few input-output examples.
While ICL has been extensively studied in large language models~\cite{brown2020language,chowdhery2022palmscalinglanguagemodeling,touvron2023llamaopenefficientfoundation}, its application to visual tasks has only recently emerged~\cite{zhang2023makes,zhang2024out,qin2024factors,Sun_2024_CVPR}.
These works suggest that visual ICL can unlock latent discriminative and reasoning abilities by allowing models to access additional cues, potentially narrowing the gap with specialized discriminative systems.

We begin with an in-depth analysis of ICL applied to several LMMs in the closed-world classification setting, comparing their behavior with caching-based strategies for VLMs~\cite{zhang2021tip}, where few-shot examples are encoded into a key–value memory.
Our experiments show that LMMs, when conditioned on annotated examples, exhibit substantial performance gains and, in several configurations, close or even surpass the gap with contrastive VLMs.
This finding challenges the conventional assumption that LMMs are inherently weaker at discriminative perception tasks.

Motivated by these results, we extend our investigation to the more challenging open-world classification scenario~\cite{conti2025large}.
This setting introduces two major challenges:
(1) the absence of fixed class labels, preventing balanced per-class sampling,
and (2) the lack of human supervision, requiring automatic labeling of in-context examples.
To address these issues, we propose a novel, annotation-free method that leverages ICL to refine pseudo-labels for unlabeled images iteratively.
Our approach, termed \ourslong~(\ours), employs a \emph{circular iterative procedure} in which pseudo-labels assigned to in-context examples are progressively updated across multiple rounds, allowing the model to self-correct and dynamically infer the level of visual granularity required for the task. 
This mechanism enhances LMM performance in open-world scenarios without requiring external supervision, consistently outperforming VLMs and challenging the common assumption that VLMs are superior to LMMs at discriminative tasks.

\myparagraph{Contributions.}
In summary, our main contributions are:
\begin{itemize}
    \item We provide the first systematic analysis of ICL in LMMs for closed-world image classification, and compare their behavior with caching-based VLMs, showing that LMMs with ICL can match and even surpass VLMs.
    \item We present \ours, a new approach that enhances LMMs for open-world classification using only unlabeled images as ICL examples, iteratively refining their pseudo-labels. 
    \item With an extensive benchmark on open-world classification, we show that, while na\"ive ICL struggles in this setting, \ours largely improves the performance of the base model, consistently surpassing those of VLMs, making a valid case for adopting LMMs for discriminative tasks.
\end{itemize}

\section{Related work}
\label{sec:related-work}

\myparagraph{Vision-Language Models (VLMs) as Classifiers.}
VLMs, such as CLIP~\cite{radford2021learning} or SigLIP~\cite{zhai2023sigmoid}, align image and text representations in a shared, dual-encoder embedding space.
These models, trained on web-crawled datasets (\eg, 400M image-text pairs in~\cite{radford2021learning}), enable zero-shot classification by computing the cosine similarity between an image embedding and the text embeddings of class names.
Despite their impressive zero-shot capabilities, previous studies~\cite{radford2021learning,Zhou_2022_LearningToPrompt,Zhou_2022_ConditionalPrompt} have shown that VLMs exhibit limited generalization to fine-grained classification tasks (\eg, distinguishing between types of flowers or cars) and specialized domains that are underrepresented in the training data (\eg, satellite image classification).
Consequently, significant research has focused on improving VLM performance without updating parameters.
These training-free adaptation strategies typically involve incorporating additional domain-specific knowledge into textual prompts or leveraging a cache of labelled images~\cite{zhang2021tip,udandarao2023sus,garosi2025compositional}.
For example, Tip-Adapter~\cite{zhang2021tip} uses similarities between input image and cached examples for prediction, SuS-X~\cite{udandarao2023sus} automatically constructs its cache through generative models or retrieval, and \textsc{ComCa}~\cite{garosi2025compositional} adapts the cache to the task by analyzing web-scale databases and querying LLMs.
A key architectural constraint of dual-encoder VLMs is their inherent restriction to the closed-world setting, where the prediction space is explicitly defined by a user-provided list of admissible classes~\cite{radford2021learning,zhai2023sigmoid}.
However, recent works~\cite{conti2023vocabulary,liu2024rar,conti2024vocabulary} have explored ways to adapt VLMs for open-world classification.
For instance, CaSED~\cite{conti2023vocabulary} leverages an external vision-language database to allow CLIP to operate effectively in an open-world setting.

\myparagraph{Large Multimodal Models as Classifiers.}
Several studies have focused on assessing the general capabilities of LMMs~\cite{fu2023mme,liu2024mmbench,yue2023mmmu,zhang2025lmms}.
Specifically, for image classification, prior works have examined performance in both closed-world settings~\cite{liu2024revisiting,zhang2024visually} and open-world scenarios~\cite{conti2025large,yue2024object}.
Notably, Zhang \textit{et al.}~\cite{zhang2024visually} evaluated multiple LMMs across several datasets, finding that generative LMMs generally underperform compared to contrastive models. Furthermore, Conti \textit{et al.}~\cite{conti2025large} showed that LMMs often struggle with label granularity, tending to predict generic terms (\eg, ``flower'') rather than precise categories (\eg, ``water lily'').
However, previous studies have not studied the role of in-context examples in this setting.
In this paper, we challenge the assumption that contrastive VLMs outperform LMMs, showing that (i) context is crucial and (ii) LMMs equipped with ICL can achieve superior accuracy under the same conditions.

\myparagraph{In-Context Learning.}
Since the advent of LMMs, one of the biggest challenges is adapting these models to specific tasks due to their large parameter size.
In-Context Learning \cite{brown2020language}, a technique well-known in natural language processing but only recently explored in computer vision \cite{zhang2023makes,zhang2024out,qin2024factors}, has emerged as a strategy to address this issue. 
Previous works \cite{qin2024factors,zhang2024out} on visual ICL have investigated how the selection of in-context examples affects downstream performance in several visual tasks, such as semantic segmentation or detection, revealing that different examples can lead to significantly different results.
Other studies \cite{zhang2023makes} focused on proposing strategies for selecting in-context examples, considering both unsupervised and supervised approaches. 
In this work, we investigate the connection between LMMs with ICL and CLIP-based caching methods, and introduce a novel ICL strategy that leverages unlabeled images as context.
We show that \ours significantly improves classification performance in the challenging open-world setting.

\section{Closed-world classification}
\label{sec:closed-world-classification}

This section provides a comparative analysis of Few-Shot and In-Context Learning for Closed-World Classification, \ie, the standard scenario where the set of classes is known \textit{a priori}, with particular focus on comparing traditional contrastive VLMs with generative LMMs.
We first formalize the classification setting for both model architectures (\cref{sec:cwc:preliminaries}) and then detail their primary mechanisms for Few-Shot Learning: caching for VLMs and In-Context Learning (ICL) for LMMs (\cref{sec:cwc:incontext}).
Finally, we introduce our experimental setup and present a detailed analysis of the results.

\subsection{Preliminaries}
\label{sec:cwc:preliminaries}
In a nutshell, classification aims to assign a semantic label to an input image.
Traditionally, a classifier $\phi$ is a mapping $\phi:\mathcal{V}\rightarrow \mathcal{S}$, \ie, from an image $v \in \mathcal{V}$ to a label $s \in \mathcal{S}$, with $\mathcal{V}$ denoting the image space and $\mathcal{S}$ denoting a set of semantic labels.
In standard classification, $\mathcal{S}=\{s_1,\cdots, s_n\}$ is assumed to be a finite set with \emph{known} elements.
In other words, $\mathcal{S}$ is ``closed'', inspiring the naming of the established Closed-World Classification (CWC) setting.

\myparagraph{CWC with Contrastive VLMs.}
Contrastive Vision-Language Models can be readily employed as zero-shot closed-world classifiers by leveraging their aligned encoders.
We denote the visual encoder with $\phi_{\text{vis}}^{\texttt{VLM}}:\mathcal{V}\rightarrow \mathcal{Z}$ and the textual encoder with $\phi_{\text{text}}^{\texttt{VLM}}:\mathcal{T}\rightarrow \mathcal{Z}$, where, for an hidden size $h$, $\mathcal{Z}$ is the $h{-}1$ unit-hypersphere manifold, and $\mathcal{T}$ is the text space.
When classes in $\mathcal{S}$ are expressed in natural language (\ie, $\mathcal{S} \subset \mathcal{T}$), classification for an image $v \in \mathcal{V}$, is carried out by finding the class whose text embedding has the highest cosine similarity with the embedding of $v$.
Formally, the predicted label $\hat{s}$ is:\footnote{For ease of notation, we omit any template prepend to class names, \eg, ``A photo of a'', ``itap of a'', etc.}
\begin{equation}\label{eq:cwc:clip}
    \hat{s} = \underset{s\in \mathcal{S}}{\text{argmax}}\; \langle
    \phi_{\text{vis}}^{\texttt{VLM}}(x),\phi_{\text{text}}^{\texttt{VLM}}(s)\rangle,
\end{equation}
where $\langle\cdot,\cdot\rangle$ denotes cosine similarity.

\myparagraph{CWC with LMMs.}
Unlike Contrastive VLMs, LMMs typically comprise a vision encoder $\phi_{\text{vis}}^{\texttt{LMM}}$, a connector $\phi_{\text{conn}}^{\texttt{LMM}}$, and an LLM decoder $\phi_{\text{text}}^{\texttt{LMM}}$. 
For an hidden size $h$, an input image $v$ and a textual query $q$, the forward pass takes the form:
\begin{align}
    & \mathbf{V} = \phi_{\text{conn}}^{\texttt{LMM}}(\phi_{\text{vis}}^{\texttt{LMM}}(v)) \in \mathbb{R}^{L_v \times h}; \label{eq:lmm-fw} \\
    & y = \phi_{\text{text}}^{\texttt{LMM}}\bigg( [\mathbf{V}, \mathbf{Q}]\bigg ) \in \mathcal{T}, \label{eq:lmm-fw-2}
\end{align}
where $[\cdot, \cdot]$ denotes first-axis concatenation, $L_v$ is the number of image tokens, and $\mathbf{Q} \in \mathbb{R}^{L_q \times h}$ the tokenized representation of $q$ with length $L_q$.
The output $y \in \mathcal{T}$ is a free-form text as a result of autoregressive generation: this requires to reformulate CWC from textual responses.
An established way is to format $q$ as a Multiple-Choice Question (MCQ) followed by class options in natural language, \eg, ``A: water lily, B: sunflower, C: daffodil''. In the following, we will denote with $q_\mathcal{S}$ the MCQ for the label set $\mathcal{S}$.
For a given image $v$ we obtain a textual output through \cref{eq:lmm-fw-2}. A prediction $\hat{s}$ is then typically parsed via string exact or fuzzy matching. 

\begin{table}
  \caption{\textbf{Closed-world results.} Averaged accuracy over the ten datasets. Higher is better, \textbf{bold} indicates best. See the \suppmat{} for detailed per-dataset results and different context sizes.}
  \centering
  \newcolumntype{C}{>{$}c<{$}}
  \resizebox{\linewidth}{!}{
  \begin{tabular}{cccccccc}
    \toprule

    \multirow{2.5}{*}{\textbf{Model}} & \multirow{2.5}{*}{\textbf{Zero-Shot}} & \multicolumn{6}{c}{\textbf{Few-Shot \& In-Context Learning}} \\

    \cmidrule(lr){3-8}
    & & \text{4-shot} & $\Delta$ & \text{8-shot} & $\Delta$ & \text{16-shot} & $\Delta$\\
    \midrule

    \rowcolor{DrawioYellow}
    \multicolumn{8}{c}{\emph{Contrastive Vision-Language Models (``Few-Shot'' = Tip-Adapter)}} \\ [-2ex] \\
    CLIP ViT-B/32 & 62.6 & 66.0 & \ccc{+3.4} & 68.2 & \ccc{+5.6} & 70.1 & \ccc{+7.5} \\
    k-NN & \text{N/A} & 54.2 & \ccc{-8.4} & 60.3 & \ccc{-2.3} & 64.8 & \ccc{+2.2} \\
    CLIP ViT-B/16 & 65.6 & 69.7 & \ccc{+4.1} & 71.4 & \ccc{+5.8} & 74.4 & \ccc{+8.8} \\
    k-NN & \text{N/A} & 61.6 & \ccc{-4.0} & 66.8 & \ccc{+1.2} & 70.9 & \ccc{+5.3} \\
    CLIP ViT-L/14 & 72.9 & 75.6 & \ccc{+2.7} &77.5 & \ccc{+4.6} & 79.8 & \ccc{+6.9} \\
    k-NN & \text{N/A} & 67.2 & \ccc{-5.7} & 73.1 & \ccc{+0.2} & 76.8 & \ccc{+3.9} \\

    \cmidrule(lr){1-8}
    \rowcolor{DrawioYellow}
    \multicolumn{8}{c}{\emph{Large Multimodal Models (``Few-Shot'' = Vanilla ICL)}} \\ [-2ex] \\
    Qwen-2-VL 7B & 61.3 & 68.1 & \ccc{+6.8} & 73.8 & \ccc{+12.5} & 79.0 & \ccc{+17.7} \\
    Qwen-2.5-VL 7B & 61.2 & 61.3 & \ccc{+0.1} & 70.1 & \ccc{+8.9} & 76.5 & \ccc{+15.3} \\
    LLaVa OneVision 7B & 55.8 & 60.4 & \ccc{+4.6} & 60.8 & \ccc{+5.0} & 60.8 & \ccc{+5.0} \\
    Phi-3.5-Vision & 38.5 & 50.5 & \ccc{+12.0} & 59.1 & \ccc{+20.6} & 67.7 & \ccc{+29.2} \\
    Phi-4-MM & 39.6 & 47.8 & \ccc{+8.2} & 51.6 & \ccc{+12.0} & 53.0 & \ccc{+13.4} \\
    \bottomrule
  \end{tabular}
  }
  \vspace{-5pt}
  \label{tab:cw:icl_vs_models}
\end{table}

\subsection{Few-Shot and In-Context Learning for CWC}
\label{sec:cwc:incontext}
Although Zero-Shot classification is arguably the most widely adopted protocol, both Contrastive VLMs and LMMs benefit from a support set of labeled examples: a \emph{context}.

\myparagraph{\text{For} Contrastive VLMs}, a simple and effective way to exploit an available context is to refine the prediction for $x$ according to its similarity with the context images.
Thanks to its simplicity and influential impact on a variety of follow-up works (\eg, \cite{udandarao2023sus, garosi2025compositional}), we consider the approach introduced by Tip-Adapter \cite{zhang2021tip} as the central Few-Shot Learning method for contrastive VLMs in this work.
In a nutshell, with $\mathcal{C}$ being a uniform $k{\times}|\mathcal{S}|$-sized context comprised of $k{\in}\mathbb{N}$ samples for each semantic category in $\mathcal{S}$, Few-Shot Learning boils down to logit refinement through visual similarity between a query image $v$ and the context images:
\begin{equation}
    \label{eq:cwc:clip-cache}
\underset{s\in \mathcal{S}}{\text{argmax}}\;
\bigg[
\underbrace{\langle\phi_{\text{vis}}^{\texttt{VLM}}(v),\phi_{\text{text}}^{\texttt{VLM}}(s)\rangle}_{\text{Zero-shot score}} +
\underbrace{
    \omega\sum_{x \in C_s} \langle\phi_{\text{vis}}^{\texttt{VLM}}(v), \phi_{\text{vis}}^{\texttt{VLM}}(x)\rangle
}_{\text{In-Context Refinement}} \bigg].
\end{equation}
Here, $\omega$ is a weighting factor and $\mathcal{C}_s \subset \mathcal{C}$ is the subset of $k$ context examples belonging to class $s$.
Importantly, the logit for any $s$ can only be refined if $\mathcal{C}_s$ is \emph{not} empty, which implies that $\mathcal{C}$ must contain samples for \textit{all} pre-defined categories.

\myparagraph{For LMMs,} a natural way to exploit $\mathcal{C}$ is to plug samples within the so-called ``context window'', which enables implicit adaptation through attention.
For $n$ in-context examples, let $\mathbf{X}_i = \phi_{\text{conn}}^{\texttt{LMM}}(\phi_{\text{vis}}^{\texttt{LMM}}(x_i))$ be the encoded image patches of the i-th context image, and $\mathbf{T}_i$ be the tokenized representation of its class name.
Among many possible orderings, we consider a ``Vanilla ICL'' setup of the form:
\begin{align}\label{eq:lmm-icl}
    & y = \phi_{\text{text}}^{\texttt{LMM}}\bigg( [\underbrace{\mathbf{X}_1, \mathbf{T}_1, \ldots, \mathbf{X}_n, \mathbf{T}_n}_{\text{Context } \mathcal{C}}, \underbrace{\mathbf{V}, \mathbf{Q_\mathcal{S}}}_{\text{Img and MCQ}}]\bigg ), 
\end{align}
with $\mathbf{Q}_\mathcal{S}$ being the tokenized representation of the MCQ $q_\mathcal{S}$.
In the next Section, we try to answer the following research question: \emph{Are Contrastive VLMs truly better than LMMs for discriminative tasks?}

\subsection{Are CLIP-like VLMs really better than LMMs?}
\label{sec:cwc:experiments}
Prior work~\cite{conti2025large} conveyed a strong message: despite their strengths in generative tasks, LMMs are largely outperformed by Contrastive VLMs when it comes to discriminative ones. 
In this Section, we examine whether such a claim holds when a shared context is available. 
To this end, we compare Tip-Adapter~\cite{zhang2021tip} as a representative for VLMs, and Vanilla ICL as the nearest equivalent for LMMs.

\myparagraph{Datasets.} We use the established Few-Shot Classification suite~\cite{conti2023vocabulary,conti2025large,shu2022tpt,Zhou_2022_LearningToPrompt}, including Caltech101 \cite{fei2004caltech101} SUN397 \cite{xiao2010sun}, Flowers102 \cite{nilsback2008flowers}, Food101 \cite{bossard2014food}, Oxford Pets \cite{parkhi2012pets}, FGVC Aircraft \cite{maji2013aircraft} Stanford Cars \cite{krause20133cars}, DTD \cite{cimpoi2014dtd}, UCF101 \cite{soomro2012ucf101}, and EuroSAT \cite{helber2019eurosat}.
To ensure an exact comparison with \cite{conti2025large}, we borrow the benchmark categorization: \ding{172} \emph{Prototypical Datasets}, including Caltech101 and SUN397; \ding{173} \emph{Non-Prototypical Datasets}, encompassing DTD, UCF101, and EuroSAT; \ding{174} \emph{Fine-grained Datasets}, with Oxford Pets, Food101, and Flowers102; and \ding{175} \emph{Very Fine-grained Datasets}, with Stanford Cars and FGVC Aircraft.

\begin{figure*}[t!]
\centering
\includegraphics[width=\linewidth]{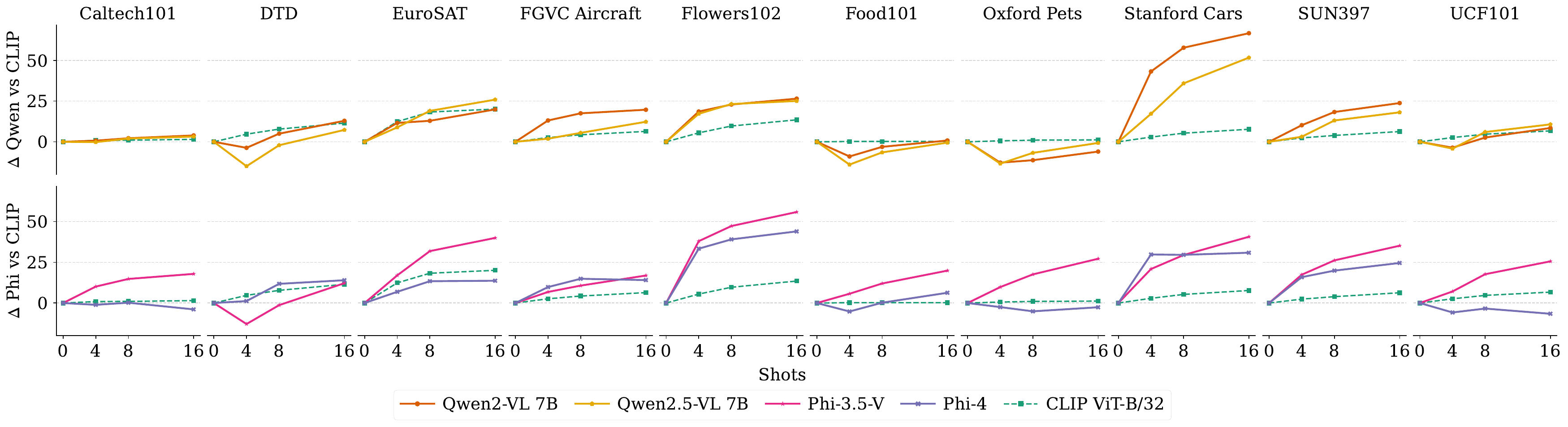}
\vspace{-5mm}
\caption{
    \textbf{Sample efficiency.}
    We visualize the \emph{relative improvement} of a $k$-shot context \wrt the corresponding zero-shot model.
    For contrastive VLMs (dashed lines), we use Tip-Adapter~\cite{zhang2021tip}.
    For LMMs (solid lines), a simple Vanilla ICL setup.
    We report both the Qwen family (top row) and the Phi series (bottom row).
    LMMs benefit much more from additional context than VLMs on most datasets, with peaks of up to $>+50\%$ (\eg, Qwen2-VL-7B on Stanford Cars, Phi3.5-V on Flowers102).
    In contrast, CLIP-ViT-B/32 peaks at $\approx +25\%$. 
}
\label{fig:cw:efficiency}
\end{figure*}

\myparagraph{Models.}
For Contrastive VLMs, we evaluate the three most common CLIP \cite{radford2021learning} variants using ViT-B/32, ViT-B/16, and ViT-L/14 \cite{dosovitskiy2020image}.
For LMMs, we consider a broad spectrum of publicly available models, including the Qwen series (Qwen2-VL \cite{wang2024qwen2}, Qwen2.5-VL \cite{bai2025qwen25vltechnicalreport}), LLaVA OneVision \cite{li2024llava-ov}, and the Phi series (Phi-3.5.Vision \cite{abdin2024phi} and Phi-4-MultiModal~\cite{microsoft2025phi4minitechnicalreportcompact}), ensuring coverage across several design choices for multimodal pretraining, \eg, LLM decoders, pretraining data, vision encoders, and alignment tasks.

\myparagraph{Methods.}
Following established literature, we evaluate at $k \in \{4,8,16\}$ shot availabilities, which implies a $k{\times}|\mathcal{S}|$-sized \emph{shared} context for both VLMs and LMMs.
For the latter, to avoid overly long sequence lengths leading to intractable memory usage, we reduce the effective context size to exactly $k$ (image, label) pairs by drawing from $\mathcal{C}$ in two flavors: \emph{random} (see the \suppmat{}) and \emph{similarity-based} sampling, for which we use the smaller CLIP ViT-B/32 to retrieve the $k$ most similar samples to the given query.
This might naturally raise a critical confounder, \ie, whether any performance gain observed with LMMs \emph{solely} stems from correctly labeled retrieved examples, making ICL collapse to a majority vote within the available context \cite{baldassini2024makes}.
Note that the pipeline $\text{\ding{172} unlabeled query} \rightarrow \text{\ding{173}}~k~\text{retrieved samples} \rightarrow \text{\ding{174} majority vote}$ exactly corresponds to the $k$-NN algorithm.
Hence, to avoid such a confounder, we report $k$-NN results within the available context for all different CLIP models.

\myparagraph{Metrics.} Since $\mathcal{S}$ is closed, we use the \emph{Textual Inclusion} (\textbf{\texttt{TI}}) metric from \cite{conti2025large}, which simply resorts to substring matching.
For CLIP models, \textbf{\texttt{TI}} is equivalent to top-1 accuracy.

\myparagraph{Experimental results,} averaged across all datasets, are given in \cref{tab:cw:icl_vs_models}\footnote{Please refer to the \suppmat{} for the detailed results.} and convey a strong message:
\emph{despite starting from significantly lower zero-shot performance, LMMs can fill the gap with Contrastive VLMs as the context size increases}.
\noindent Specifically, we observe that at lower shot counts (\eg, 4), LMMs often underperform even a simple CLIP k-NN majority vote, but the picture changes dramatically with higher shot counts (\eg, 16). 
For the strongest LMMs, we observe performance boosts relative to zero-shot inference up to $+29.2\%$ for Phi-3.5-Vision and $+17.7\%$ for Qwen2-VL-7B on average.
Importantly, \emph{the strongest LMMs can even match the strongest VLMs when provided with sufficient context}. 
For example, Qwen2-VL-7B matches CLIP-ViT-L/14, the strongest contrastive model, when $k{=}16$ in-context examples are given.
These results show, for the first time, that generative models might serve as a drop-in replacement to VLMs when facing discriminative tasks, and further allows to see contrastive VLMs through different lenses: instead of directly solving a discriminative task, in the future they might serve better as context builders for LMMs.
The most significant advantage relates to \textbf{sample efficiency}, as shown in \cref{fig:cw:efficiency}.
Using only $k{=}16$ shots, LMMs can improve up to $>+50\%$, \wrt their zero-shot behavior,  while CLIP ViT-B/32 (which shows greater relative improvements than the stronger ViT-L/14 variant), peaks at $\approx +25\%$, which entails $2\times$ sample efficiency in terms of relative gains.
Through these results, we speculate that with well-engineered context curation, LMMs might surpass VLMs in future research, even in a closed-world setup that naturally suits the latter.

\section{Open-world classification}
\label{sec:open-world-classification}

In this section, we expand our analysis to Open-World Classification (OWC), where the goal is to classify an image \emph{without} a predefined set of classes. 
In other words, $\mathcal{S}$ is ``open'' and its elements are not known a priori.

\begin{algorithm}[t]
\caption{PyTorch-style code for \ours}
\label{alg:circle-pytorch}
\vspace{-1.ex}
\begin{lstlisting}[style=Pytorch,escapeinside={(@}{@)}]
# image_shots = given unlabeled images (m,3,H,W)
# iters = number of iterations
def classify(model, images):
    context = build_context(model)
    response = model(images, context)
    return response

def build_context(model):
    # step1: pseudo-label images
    labels = forward(model, image_shots)
    context = zip(image_shots, labels)
    # step2: recursive steps
    for iter in iters:
        labels = forward(model, image_shots, context)
        context = zip(image_shots, labels)
    return context

def forward(model, images, context=None):
    responses = []
    for i, image in enumerate(images):
        # context contains the other m-1 images
        cur_context = None
        if context:
            cur_context = context.copy()
            cur_context.remove(i)
        responses[i] = model(image, cur_context)
    return responses
\end{lstlisting}
\vspace{-1.ex}
\end{algorithm}

\subsection{Preliminaries}
\myparagraph{OWC with Contrastive VLMs.}
Thanks to their architectural design, Contrastive VLMs are ultimately image-text retrieval systems.
Therefore, a popular approach to enable OWC is to equip them with a large natural language corpus and consider the most similar caption to the input image as their response.
An extension of this approach is CaSED \cite{conti2023vocabulary}, which extracts candidate class names after retrieving captions from a massive pre-built textual index.

\myparagraph{OWC with LMMs} is a natural scenario, which does not require to reformat a query $q$ into an MCQ. 
Hence, the design of LMMs is naturally suited for this task.

\subsection{Few-shot and In-Context Learning for OWC}
In this setting, we assume access to an \emph{unlabeled} set of $m$ in-context images $\mathcal{C} = \{x_1, \ldots, x_m\}$, and consider using pseudo-labeling to account for the lack of supervision.

\myparagraph{For Contrastive VLMs}, the advantage of having a cache for open-world is dubious, since abundant support data (\ie, the retrieval index) are assumed to be available already. 
Moreover, as stated in \cref{sec:cwc:incontext}, in-context examples for VLMs primarily serve for logit refinement, which is ill-posed in OWC since the notion of ``logit'' is inherently attached to a pre-defined category, which is absent in this scenario.

\myparagraph{For LMMs}, on the other hand, the context can still induce better responses not by steering the predictions towards a specific label set, but by conveying the task to the generative model~\cite{min2022rethinking}, or by narrowing generation down to the visual domain depicted by the context.

\subsection{What can a Context tell about the Open World?}
From \cref{sec:closed-world-classification}, we have empirical evidence that labeled in-context examples significantly benefit CWC.
However, there is no exact notion of ``label'' in OWC, and we assume an unlabeled context comprised of $m$ images only.
An intuitive way to mimic the setting of \cref{sec:closed-world-classification} is to let the LMM generate pseudo labels $\hat{y}_i$ for each of the context images $x_i$, and to use their tokenized representations $\mathbf{\hat{T}}_i$ as a context.
With this strategy, the Vanilla ICL setup of \cref{eq:lmm-icl} now reads:
\begin{align}\label{eq:lmm-pseudo-icl}
    & y = \phi_{\text{text}}^{\texttt{LMM}}\bigg( [\mathbf{X}_1, \mathbf{\hat{T}}_1, \ldots, \mathbf{X}_m, \mathbf{\hat{T}}_m, \mathbf{V}, \mathbf{Q}]\bigg ). 
\end{align}
\noindent We call such a setup \emph{Pseudo In-Context Learning}. 

The key limitation of this na\"ive approach is that it does not capture the inter-sample dependencies within $\mathcal{C}$, which is crucial for disambiguating user intent and converging on a consistent semantic granularity. In particular, we have seen how in-context examples improve classification performance. Thus, we can apply the same principle to improve the pseudo-labels of our in-context examples.
We therefore introduce a new approach, \ourslong (\ours), with a simple key idea: using the context $\mathcal{C}$ as a context for its own examples, recursively.
\ours treats $\mathcal{C}$ itself as a malleable resource, where each pseudo label is refined according to the state of all other in-context examples.
Formally, let $\hat{\mathcal{C}}^t$ denote the state of the context at time $t$, with images $x_i$ (immutable), and pseudo labels $\hat{y}_i^{t}$ evolving over time.
At time $t=0$, the LMM provides independent per-sample pseudo labels $\hat{y}_i^{t=0}$.
From time $t=1$ onward, the context for the $j$-th (in-context) example is the concatenation of all but the $j$-th sample itself, $\{(x_i, \hat{y}_i^{t-1}) : i \neq j ,\,~\forall~ i \in [1, \ldots, m] \}$, 
which has a tokenized representation $\mathbf{C}^t_{i \neq j}$ of the form:
\begin{equation}\label{eq:ineqj-ctx}
    \mathbf{C}^t_{i \neq j} = [\{\mathbf{X}_i, \mathbf{\hat{T}}_i^{t-1} : i\neq j , \forall~ i \in [1, \ldots, m]\}]. \\
\end{equation}
Here, $\mathbf{\hat{T}}_i^{t-1}$ denotes the tokenized representation of the running pseudo label $\hat{y}_i^{t-1}$.
Intuitively, $\mathbf{C}^t_{i \neq j}$ is a ``leave-one-out'' context with all but the $j$-th sample, which we can use to obtain a contextual pseudo label $\hat{y}_j^t$ for sample $x_j$ as follows: 
\begin{equation}\label{eq:ricl-t}
    \hat{y}_j^t = \phi_{\text{text}}^{\texttt{LMM}}\bigg( [\mathbf{C}^t_{i \neq j}, \mathbf{X}_j, \mathbf{Q}]\bigg )
\end{equation}

\noindent This operation is parallelized across the context samples, obtaining new contextual pseudo labels $\{\hat{y}_i^t\}_{i=1}^m$ that capture inter-sample dependencies within the context.
From the hints in \cref{sec:closed-world-classification}, we expect that leveraging auxiliary information makes $\hat{y}_i^t$ more accurate than $\hat{y}_i^{t-1}$.
Repeating this cyclical procedure $T$ times yields a final context $\mathcal{\hat{C}}^T = \{x_i, \hat{y}_i^T\}_{i=1}^m$, which serves as context for the query image.

We report the pseudo-code for \ours in \cref{alg:circle-pytorch}.
We highlight that there are three main steps: \ding{151} pseudo-labeling the context images, \ding{152} recursively refining their labels using the others as context, and \ding{153} classifying an input image.

\subsection{Evaluation protocol}
\label{sec:ow:eval-protocol}
Here, we examine OWC using contrastive VLMs and LMMs, along with their ICL extensions when applicable.
The protocol maintains the outline from the closed-world setting, using the same ten datasets and models.

\myparagraph{Methods.} For contrastive VLMs, we report the open-world baselines CaSED \cite{conti2023vocabulary} and CLIP-Retrieval.
For LMMs, we start by reporting a \textit{Random} baseline for which we assume manually annotated ground truths (which, in practice, can easily be human-annotated given the modest context size).
We then report \textit{Pseudo} ICL and \ours, assuming an unlabeled context with fully automated refinement.
Importantly, all ICL variants use the \emph{same} $m{=}16$ in-context images.
Results at different shot availabilities are in the \suppmat{}

\begin{figure}[t!]
    \makebox[\linewidth][c]{%
        \includegraphics[width=0.95\columnwidth]{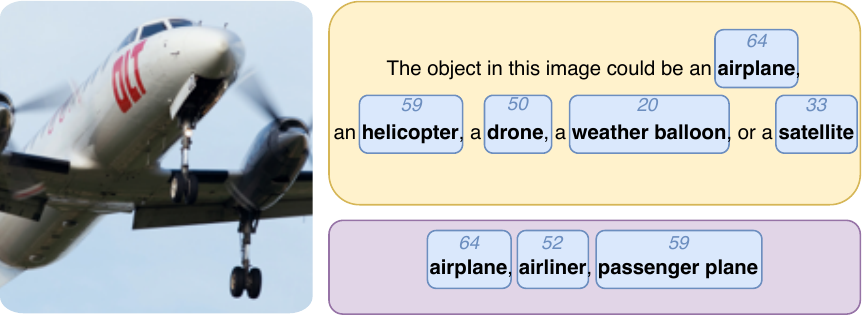}%
    }
    \caption{
        \textbf{Concept similarity issues.}
        \texttt{bCS} can be misleading, for instance by favoring comprehensive yet ungrounded lists of candidates (64 for \inlineColorbox{DrawioYellow}{\textit{Vanilla}} and \inlineColorbox{DrawioPurple}{\ours}).
        \texttt{mCS} instead rewards more coherent and precise answers (50 for \inlineColorbox{DrawioYellow}{\textit{Vanilla}}, 59 for \inlineColorbox{DrawioPurple}{\ours}).
    }
    \label{fig:metric-issue}
    \vspace{-1mm}
\end{figure}

\myparagraph{Metrics.}
Following \cite{conti2025large}, we report performance using four distinct metrics designed to assess the correctness and relevance of generative outputs.
For correctness, we evaluate the presence of the label in the output with \textit{Llama Inclusion} (\textbf{\texttt{LI}}), which uses LLM-as-a-judge to compare the generated outputs with the ground truth\footnote{As \ours generates a list of comma-separated labels, we wrap its output in a fixed template to allow the LLM to process it correctly. More details available in the \suppmat{}}.
For relevance, we estimate the similarity of the generation to the ground truth on a sentence-level (\ie, \textit{Semantic Similarity}, \textbf{\texttt{SS}}), and on a concept-level.
For the latter, we extract all concepts from the free-form output using spaCy and compute the maximum SentenceBERT~\cite{reimers2019sentence} similarity with the ground truth.
We denote this metric as \textit{Best Concept Similarity}, \textbf{${\texttt{bCS}}$}.
Additionally, since \textbf{${\texttt{bCS}}$} might lead to inflated and misleading scores (see \cref{fig:metric-issue}), we further report the \textit{Median Concept Similarity}, \textbf{${\texttt{mCS}}$}.

\begin{table*}
\caption{\textbf{Open-world results.}
  We report results for \textit{Llama Inclusion} (\texttt{LI}), \textit{Semantic Similarity} (\texttt{SS}), \textit{Concept Similarity} ($\texttt{bCS}$), and \textit{Median Concept Similarity} ($\texttt{mCS}$).
  \inlineColorbox{DrawioPurple}{Purple} indicates our \ours.
  Higher is better on all metrics.
  For each LMM, \textbf{bold} indicates the best result.
  See the \suppmat{} for an extended version of the table.
  }
  \centering
  \resizebox{\linewidth}{!}{
  \begin{tabular}{lr c cccc c cccc c cccc c cccc}
    \toprule
    \multirow[c]{2.5}{*}{Model} & \multirow[c]{2.5}{*}{Method} && \multicolumn{4}{c}{Prototypical} && \multicolumn{4}{c}{Non-prototypical} && \multicolumn{4}{c}{Fine-grained} && \multicolumn{4}{c}{Very fine-grained} \\
    \cmidrule{4-7} \cmidrule{9-12} \cmidrule{14-17} \cmidrule{19-22}
    &&&
        \texttt{LI} & \texttt{SS} & \texttt{bCS} & \texttt{mCS}
    &&
        \texttt{LI} & \texttt{SS} & \texttt{bCS} & \texttt{mCS}
    &&
        \texttt{LI} & \texttt{SS} & \texttt{bCS} & \texttt{mCS}
    &&
        \texttt{LI} & \texttt{SS} & \texttt{bCS} & \texttt{mCS} \\
    \midrule

    \rowcolor{DrawioYellow}
    \multicolumn{22}{c}{\emph{Contrastive Vision-Language Models}} \\ [-2ex] \\
    
    CaSED \cite{conti2023vocabulary} (ViT-L/14)
        & &&
        46.3 & 58.9 & 59.8 & 58.8 &&
        18.6 & 41.8 & 42.4 & 41.8 &&
        46.6 & 60.7 & 61.6 & 60.7 &&
        47.1 & 38.5 & 38.5 & 38.5
        \\

    CLIP retrieval (ViT-L/14)
        & &&
        42.9 & 40.2 & 60.6 & 31.1 &&
        28.1 & 43.4 & 32.4 & 23.8 &&
        45.4 & 42.9 & 65.4 & 31.8 &&
        18.1 & 39.7 & 56.1 & 29.5
        \\

    \cmidrule(lr){1-22}
    \rowcolor{DrawioYellow}
    \multicolumn{22}{c}{\emph{Large Multimodal Models}} \\ [-2ex] \\

    &  \textit{Zero-Shot} &&
        78.7 & 51.9 & 76.0 & 43.7 &&
        42.6 & 30.8 & 49.8 & 29.2 &&
        64.0 & 39.2 & 62.9 & 31.9 &&
        63.0 & 34.5 & 43.4 & 33.1
        \\
    & \emph{Random Ctx} &&
        24.4 & 41.4 & 52.7 & 39.7 &&
        17.1 & 23.4 & 41.3 & 23.0 &&
        31.7 & 34.4 & 44.8 & 34.6 &&
        31.1 & 29.2 & 34.1 & 27.9
        \\
    & \emph{Pseudo ICL} &&
        81.1 & 53.4 & \textbf{76.2} & 44.4 &&
        42.8 & 31.2 & \textbf{50.1} & 26.9 &&
        53.1 & 40.2 & 64.4 & 30.7 &&
        49.1 & 38.9 & 49.1 & 38.6
        \\
    \rowcolor{DrawioPurple}
    \cellcolor{white} \multirow[c]{-4}{*}{\raggedleft Qwen2-VL 7B~\cite{bai2023qwen}}
    & \ours &&
        \textbf{91.5} & \textbf{65.6} & 74.3 & \textbf{63.5} &&
        \textbf{61.6} & \textbf{41.9} & 49.4 & \textbf{40.5} &&
        \textbf{87.3} & \textbf{61.1} & \textbf{72.0} & \textbf{57.3} &&
        \textbf{91.5} & \textbf{42.5} & \textbf{50.2} & \textbf{39.8}
        \\
    
    \cmidrule{1-22}
    &  \textit{Zero-Shot} &&
        82.9 & 47.9 & \textbf{79.9} & 31.1 &&
        45.9 & 30.5 & 54.0 & 24.8 &&
        73.8 & 47.0 & \textbf{78.9} & 29.5 &&
        69.0 & 45.8 & 68.6 & 27.1
        \\
    & \emph{Random Ctx} &&
        82.5 & 53.5 & 78.2 & 43.1 &&
        48.2 & 35.7 & \textbf{58.9} & 27.3 &&
        70.6 & 49.0  & 76.7 & 36.3 &&
        34.7 & \textbf{52.4} & \textbf{70.2} & \textbf{42.0}
        \\
    & \emph{Pseudo ICL} &&
        80.6 & 49.3 & 78.6 & 31.0 &&
        42.7 & 31.6 & 53.2 & 24.2 &&
        63.8 & 45.1 & 74.8 & 28.4 &&
        39.7 & 46.0 & 63.4 & 25.5
        \\
    \rowcolor{DrawioPurple}
    \cellcolor{white} 
    \multirow[c]{-4}{*}{\raggedleft Qwen2.5-VL 7B~\cite{bai2025qwen25vltechnicalreport}}
    & \ours &&
        \textbf{94.9} & \textbf{67.7} & 68.1 & \textbf{67.2} &&
        \textbf{67.6} & \textbf{42.6} & 45.1 & \textbf{42.3} &&
        \textbf{86.3} & \textbf{60.1} & 60.9 & \textbf{59.7} &&
        \textbf{93.6} & 36.4 & 36.6 & 36.5
        \\
    
    \cmidrule{1-22}
    &  \textit{Zero-Shot} &&
        53.2 & 56.2 & 62.0 & 53.4 &&
        28.1 & 31.6 & 43.8 & 30.2 &&
        40.4 & 39.0 & 43.9 & 37.2 &&
        \textbf{76.7} & 31.8 & 32.3 & 30.9
        \\
    & \emph{Random Ctx} &&
        14.0 & 29.3 & 36.7 & 29.6 &&
         8.6 & 26.0 & 39.3 & 24.1 &&
        21.0 & 33.2 & 35.8 & 33.4 &&
        75.8 & 30.8 & 30.8 & 30.8
        \\
    & \emph{Pseudo ICL} &&
        19.7 & 31.2 & 41.1 & 30.5 &&
         3.3 & 18.3 & 31.5 & 19.7 &&
        20.3 & 35.5 & 40.0 & 34.7 &&
        70.4 & 30.6 & 31.1 & 29.8
        \\
    \rowcolor{DrawioPurple}
    \cellcolor{white} 
    \multirow[c]{-4}{*}{\raggedleft LLaVa OV 7B~\cite{li2024llava-ov}}
    & \ours &&
        \textbf{72.2} & \textbf{74.0} & \textbf{74.0} & \textbf{74.0} &&
        \textbf{61.7} & \textbf{55.3} & \textbf{55.3} & \textbf{55.3} &&
        \textbf{55.1} & \textbf{46.0} & \textbf{46.0} & \textbf{46.0} &&
        74.2 & \textbf{32.9} & \textbf{32.8} & \textbf{32.9}
        \\
    
    \cmidrule{1-22}
    &  \textit{Zero-Shot} &&
        60.7 & 48.2 & \textbf{65.6} & 46.1 &&
        28.7 & 24.9 & \textbf{36.7} & 24.1 &&
        50.7 & 32.1 & 47.2 & 31.3 &&
        54.2 & 29.5 & 36.3 & 29.8
        \\
    & \emph{Random Ctx} &&
        41.1 & 48.3 & 54.4 & 48.4 &&
        10.7 & 26.3 & 36.4 & 27.4 &&
        25.2 & 26.3 & 36.4 & 27.4 &&
        59.2 & 27.1 & 31.7 & 26.3
        \\
    & \emph{Pseudo ICL} &&
        54.1 & 44.1 & 61.7 & 40.1 &&
        23.7 & 22.8 & 35.1 & 21.5 &&
        43.1 & 33.0 & \textbf{48.6} & 29.7 &&
        24.9 & 32.4 & \textbf{41.8} & 32.5
        \\
    \rowcolor{DrawioPurple}
    \cellcolor{white} 
    \multirow[c]{-4}{*}{\raggedleft Phi-3.5-V~\cite{abdin2024phi}}
    & \ours &&
        \textbf{92.1} & \textbf{59.7} & 63.2 & \textbf{60.3} &&
        \textbf{58.3} & \textbf{30.0} & 35.2 & \textbf{31.7} &&
        \textbf{88.1} & \textbf{39.2} & 45.7 & \textbf{42.1} &&
        \textbf{99.6} & \textbf{33.0} & 33.6 & \textbf{33.1}
        \\
    
    \cmidrule{1-22}
    &  \textit{Zero-Shot} &&
        49.8 & 57.4 & 58.7 & 57.2 &&
        21.2 & 29.2 & 32.7 & 29.2 &&
        37.7 & 39.2 & 39.2 & 39.1 &&
        73.6 & 31.6 & 31.7 & 31.6
        \\
    & \emph{Random Ctx} &&
        11.6 & 32.8 & 32.8 & 32.8 &&
         5.9 & 28.5 & 28.6 & 28.5 &&
        31.3 & 37.9 & 37.9 & 37.9 &&
        74.6 & 30.8 & 30.7 & 30.6
        \\
    & \emph{Pseudo ICL} &&
        51.9 & 61.5 & 62.0 & 61.5 &&
        15.1 & 26.6 & 31.2 & 26.7 &&
        37.7 & 41.6 & 41.7 & 41.5 &&
        72.8 & 31.9 & 31.9 & 31.9
        \\
    \rowcolor{DrawioPurple}
    \cellcolor{white} 
    \multirow[c]{-4}{*}{\raggedleft Phi-4-MM~\cite{microsoft2025phi4minitechnicalreportcompact}}
    & \ours &&
        \textbf{91.5} & \textbf{65.5} & \textbf{70.1} & \textbf{66.4} &&
        \textbf{67.6} & \textbf{43.2} & \textbf{46.1} & \textbf{43.4} &&
        \textbf{79.1} & \textbf{53.3} & \textbf{55.5} & \textbf{53.0} &&
        \textbf{75.2} & \textbf{40.2} & \textbf{42.5} & \textbf{37.9}
        \\
    
    \bottomrule
  \end{tabular}}
  \label{tab:ow:main}
  \vspace{-4mm}
\end{table*}

\begin{figure}[t!]
    \makebox[\linewidth][c]{%
\includegraphics[trim={0.7cm 0.5cm 0.7cm 0.6cm},clip,width=0.95\linewidth]{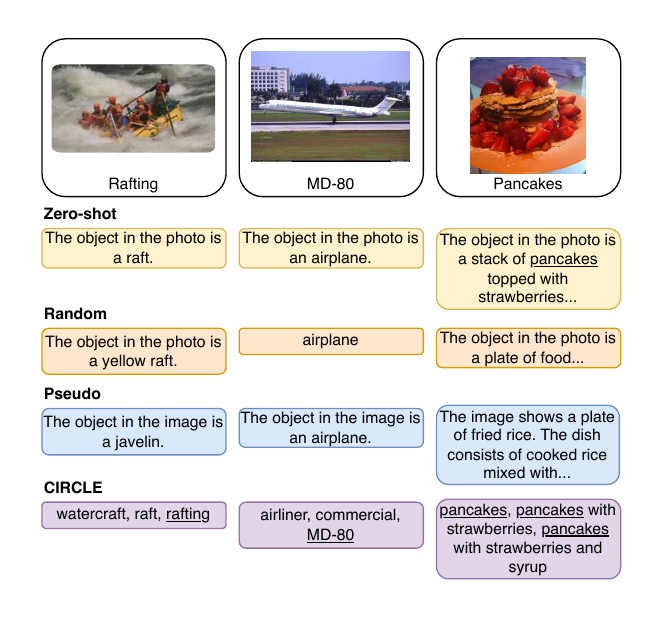}%
    }
    \vspace{-5mm}
    \caption{
        \textbf{Qualitative results.} We visualize some qualitative results on examples from UCF101, FGVC Aircraft, and Food101. \underline{Underline} indicates a correct prediction of the ground truth.
    }
    \label{fig:qualitative}
    \vspace{-4mm}
\end{figure}

\myparagraph{Experimental results.}
We present OWC results in \cref{tab:ow:main} and visualize some qualitative examples of the predictions of these methods in \cref{fig:qualitative}.
There are three clear observations from the table. The first is the low performance of LMMs zero-shot w.r.t. VLMs models designed for OWC on semantic similarity metrics (\eg, +15 \texttt{mCS} and +7 \texttt{SS} for CaSED in \textit{Prototypical} vs. Qwen2-VL), they lag behind in \texttt{LI} (\eg, -32 for the same models), as shown also in \cite{conti2025large}. The second is that na\"ive ICL may degrade the performance of the zero-shot baseline across both correctness (\eg, -20 \texttt{LI} for LLaVa OV Random on \textit{Non-prototypical}) and relevance metrics (\eg, -27 for \texttt{SS} for the same model). Additionally, pseudo-labeling alone does not mitigate this issue, showing similar performance degradation (\eg, -33.5 \texttt{LI} and -35 \texttt{SS} for the same settings). \Cref{fig:qualitative} shows some examples where the baselines fail to steer the model to the correct generation.

The third observation is that \ours inverts these trends, providing results consistently higher than any ICL counterpart and VLM, across both correctness and relevance metrics and for all dataset categories. 
For instance, on the \textit{Prototypical} tasks, \ours improves the \texttt{LI} score of Qwen2-VL to 91.5, significantly outperforming the second-best baseline, \textit{Pseudo ICL} (81.1).
Similarly, it improves the performance of \textit{Zero-Shot} Phi-3.5-V (60.7) by +31.4\% on \textit{Prototypical} tasks.
This trend holds across all models and task groups.

Crucially, this notable increase in \texttt{LI} is not achieved at the expense of semantic quality, a common trade-off for baselines.
For example, \textit{Zero-Shot} Qwen2.5-VL achieves high \texttt{LI} (82.9) but comparatively low \texttt{SS} (47.9) and $\texttt{mCS}$ (31.1), suggesting verbose or semantically imprecise outputs.
In contrast, \ours shows strong performance on \emph{both} inclusion and semantic relevance metrics simultaneously.

Through these results, we can draw two key conclusions:
(i) ICL does not guarantee improvements of OWC performance of LMMs;
(ii) \ours, a simple yet carefully designed strategy that exploits ICL itself to refine the context, overcomes these limitations, and consistently outperforms both the base model and VLM counterparts. Thus, with the right strategy, LMMs are more suitable than VLMs for open-world classification.
We refer to the \suppmat{} for a comprehensive, metric-by-metric analysis of the results.

\begin{figure*}[!t]
\includegraphics[width=\linewidth, valign=t]{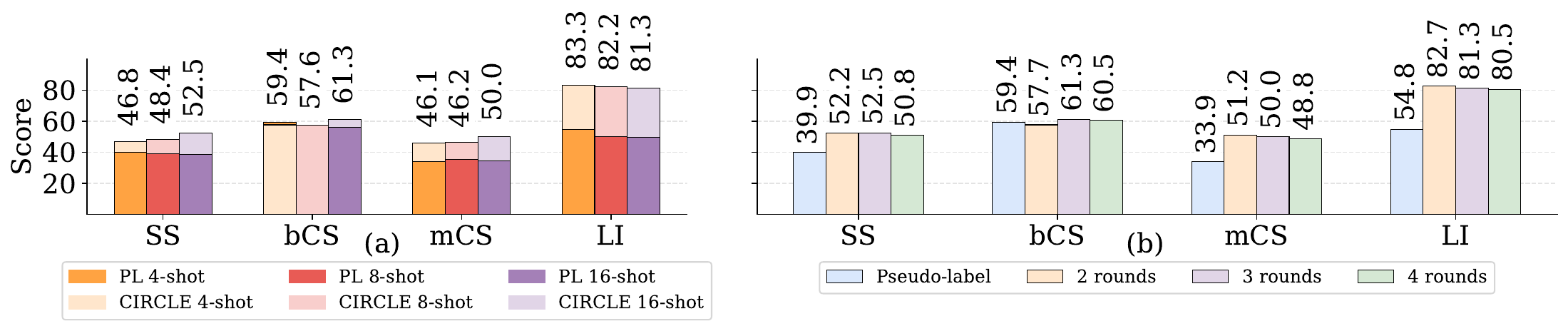}
\vspace{-2mm}
    \captionof{figure}{
    \inlineColorbox{DrawioPurple}{Purple} denotes our default configuration, applied to Qwen2-VL 7B~\cite{wang2024qwen2}.
    We report \textit{semantic similarity} (\texttt{SS}), \textit{Concept Similarity} (\texttt{bCS}), \textit{Median Concept Similarity} (\texttt{mCS}), and \textit{Llama Inclusion} (\texttt{LI}).
    \textbf{(a) Ablation on context size.}
    Results are reported for \textit{Pseudo-labeling} (\texttt{PL}) and \ours, with 4, 8, and 16 shots.
    \textbf{(b) Ablation on \ours rounds.}
    We report results comparing \textit{Pseudo-labeling} against increasing numbers of \ours refinement rounds.
    See the \suppmat{} for an extended version of these results.
    }
    \vspace{-4mm}
    \label{fig:ow:ablations}
\end{figure*}

\subsection{Ablation study}
In this section, we analyze \ours's components, studying the impact of context size and test-time compute available.

\myparagraph{Context size.}
In \cref{fig:ow:ablations}(a), we analyze the effect of adding more samples to the context.
We observe that adding examples has a positive effect on semantic-related metrics (\texttt{SS}, \texttt{bCS}, \texttt{mCS}), while \texttt{LI} remains stable.
As the LMM has to find the correct level of information to solve the task, providing more examples enhances the model's overall perspective.

\myparagraph{Test-time compute.}
We measure the impact of the amount of test-time compute in \cref{fig:ow:ablations}(b), showing how performance varies as the number of recursive rounds for \ours increases.
While allowing the model to spend more compute resources (\ie, iterating more times) has a strongly beneficial effect over the \textit{pseudo-labeling} setting (no refinement rounds), we find that it leads to diminishing returns.

\subsection{Streaming ICL}
\label{sec:streaming}
\begin{figure}[!t]
\includegraphics[width=\columnwidth, valign=t]{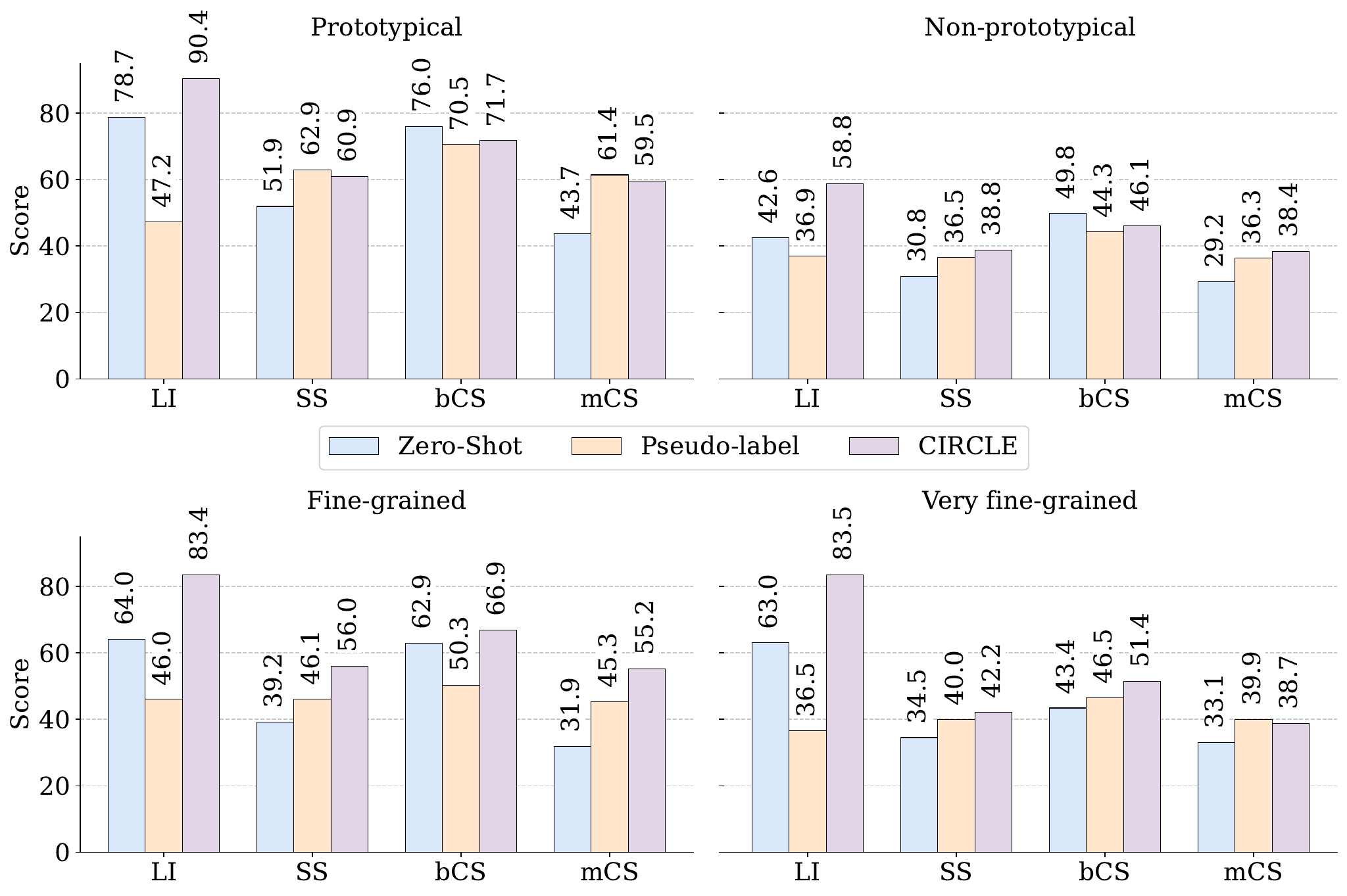}
\vspace{-2mm}
    \captionof{figure}{
    \inlineColorbox{DrawioPurple}{Purple} indicates our \ours, applied to Qwen2-VL 7B~\cite{wang2024qwen2}.
    We report \textit{semantic similarity} (\texttt{SS}), \textit{Concept Similarity} (\texttt{bCS}), \textit{Median Concept Similarity} (\texttt{mCS}), and \textit{Llama Inclusion} (\texttt{LI}).
    See the \suppmat{} for an extended version of these results.
    }
    \label{fig:ow:streaming}
    \vspace{-5mm}
\end{figure}

\ours's simple design enables flexible adaptation to multiple scenarios.
We show this concept in the context of online ICL, taking inspiration from online cache building in VLM classification~\cite{karmanov2024efficient}. 
We extend our method to work online, \ie, on a stream of samples. 
To avoid designing ad-hoc filtering protocols, we explore a simple strategy: selecting  $m$ random unlabeled examples from the history of test data.
From there, na\"ive pseudo-labeling or \ours can be employed to assign labels to the in-context samples.
We report the results for this experiment in \cref{fig:ow:streaming} for Qwen2-VL, comparing zero-shot predictions with pseudo-labeling and \ours.
Even in this setting, \ours consistently improves over the Pseudo-labeling, with the latter sometimes even decreasing the performance of the base model (\eg, -18 \texttt{LI} on fine-grained).
In particular, it consistently achieves the best \texttt{LI} (\eg, +16 on non-prototypical), and it is either best or second-best in all settings and for all relevance metrics.
These results confirm the robustness of our approach even in a difficult streaming environment.
A comprehensive analysis for all models and tasks is in the \suppmat{}

\section{Conclusions}
\label{sec:conclusions}

In this paper, we demonstrated that LMMs show strong data efficiency in closed-world image classification through ICL. Still, their predictions are fragile and highly sensitive to context noise. These issues become more severe in open-world settings, where standard ICL fails to form consistent semantic interpretations. To address this, we introduced \ourslong, a training-free self-refinement mechanism that models dependencies across unlabeled in-context examples, enabling the LMM to construct a coherent, task-aligned structure. Our experiments show that this refinement is essential: \ours consistently stabilizes LMM outputs and yields high-precision predictions that outperform all baselines in the OW setting, demonstrating that targeted refinement is necessary to unlock LMMs’ discriminative capabilities.

\noindent\textbf{Limitations.}
Although \ours requires no human annotations, this lack of supervision may make the refinement converge to semantically coherent but task-misaligned label interpretations.
In addition, the streaming variant's dynamic memory updates can introduce non-trivial computational overhead when processing large or continuous data streams.

\noindent\textbf{Future work.}
Two promising directions exist:
(i) incorporating lightweight training or parameter-efficient tuning to stabilize refinement and potentially recover the task structure with ambiguous unlabeled data;
(ii) improving the streaming mechanism's efficiency (\eg, via memory compression, selective updating, scalable retrieval strategies), enabling broader deployment in resource-constrained scenarios.

\noindent\textbf{Acknowledgments.}
We acknowledge ISCRA for awarding this project access to the LEONARDO supercomputer, owned by the EuroHPC Joint Undertaking, hosted by CINECA (Italy).
This work was supported by the EU Horizon ELIAS (No. 101120237), ELLIOT (No. 101214398), and TURING (No. 101215032) projects.

{
    \small
    \bibliographystyle{ieeenat_fullname}
    \bibliography{main}

@String(IJCV = {Int. J. Comput. Vis.})

@String(CVPR= {IEEE Conf. Comput. Vis. Pattern Recog.})

@String(ICCV= {Int. Conf. Comput. Vis.})

@String(ECCV= {Eur. Conf. Comput. Vis.})

@String(CVPRW= {IEEE Conf. Comput. Vis. Pattern Recog. Worksh.})

@String(IJCV  = {IJCV})

@String(CVPR  = {CVPR})

@String(ICCV  = {ICCV})

@String(ECCV  = {ECCV})

@String(CVPRW= {CVPRW})

@inproceedings{radford2021learning,
  title={Learning transferable visual models from natural language supervision},
  author={Radford, Alec and Kim, Jong Wook and Hallacy, Chris and Ramesh, Aditya and Goh, Gabriel and Agarwal, Sandhini and Sastry, Girish and Askell, Amanda and Mishkin, Pamela and Clark, Jack and others},
  booktitle={ICML},
  year={2021},
}

@inproceedings{zhai2023sigmoid,
  title={Sigmoid loss for language image pre-training},
  author={Zhai, Xiaohua and Mustafa, Basil and Kolesnikov, Alexander and Beyer, Lucas},
  booktitle={ICCV},
  year={2023}
}

@article{Zhou_2022_LearningToPrompt,
  author       = {Zhou, Kaiyang and Yang, Jingkang and Loy, Chen Change and Liu, Ziwei},
  title        = {Learning to Prompt for Vision‑Language Models},
  journal      = {IJCV},
  year         = {2022},
}

@inproceedings{Zhou_2022_ConditionalPrompt,
  author       = {Zhou, Kaiyang and Yang, Jingkang and Loy, Chen Change and Liu, Ziwei},
  title        = {Conditional Prompt Learning for Vision‑Language Models},
  booktitle    = {CVPR},
  year         = {2022},
}

@article{conti2023vocabulary,
  title={Vocabulary-free image classification},
  author={Conti, Alessandro and Fini, Enrico and Mancini, Massimiliano and Rota, Paolo and Wang, Yiming and Ricci, Elisa},
  journal={NeurIPS},
  year={2023}
}

@article{fu2023mme,
  title        = {MME: A Comprehensive Evaluation Benchmark for Multimodal Large Language Models},
  author       = {Fu, Chaoyou and Chen, Peixian and Shen, Yunhang and Qin, Yulei and Zhang, Mengdan and Lin, Xu and Yang, Jinrui and Zheng, Xiawu and Li, Ke and Sun, Xing and Wu, Yunsheng and Ji, Rongrong},
  journal      = {arXiv},
  year         = {2023}
}

@inproceedings{liu2024mmbench,
  title        = {MMBench: Is Your Multi‑modal Model an All‑around Player?},
  author       = {Liu, Yuan and Duan, Haodong and Zhang, Yuanhan and Li, Bo and Zhang, Songyang and Zhao, Wangbo and Yuan, Yike and Wang, Jiaqi and He, Conghui and Liu, Ziwei and Chen, Kai and Lin, Dahua},
  booktitle    = {ECCV},
  year         = {2024},
}

@inproceedings{yue2023mmmu,
  title        = {MMMU: A Massive Multi‑discipline Multimodal Understanding and Reasoning Benchmark for Expert AGI},
  author       = {Xiang Yue and Yuansheng Ni and Kai Zhang and Tianyu Zheng and Ruoqi Liu and Ge Zhang and Samuel Stevens and Dongfu Jiang and Weiming Ren and Yuxuan Sun and Cong Wei and Botao Yu and Ruibin Yuan and Renliang Sun and Ming Yin and Boyuan Zheng and Zhenzhu Yang and Yibo Liu and Wenhao Huang and Huan Sun and Yu Su and Wenhu Chen},
  booktitle    = {CVPR},
  year         = {2024}
}

@article{liu2024revisiting,
  title={Revisiting mllms: An in-depth analysis of image classification abilities},
  author={Liu, Huan and Xiao, Lingyu and Liu, Jiangjiang and Li, Xiaofan and Feng, Ze and Yang, Sen and Wang, Jingdong},
  journal={arXiv},
  year={2024}
}

@inproceedings{zhang2025lmms,
  title={Lmms-eval: Reality check on the evaluation of large multimodal models},
  author={Zhang, Kaichen and Li, Bo and Zhang, Peiyuan and Pu, Fanyi and Cahyono, Joshua Adrian and Hu, Kairui and Liu, Shuai and Zhang, Yuanhan and Yang, Jingkang and Li, Chunyuan and others},
  booktitle={NAACL},
  year={2025}
}

@article{conti2025large,
  title={On large multimodal models as open-world image classifiers},
  author={Conti, Alessandro and Mancini, Massimiliano and Fini, Enrico and Wang, Yiming and Rota, Paolo and Ricci, Elisa},
  journal={ICCV},
  year={2025}
}

@inproceedings{yue2024object,
  title={Object recognition as next token prediction},
  author={Yue, Kaiyu and Chen, Bor-Chun and Geiping, Jonas and Li, Hengduo and Goldstein, Tom and Lim, Ser-Nam},
  booktitle={CVPR},
  year={2024}
}

@article{zhang2024visually,
  title={Why are visually-grounded language models bad at image classification?},
  author={Zhang, Yuhui and Unell, Alyssa and Wang, Xiaohan and Ghosh, Dhruba and Su, Yuchang and Schmidt, Ludwig and Yeung-Levy, Serena},
  journal={NeurIPS},
  year={2024}
}

@article{brown2020language,
  title={Language Models are Few-Shot Learners},
  author={Brown, Tom B. and Mann, Benjamin and Ryder, Nick and Subbiah, Melanie and Kaplan, Jared D. and Dhariwal, Prafulla and Neelakantan, Arvind and Shyam, Pranav and Sastry, Girish and Askell, Amanda and others},
  journal={arXiv},
  year={2020}
}

@article{zhang2023makes,
  title={What makes good examples for visual in-context learning?},
  author={Zhang, Yuanhan and Zhou, Kaiyang and Liu, Ziwei},
  journal={NeurIPS},
  year={2023}
}

@article{qin2024factors,
  title={What factors affect multi-modal in-context learning? an in-depth exploration},
  author={Qin, Libo and Chen, Qiguang and Fei, Hao and Chen, Zhi and Li, Min and Che, Wanxiang},
  journal={NeurIPS},
  year={2024}
}

@article{zhang2024out,
  title={On the out-of-distribution generalization of multimodal large language models},
  author={Zhang, Xingxuan and Li, Jiansheng and Chu, Wenjing and Hai, Junjia and Xu, Renzhe and Yang, Yuqing and Guan, Shikai and Xu, Jiazheng and Cui, Peng},
  journal={arXiv},
  year={2024}
}

@article{zhang2021tip,
  title={Tip-adapter: Training-free clip-adapter for better vision-language modeling},
  author={Zhang, Renrui and Fang, Rongyao and Zhang, Wei and Gao, Peng and Li, Kunchang and Dai, Jifeng and Qiao, Yu and Li, Hongsheng},
  journal={arXiv},
  year={2021}
}

@article{liu2024rar,
  title={Rar: Retrieving and ranking augmented mllms for visual recognition},
  author={Liu, Ziyu and Sun, Zeyi and Zang, Yuhang and Li, Wei and Zhang, Pan and Dong, Xiaoyi and Xiong, Yuanjun and Lin, Dahua and Wang, Jiaqi},
  journal={arXiv},
  year={2024}
}

@inproceedings{udandarao2023sus,
  title={Sus-x: Training-free name-only transfer of vision-language models},
  author={Udandarao, Vishaal and Gupta, Ankush and Albanie, Samuel},
  booktitle={ICCV},
  year={2023}
}

@misc{touvron2023llamaopenefficientfoundation,
      title={LLaMA: Open and Efficient Foundation Language Models}, 
      author={Hugo Touvron and Thibaut Lavril and Gautier Izacard and Xavier Martinet and Marie-Anne Lachaux and Timothée Lacroix and Baptiste Rozière and Naman Goyal and Eric Hambro and Faisal Azhar and Aurelien Rodriguez and Armand Joulin and Edouard Grave and Guillaume Lample},
      year={2023},
      journal={arXiv},
}

@misc{chowdhery2022palmscalinglanguagemodeling,
      title={PaLM: Scaling Language Modeling with Pathways}, 
      author={Aakanksha Chowdhery and Sharan Narang and Jacob Devlin and Maarten Bosma and Gaurav Mishra and Adam Roberts and Paul Barham and Hyung Won Chung and Charles Sutton and Sebastian Gehrmann and Parker Schuh and Kensen Shi and Sasha Tsvyashchenko and Joshua Maynez and Abhishek Rao and Parker Barnes and Yi Tay and Noam Shazeer and Vinodkumar Prabhakaran and Emily Reif and Nan Du and Ben Hutchinson and Reiner Pope and James Bradbury and Jacob Austin and Michael Isard and Guy Gur-Ari and Pengcheng Yin and Toju Duke and Anselm Levskaya and Sanjay Ghemawat and Sunipa Dev and Henryk Michalewski and Xavier Garcia and Vedant Misra and Kevin Robinson and Liam Fedus and Denny Zhou and Daphne Ippolito and David Luan and Hyeontaek Lim and Barret Zoph and Alexander Spiridonov and Ryan Sepassi and David Dohan and Shivani Agrawal and Mark Omernick and Andrew M. Dai and Thanumalayan Sankaranarayana Pillai and Marie Pellat and Aitor Lewkowycz and Erica Moreira and Rewon Child and Oleksandr Polozov and Katherine Lee and Zongwei Zhou and Xuezhi Wang and Brennan Saeta and Mark Diaz and Orhan Firat and Michele Catasta and Jason Wei and Kathy Meier-Hellstern and Douglas Eck and Jeff Dean and Slav Petrov and Noah Fiedel},
      year={2022},
      journal={arXiv},
}

@InProceedings{Sun_2024_CVPR,
    author    = {Sun, Quan and Cui, Yufeng and Zhang, Xiaosong and Zhang, Fan and Yu, Qiying and Wang, Yueze and Rao, Yongming and Liu, Jingjing and Huang, Tiejun and Wang, Xinlong},
    title     = {Generative Multimodal Models are In-Context Learners},
    booktitle = {CVPR},
    year      = {2024},
}

@inproceedings{garosi2025compositional,
  title={Compositional Caching for Training-free Open-vocabulary Attribute Detection},
  author={Garosi, Marco and Conti, Alessandro and Liu, Gaowen and Ricci, Elisa and Mancini, Massimiliano},
  booktitle={CVPR},
  year={2025}
}

@article{dosovitskiy2020image,
  title={An image is worth 16x16 words: Transformers for image recognition at scale},
  author={Dosovitskiy, Alexey},
  journal={arXiv},
  year={2020}
}

@article{bai2023qwen,
  title={Qwen-vl: A frontier large vision-language model with versatile abilities},
  author={Bai, Jinze and Bai, Shuai and Yang, Shusheng and Wang, Shijie and Tan, Sinan and Wang, Peng and Lin, Junyang and Zhou, Chang and Zhou, Jingren},
  journal={arXiv},
  year={2023}
}

@article{abdin2024phi,
  title={Phi-3 technical report: A highly capable language model locally on your phone},
  author={Abdin, Marah and Aneja, Jyoti and Awadalla, Hany and Awadallah, Ahmed and Awan, Ammar Ahmad and Bach, Nguyen and Bahree, Amit and Bakhtiari, Arash and Bao, Jianmin and Behl, Harkirat and others},
  journal={arXiv},
  year={2024}
}

@article{wang2024qwen2,
  title={Qwen2-vl: Enhancing vision-language model's perception of the world at any resolution},
  author={Wang, Peng and Bai, Shuai and Tan, Sinan and Wang, Shijie and Fan, Zhihao and Bai, Jinze and Chen, Keqin and Liu, Xuejing and Wang, Jialin and Ge, Wenbin and others},
  journal={arXiv},
  year={2024}
}

@article{li2024llava-ov,
  title={Llava-onevision: Easy visual task transfer},
  author={Li, Bo and Zhang, Yuanhan and Guo, Dong and Zhang, Renrui and Li, Feng and Zhang, Hao and Zhang, Kaichen and Zhang, Peiyuan and Li, Yanwei and Liu, Ziwei and others},
  journal={arXiv},
  year={2024}
}

@inproceedings{chen2024internvl,
  title={Internvl: Scaling up vision foundation models and aligning for generic visual-linguistic tasks},
  author={Chen, Zhe and Wu, Jiannan and Wang, Wenhai and Su, Weijie and Chen, Guo and Xing, Sen and Zhong, Muyan and Zhang, Qinglong and Zhu, Xizhou and Lu, Lewei and others},
  booktitle={CVPR},
  year={2024}
}

@misc{bai2025qwen25vltechnicalreport,
      title={Qwen2.5-VL Technical Report}, 
      author={Shuai Bai and Keqin Chen and Xuejing Liu and Jialin Wang and Wenbin Ge and Sibo Song and Kai Dang and Peng Wang and Shijie Wang and Jun Tang and Humen Zhong and Yuanzhi Zhu and Mingkun Yang and Zhaohai Li and Jianqiang Wan and Pengfei Wang and Wei Ding and Zheren Fu and Yiheng Xu and Jiabo Ye and Xi Zhang and Tianbao Xie and Zesen Cheng and Hang Zhang and Zhibo Yang and Haiyang Xu and Junyang Lin},
      year={2025},
      journal={arXiv},
}

@misc{microsoft2025phi4minitechnicalreportcompact,
      title={Phi-4-Mini Technical Report: Compact yet Powerful Multimodal Language Models via Mixture-of-LoRAs}, 
      author={Microsoft and : and Abdelrahman Abouelenin and Atabak Ashfaq and Adam Atkinson and Hany Awadalla and Nguyen Bach and Jianmin Bao and Alon Benhaim and Martin Cai and Vishrav Chaudhary and Congcong Chen and Dong Chen and Dongdong Chen and Junkun Chen and Weizhu Chen and Yen-Chun Chen and Yi-ling Chen and Qi Dai and Xiyang Dai and Ruchao Fan and Mei Gao and Min Gao and Amit Garg and Abhishek Goswami and Junheng Hao and Amr Hendy and Yuxuan Hu and Xin Jin and Mahmoud Khademi and Dongwoo Kim and Young Jin Kim and Gina Lee and Jinyu Li and Yunsheng Li and Chen Liang and Xihui Lin and Zeqi Lin and Mengchen Liu and Yang Liu and Gilsinia Lopez and Chong Luo and Piyush Madan and Vadim Mazalov and Arindam Mitra and Ali Mousavi and Anh Nguyen and Jing Pan and Daniel Perez-Becker and Jacob Platin and Thomas Portet and Kai Qiu and Bo Ren and Liliang Ren and Sambuddha Roy and Ning Shang and Yelong Shen and Saksham Singhal and Subhojit Som and Xia Song and Tetyana Sych and Praneetha Vaddamanu and Shuohang Wang and Yiming Wang and Zhenghao Wang and Haibin Wu and Haoran Xu and Weijian Xu and Yifan Yang and Ziyi Yang and Donghan Yu and Ishmam Zabir and Jianwen Zhang and Li Lyna Zhang and Yunan Zhang and Xiren Zhou},
      year={2025},
      journal={arXiv},
}

@article{min2022rethinking,
  title={Rethinking the role of demonstrations: What makes in-context learning work?},
  author={Min, Sewon and Lyu, Xinxi and Holtzman, Ari and Artetxe, Mikel and Lewis, Mike and Hajishirzi, Hannaneh and Zettlemoyer, Luke},
  journal={arXiv},
  year={2022}
}

@inproceedings{reimers2019sentence,
  title={Sentence-BERT: Sentence Embeddings using Siamese BERT-Networks},
  author={Reimers, Nils and Gurevych, Iryna},
  booktitle={EMNLP-IJCNLP},
  year={2019}
}

@article{shu2022tpt,
  title={Test-time prompt tuning for zero-shot generalization in vision-language models},
  author={Shu, Manli and Nie, Weili and Huang, De-An and Yu, Zhiding and Goldstein, Tom and Anandkumar, Anima and Xiao, Chaowei},
  journal={NeurIPS},
  year={2022}
}

@inproceedings{fei2004caltech101,
  title={Learning generative visual models from few training examples: An incremental bayesian approach tested on 101 object categories},
  author={Fei-Fei, Li and Fergus, Rob and Perona, Pietro},
  booktitle=cvprw,
  year={2004},
}

@inproceedings{bossard2014food,
  title={Food-101--mining discriminative components with random forests},
  author={Bossard, Lukas and Guillaumin, Matthieu and Van Gool, Luc},
  booktitle=eccv,
  year={2014}
}

@inproceedings{nilsback2008flowers,
  title={Automated flower classification over a large number of classes},
  author={Nilsback, Maria-Elena and Zisserman, Andrew},
  booktitle={Indian conference on computer vision, graphics \& image processing},
  year={2008},
}

@inproceedings{parkhi2012pets,
  title={Cats and dogs},
  author={Parkhi, Omkar M and Vedaldi, Andrea and Zisserman, Andrew and Jawahar, CV},
  booktitle=cvpr,
  year={2012}
}

@inproceedings{xiao2010sun,
  title={Sun database: Large-scale scene recognition from abbey to zoo},
  author={Xiao, Jianxiong and Hays, James and Ehinger, Krista A and Oliva, Aude and Torralba, Antonio},
  booktitle=cvpr,
  year={2010}
}

@inproceedings{cimpoi2014dtd,
  title={Describing textures in the wild},
  author={Cimpoi, Mircea and Maji, Subhransu and Kokkinos, Iasonas and Mohamed, Sammy and Vedaldi, Andrea},
  booktitle=cvpr,
  year={2014}
}

@inproceedings{krause20133cars,
  title={3d object representations for fine-grained categorization},
  author={Krause, Jonathan and Stark, Michael and Deng, Jia and Fei-Fei, Li},
  booktitle={ICCV-WS},
  year={2013}
}

@article{maji2013aircraft,
  title={Fine-grained visual classification of aircraft},
  author={Maji, Subhransu and Rahtu, Esa and Kannala, Juho and Blaschko, Matthew and Vedaldi, Andrea},
  journal={arXiv},
  year={2013}
}

@article{soomro2012ucf101,
  title={UCF101: A dataset of 101 human actions classes from videos in the wild},
  author={Soomro, Khurram and Zamir, Amir Roshan and Shah, Mubarak},
  journal={arXiv},
  year={2012}
}

@article{helber2019eurosat,
  title={Eurosat: A novel dataset and deep learning benchmark for land use and land cover classification},
  author={Helber, Patrick and Bischke, Benjamin and Dengel, Andreas and Borth, Damian},
  journal={IEEE Journal of Selected Topics in Applied Earth Observations and Remote Sensing},
  year={2019},
}

@article{conti2024vocabulary,
  title={Vocabulary-free image classification and semantic segmentation},
  author={Conti, Alessandro and Fini, Enrico and Mancini, Massimiliano and Rota, Paolo and Wang, Yiming and Ricci, Elisa},
  journal={arXiv},
  year={2024}
}

@inproceedings{baldassini2024makes,
  title={What makes multimodal in-context learning work?},
  author={Baldassini, Folco Bertini and Shukor, Mustafa and Cord, Matthieu and Soulier, Laure and Piwowarski, Benjamin},
  booktitle={CVPR-WS},
  year={2024}
}

@inproceedings{karmanov2024efficient,
  title={Efficient test-time adaptation of vision-language models},
  author={Karmanov, Adilbek and Guan, Dayan and Lu, Shijian and El Saddik, Abdulmotaleb and Xing, Eric},
  booktitle={CVPR},
  year={2024}
}
}

\appendix
\clearpage
\setcounter{page}{1}
\maketitlesupplementary

This supplementary material provides additional technical and implementation details, alongside extended experimental results and further analyses that complement the main paper.
Specifically, the material is organized as follows:
\begin{itemize}
    \item \cref{supp:sec:datasets} provides details regarding the datasets.

    \item \cref{supp:sec:cw} extends the \emph{closed-world} analysis from \cref{tab:cw:icl_vs_models} (\textit{Main}) by reporting granular, per-category performance breakdowns.

    \item \cref{supp:sec:ow} presents the full set of results for the \emph{open-world} setting (\cref{tab:ow:main} in \textit{Main}), including evaluations on 4-shot and 8-shot configurations.

    \item \cref{supp:sec:ow_streaming} provides the complete data for the \emph{streaming} experiments visualized in \cref{fig:ow:streaming} (\textit{Main}).

    \item \cref{supp:sec:impl-details} outlines the implementation specifics.

    \item \cref{supp:sec:qualitatives} showcases an extended gallery of qualitative examples, complementing the visualizations in \cref{fig:qualitative} (\textit{Main}).
\end{itemize}

\section{Datasets}
\label{supp:sec:datasets}

We summarize the evaluation datasets in \cref{supp:tab:dataset_details}.
The experiments utilized the same training and test splits as previous work~\cite{conti2025large,conti2023vocabulary}.

\section{Closed-world results}
\label{supp:sec:cw}

This section presents the detailed, per-dataset results of the experiments described in \cref{sec:cwc:experiments,tab:cw:icl_vs_models} (\textit{Main}).

As they did not fit on a single page, we split the table into two parts, showing contrastive VLMs in \cref{supp:tab:cw:Similarity_vs_models_vlms} and generative LMMs in \cref{supp:tab:cw:Similarity_vs_models_lmms}.

For contrastive VLMs (\cref{supp:tab:cw:Similarity_vs_models_vlms}), we report the performance of the \textit{Zero-Shot} model, Tip-Adapter~\cite{zhang2021tip} using the same backbone, and k-NN (as described in \cref{sec:cwc:experiments} in \textit{Main}).
For both Tip-Adapter and k-NN, we report results with 4, 8, and 16 shots.
For each experiment, we report the average (\textit{Avg.}) on the ten datasets, and highlight the delta $\Delta$ \wrt the average of the corresponding \textit{Zero-Shot} model.

\begin{table}[tb]
  \caption{\textbf{Dataset details.}
  Summary details of the datasets used in our experiments.}
  \centering
  \newcolumntype{C}{>{$}c<{$}}
  \resizebox{\linewidth}{!}{
  \begin{tabular}{cl cc}
    \toprule

    \textbf{Abbr.} & \textbf{Dataset} & \textbf{Images} & \textbf{Classes} \\

    \midrule

    C101 & Caltech101~\cite{fei2004caltech101} & 2,465 & 100 \\
    DTD & DTD~\cite{cimpoi2014dtd} & 1,692 & 47 \\
    ESAT & Eurosat~\cite{helber2019eurosat} & 8,100 & 10 \\
    FGVC & FGVC Aircraft~\cite{maji2013aircraft} & 3,333 & 100 \\
    FLWR & Flowers102~\cite{nilsback2008flowers} & 2,463 & 102 \\
    FOOD & Food101~\cite{bossard2014food} & 30,300 & 101 \\
    PETS & Oxford Pets~\cite{parkhi2012pets} & 3,669 & 37 \\
    CARS & Stanford Cars~\cite{krause20133cars} & 8,041 & 196 \\
    S397 & SUN397~\cite{xiao2010sun} & 19,850 & 397 \\
    U101 & UCF101~\cite{soomro2012ucf101} & 3,783 & 101 \\
    
    \bottomrule
  \end{tabular}
  }
  \label{supp:tab:dataset_details}
\end{table}

For LMMs (\cref{supp:tab:cw:Similarity_vs_models_lmms}), we report the performance of the \textit{Vanilla} model (\ie, multi-choice) and when using both a \textit{Random} context and a \textit{Similarity}-based context that retrieves images.
\textit{Similarity} ICL uses CLIP ViT-B/32~\cite{radford2021learning} to retrieve the most similar images from the few-shot pool.
We report results in the 4-, 8-, and 16-shot settings.

The results for contrastive VLMs in \cref{supp:tab:cw:Similarity_vs_models_vlms} highlight the efficacy of adapter-based methods over simple nearest-neighbor retrieval in the few-shot regime.
Tip-Adapter consistently outperforms the \textit{Zero-Shot} baseline across all backbones and shot settings, achieving a peak improvement of +8.8\% with ViT-B/16 at 16 shots. Conversely, the k-NN approach often struggles in low-shot scenarios (4 shots), frequently yielding negative deltas compared to the zero-shot baseline (\eg, -8.4\% for ViT-B/32).
However, k-NN performance recovers as the number of support examples increases to 16.
Overall, while scaling the backbone from ViT-B/32 to ViT-L/14 improves absolute accuracy metrics, the relative trends between Tip-Adapter and k-NN remain consistent.

In the case of generative LMMs (\cref{supp:tab:cw:Similarity_vs_models_lmms}), the data reveals a critical sensitivity to the quality of the in-context examples.
The \textit{Random} setting proves catastrophic across the board, causing massive performance degradation (\eg, up to -48.7\% for Qwen-2-VL), suggesting that irrelevant context acts as noise that confuses the model rather than aiding it.
In contrast, \textit{Similarity}-based retrieval unlocks significant performance gains.
This is most notable in Phi-3.5-Vision, which scores +29.2\% at 16 shots, and the Qwen family, where Qwen-2-VL achieves a +17.7\% boost.

\begin{table*}
\caption{
\textbf{Closed-world results.}
Accuracy on the ten datasets.
\textbf{Bold} indicates the best result for each VLM.
$\Delta$ computed \wrt the Avg. of \textit{Zero-Shot}.
}
\begin{subtable}{\textwidth}
  \caption{\textbf{Vision-Language Models (VLMs).}}
  \centering
  \newcolumntype{C}{>{$}c<{$}}
  \resizebox{\linewidth}{!}{
  \begin{tabular}{cc cccccccccccc}
    \toprule

    \multirow{1}{*}{\textbf{Model}} & \multirow{1}{*}{\textbf{Shots}} &
    C101 & DTD & ESAT & FGVC & FLWR & FOOD & PETS & CARS & S397 & U101 & Avg. & $\Delta$ \\
    
    \midrule

    \rowcolor{DrawioYellow}
    \multicolumn{14}{c}{\emph{CLIP ViT-B/32}} \\ [-2ex] \\
    \textit{Zero-Shot} & - & 92.4 & 44.6 & 45.1 & 19.2 & 69.3 & 80.5 & 87.3 & 60.1 & 62.5 & 64.5 & 62.6 \\
    TIP-Adapter & 4 & 93.3 & 49.3 & 57.5 & 21.8 & 74.8 & \textbf{80.7} & 87.9 & 63.0 & 64.9 & 67.1 & 66.0 & \ccc{+3.4} \\
    k-NN & 4 & 87.6 & 43.5 & 66.7 & 20.5 & 80.0 & 52.8 & 41.9 & 37.1 & 50.9 & 60.8 & 54.2 & \ccc{-8.4} \\

    TIP-Adapter & 8 & 93.4 & 52.4 & 63.4 & 23.5 & 79.0 & \textbf{80.7} & 88.3 & 65.4 & 66.4 & 69.2 & 68.2 & \ccc{+5.6} \\
    k-NN & 8 & 87.2 & 53.0 & 72.5 & 25.2 & 83.6 & 60.9 & 52.4 & 45.7 & 58.6 & 64.3 & 60.3 & \ccc{-2.3} \\

    TIP-Adapter & 16 & \textbf{93.9} & 56.1 & 65.2 & 25.6 & 82.8 & \textbf{80.7} & \textbf{88.5} & \textbf{67.8} & \textbf{68.8} & \textbf{71.2} & \textbf{70.1} & \ccc{+7.5} \\
    k-NN & 16 & 90.8 & \textbf{56.3} & \textbf{73.4} & \textbf{29.0} & \textbf{85.8} & 67.0 & 61.8 & 54.5 & 62.2 & 67.7 & 64.8 & \ccc{+2.2} \\

    \cmidrule(lr){1-14}
    \rowcolor{DrawioYellow}
    \multicolumn{14}{c}{\emph{CLIP ViT-B/16}} \\ [-2ex] \\
    \textit{Zero-Shot} & - & 94.2 & 45.7 & 48.2 & 24.8 & 71.3 & 85.8 & 89.1 & 65.6 & 63.0 & 68.5 & 65.6 \\
    TIP-Adapter & 4 & 94.8 & 49.1 & 66.6 & 27.9 & 76.8 & 86.2 & 90.1 & 68.3 & 65.8 & 71.5 & 69.7 & \ccc{+4.2} \\
    k-NN & 4 & 90.2 & 46.6 & 72.7 & 29.5 & 86.9 & 65.6 & 58.0 & 50.2 & 53.6 & 62.6 & 61.6 & \ccc{-4.0} \\

    TIP-Adapter & 8 & 94.9 & 52.2 & 69.6 & 30.5 & 79.9 & 86.2 & 90.2 & 70.0 & 67.6 & 73.2 & 71.4 & \ccc{+5.8} \\
    k-NN & 8 & 90.6 & 56.3 & 73.7 & 32.0 & 90.3 & 71.8 & 65.2 & 59.3 & 60.6 & 68.3 & 66.8 & \ccc{+1.2} \\

    TIP-Adapter & 16 & \textbf{95.6} & 57.0 & \textbf{76.4} & 33.3 & 84.8 & \textbf{86.5} & \textbf{91.9} & \textbf{73.1} & \textbf{70.0} & \textbf{75.5} & \textbf{74.4} & \ccc{+8.8} \\
    k-NN & 16 & 93.1 & \textbf{58.2} & 76.2 & \textbf{36.3} & \textbf{92.0} & 77.0 & 75.3 & 65.7 & 64.4 & 70.6 & 70.9 & \ccc{+5.3} \\

    \cmidrule(lr){1-14}
    \rowcolor{DrawioYellow}
    \multicolumn{14}{c}{\emph{CLIP ViT-L/14}} \\ [-2ex] \\
    \textit{Zero-Shot} & - & 96.8 & 53.7 & 60.2 & 32.7 & 80.8 & 91.0 & 93.5 & 76.8 & 68.0 & 75.8 & 72.9 \\
    TIP-Adapter & 4 & 97.1 & 57.3 & 68.8 & 37.1 & 84.9 & 91.2 & 93.8 & 79.0 & 70.1 & 77.0 & 75.6 & \ccc{+2.7} \\
    k-NN & 4 & 92.9 & 49.9 & 75.0 & 35.0 & 93.7 & 75.2 & 61.9 & 58.6 & 56.6 & 72.7 & 67.2 & \ccc{-5.7} \\

    TIP-Adapter & 8 & 97.3 & 59.9 & 74.5 & 40.4 & 87.9 & 91.3 & 94.0 & 80.2 & 71.4 & 78.1 & 77.5 & \ccc{+4.6} \\
    k-NN & 8 & 94.1 & 59.6 & 81.0 & 40.9 & 94.4 & 81.4 & 70.9 & 67.6 & 63.7 & 77.4 & 73.1 & \ccc{+0.2} \\

    TIP-Adapter & 16 & \textbf{97.5} & \textbf{64.4} & 77.7 & \textbf{44.3} & 92.2 & \textbf{91.4} & \textbf{94.2} & \textbf{82.8} & \textbf{73.6} & \textbf{80.1} & \textbf{79.8} & \ccc{+6.9} \\
    k-NN & 16 & 96.2 & 64.1 & \textbf{82.3} & 44.2 & \textbf{96.5} & 83.9 & 81.6 & 72.8 & 67.7 & 78.4 & 76.8 & \ccc{+3.9} \\
    
    \bottomrule
  \end{tabular}
  }
  \label{supp:tab:cw:Similarity_vs_models_vlms}
\end{subtable}
\end{table*}

\begin{table*}
\ContinuedFloat
\caption{
\textbf{Closed-world results.}
Accuracy on the ten datasets.
\textbf{Bold} indicates the best result for each LMM.
$\Delta$ computed \wrt the average (Avg.) of \textit{Vanilla}.
}
\begin{subtable}{\textwidth}
  \caption{\textbf{Large Multimodal Models (LMMs).}}
  \centering
  \newcolumntype{C}{>{$}c<{$}}
  \resizebox{\linewidth}{!}{
  \begin{tabular}{cc cccccccccccc}
    \toprule

    \multirow{1}{*}{\textbf{Model}} & \multirow{1}{*}{\textbf{Shots}} &
    C101 & DTD & ESAT & FGVC & FLWR & FOOD & PETS & CARS & S397 & U101 & Avg. & $\Delta$ \\
    
    \midrule
    
    \rowcolor{DrawioYellow}
    \multicolumn{14}{c}{\emph{Qwen-2-VL 7B}} \\ [-2ex] \\
    \textit{Vanilla} & - & 91.6 & 61.9 & 33.9 & 50.3 & 69.8 & 82.3 & \textbf{86.8} & 16.4 & 48.9 & 70.9 & 61.3 \\

    \cmidrule{2-14}
    \multirow{3}{*}{\textit{Random}} & 4 & 71.5 & 41.7 & 21.0 & 11.1 & 20.0 & 64.1 & 55.3 & 7.0 & 10.8 & 37.4 & 34.0 & \ccc{-27.3} \\
    & 8 & 49.5 & 23.6 & 19.7 & 4.4 & 8.6 & 42.5 & 30.0 & 4.6 & 4.7 & 15.4 & 20.3 & \ccc{-41.0} \\
    & 16 & 20.6 & 16.9 & 17.5 & 2.3 & 3.9 & 39.0 & 15.5 & 2.7 & 1.2 & 5.9 & 12.6 & \ccc{-48.7} \\

    \cmidrule{2-14}
    \multirow{3}{*}{\textit{Similarity}} & 4 & 92.3 & 58.2 & 45.5 & 63.4 & 88.4 & 73.3 & 74.0 & 59.6 & 59.1 & 67.3 & 68.1 & \ccc{+6.8} \\
    & 8 & 93.8 & 66.9 & 46.8 & 67.8 & 92.7 & 79.2 & 75.5 & 74.2 & 67.2 & 73.5 & 73.8 & \ccc{+12.5} \\
    & 16 & \textbf{95.5} & \textbf{74.8} & \textbf{53.8} & \textbf{70.0} & \textbf{96.3} & \textbf{83.1} & 80.8 & \textbf{83.1} & \textbf{72.7} & \textbf{79.4} & \textbf{79.0} & \ccc{+17.7} \\

    \cmidrule(lr){1-14}
    \rowcolor{DrawioYellow}
    \multicolumn{14}{c}{\emph{Qwen-2.5-VL 7B}} \\ [-2ex] \\
    \textit{Vanilla} & - & 92.4 & 65.2 & 29.7 & 48.5 & 73.2 & \textbf{77.4} & \textbf{85.9} & 17.3 & 55.9 & 66.1 & 61.2 \\
    
    \cmidrule{2-14}
    \multirow{3}{*}{\textit{Random}} & 4 & 90.1 & 52.1 & 30.4 & 13.2 & 45.6 & 57.6 & 75.0 & 10.9 & 27.0 & 58.9 & 46.1 & \ccc{-15.1} \\
    & 8 & 88.6 & 51.9 & 30.8 & 12.1 & 48.3 & 53.8 & 72.2 & 11.9 & 28.2 & 58.3 & 45.6 & \ccc{-15.6} \\
    & 16 & 87.2 & 50.7 & 30.3 & 11.0 & 48.7 & 47.3 & 67.8 & 10.9 & 28.1 & 55.4 & 43.7 & \ccc{-17.5} \\

    \cmidrule{2-14}
    \multirow{3}{*}{\textit{Similarity}} & 4 & 92.2 & 50.2 & 38.6 & 50.4 & 90.4 & 63.4 & 72.5 & 34.5 & 59.0 & 61.9 & 61.3 & \ccc{+0.1} \\
    & 8 & 94.4 & 63.1 & 48.7 & 54.0 & 96.4 & 70.9 & 79.1 & 53.2 & 69.0 & 72.1 & 70.1 & \ccc{+8.9} \\
    & 16 & \textbf{95.6} & \textbf{72.5} & \textbf{55.6} & \textbf{60.8} & \textbf{98.3} & {76.9} & 85.2 & \textbf{69.0} & \textbf{74.0} & \textbf{76.8} & \textbf{76.5} & \ccc{+15.3} \\

    \cmidrule(lr){1-14}
    \rowcolor{DrawioYellow}
    \multicolumn{14}{c}{\emph{LLaVa OneVision 7B}} \\ [-2ex] \\
    \textit{Vanilla} & - & 91.5 & \textbf{73.8} & 48.9 & \textbf{51.6} & 38.5 & \textbf{69.7} & \textbf{53.3} & 14.8 & 46.6 & \textbf{69.5} & 55.8 \\

    \cmidrule{2-14}
    \multirow{3}{*}{\textit{Random}} & 4 & 16.9 & 15.2 & 17.6 & 3.6 & 4.8 & 14.5 & 21.4 & 1.3 & 10.0 & 11.9 & 11.7 & \ccc{-44.1} \\
    & 8 & 20.3 & 18.0 & 22.5 & 4.7 & 5.8 & 14.7 & 19.3 & 1.3 & 10.8 & 15.7 & 13.3 & \ccc{-42.5} \\
    & 16 & 80.2 & 69.6 & 1.5 & 11.4 & 24.6 & 11.9 & 39.1 & 3.5 & 11.2 & 23.3 & 27.6 & \ccc{-28.2} \\

    \cmidrule{2-14}
    \multirow{3}{*}{\textit{Similarity}} & 4 & \textbf{94.8} & 68.3 & \textbf{60.0} & 30.0 & 73.3 & 60.5 & 47.7 & 55.6 & 47.9 & 66.4 & 60.4 & \ccc{+4.6} \\
    & 8 & 94.1 & 71.0 & 51.9 & 30.6 & 73.1 & 63.7 & 47.2 & \textbf{55.7} & 51.2 & 69.3 & \textbf{60.8} & \ccc{+5.0} \\
    & 16 & 94.6 & 72.8 & 53.8 & 29.7 & \textbf{73.4} & 64.3 & 46.0 & 52.4 & \textbf{51.8} & 69.2 & \textbf{60.8} & \ccc{+5.0} \\

    \cmidrule(lr){1-14}
    \rowcolor{DrawioYellow}
    \multicolumn{14}{c}{\emph{Phi-3.5-Vision}} \\ [-2ex] \\
    \textit{Vanilla} & - & 76.4 & 52.4 & 33.7 & 6.8 & 27.3 & 53.1 & 49.4 & 6.5 & 31.8 & 47.6 & 38.5 \\
    
    \cmidrule{2-14}
    \multirow{3}{*}{\textit{Random}} & 4 & 60.5 & 23.6 & 23.6 & 2.0 & 10.2 & 21.1 & 23.4 & 1.3 & 12.9 & 16.3 & 19.5 & \ccc{-19.0} \\
    & 8 & 47.9 & 11.8 & 18.1 & 1.4 & 5.7 & 14.1 & 11.5 & 0.9 & 7.1 & 8.2 & 12.7 & \ccc{-25.7} \\
    & 16 & 55.7 & 26.8 & 14.0 & 2.0 & 32.0 & 6.8 & 42.1 & 0.6 & 2.9 & 21.7 & 20.5 & \ccc{-18.0} \\

    \cmidrule{2-14}
    \multirow{3}{*}{\textit{Similarity}} & 4 & 86.5 & 39.5 & 50.7 & 13.6 & 65.3 & 58.8 & 59.2 & 27.4 & 49.2 & 54.7 & 50.5 & \ccc{+12.0} \\
    & 8 & 91.2 & 51.1 & 65.6 & 17.5 & 74.6 & 65.1 & 67.0 & 36.0 & 58.0 & 65.3 & 59.1 & \ccc{+20.6} \\
    & 16 & \textbf{94.3} & \textbf{64.6} & \textbf{73.7} & \textbf{23.7} & \textbf{83.2} & \textbf{73.0} & \textbf{76.6} & \textbf{47.2} & \textbf{67.0} & \textbf{73.2} & \textbf{67.7} & \ccc{+29.2} \\

    \cmidrule(lr){1-14}
    \rowcolor{DrawioYellow}
    \multicolumn{14}{c}{\emph{Phi-4-MM}} \\ [-2ex] \\
    \textit{Vanilla} & - & 85.8 & 56.6 & 29.5 & 14.7 & 25.5 & 54.2 & \textbf{47.7} & 3.3 & 29.3 & 48.9 & 39.6 \\
    
    \cmidrule{2-14}
    \multirow{3}{*}{\textit{Random}} & 4 & 23.2 & 16.9 & 15.9 & 3.3 & 7.3 & 19.6 & 15.5 & 1.2 & 10.7 & 8.2 & 12.2 & \ccc{-27.4} \\
    & 8 & 22.0 & 10.2 & 13.7 & 1.7 & 10.0 & 11.2 & 12.2 & 1.3 & 7.0 & 8.6 & 9.8 & \ccc{-29.8} \\
    & 16 & 63.2 & 49.3 & 12.9 & 9.4 & 23.1 & 6.8 & 8.6 & 0.9 & 3.9 & 14.9 & 19.3 & \ccc{-20.3} \\

    \cmidrule{2-14}
    \multirow{3}{*}{\textit{Similarity}} & 4 & 84.7 & 57.8 & 36.4 & 24.5 & 58.9 & 49.0 & 45.2 & 33.1 & 45.2 & 43.1 & 47.8 & \ccc{+8.2} \\
    & 8 & \textbf{86.0} & 68.4 & 42.9 & \textbf{29.6} & 64.6 & 54.4 & 42.6 & 32.9 & 49.2 & \textbf{45.5} & 51.6 & \ccc{+12.0} \\
    & 16 & 81.9 & \textbf{70.6} & \textbf{43.2} & 28.8 & \textbf{69.5} & \textbf{60.5} & 45.1 & \textbf{34.2} & \textbf{53.9} & {42.3} & \textbf{53.0} & \ccc{+13.4} \\
    
    \bottomrule
  \end{tabular}
  }
  \label{supp:tab:cw:Similarity_vs_models_lmms}
\end{subtable}
\end{table*}

\section{Open-world results}
\label{supp:sec:ow}

This section details the results of our Open-World (OW) experiments.
We conduct a comprehensive evaluation across all five models and ten datasets, testing three In-Context Learning (ICL) variants: \textit{Random Context}, \textit{Pseudo ICL}, and \ours.
For each method, we report performance at 4, 8, and 16 shots.

The detailed results are organized by model family and metric as follows:
\begin{itemize}
    \item \textbf{Main results.}
    \cref{supp:tab:ow:main_qwen,supp:tab:ow:main_llava,supp:tab:ow:main_phi} present the aggregated performance for the Qwen series, LLaVa OneVision, and the Phi series, respectively.
    These tables group results by dataset type: \textit{Prototypical}, \textit{Non-prototypical}, \textit{Fine-grained}, and \textit{Very fine-grained}.

    \item \textbf{Metric-specific breakdowns.}
    We provide granular breakdowns for each metric across all ten datasets in the subsequent tables:
    \begin{itemize}
        \item \textbf{Semantic Similarity} (\texttt{SS}) in \cref{supp:tab:ow-ss-qwen,supp:tab:ow-ss-llava,supp:tab:ow-ss-phi}.

        \item \textbf{Concept Similarity} (\texttt{bCS}) in \cref{supp:tab:ow-bcs-qwen,supp:tab:ow-bcs-llava,supp:tab:ow-bcs-phi}.

        \item \textbf{Median Concept Similarity} (\texttt{mCS}) in \cref{supp:tab:ow-mcs-qwen,supp:tab:ow-mcs-llava,supp:tab:ow-mcs-phi}.

        \item \textbf{Llama Inclusion} (\texttt{LI}) in \cref{supp:tab:ow-li-qwen,supp:tab:ow-li-llava,supp:tab:ow-li-phi}.
    \end{itemize}
\end{itemize}

In all tables, we report the average score (\textit{Avg.}) and the improvement ($\Delta$) relative to the \textit{Zero-Shot} baseline.

\myparagraph{Results discussion.}
The open-world results (\cref{supp:tab:ow:main_qwen,supp:tab:ow:main_llava,supp:tab:ow:main_phi}) demonstrate the consistent superiority of \ours over both baselines and other ICL variants.
Across all model families, constructing the context with our \ours yields significant improvements in \textit{Semantic Similarity} (\texttt{SS}), \textit{Median Concept Similarity} (\texttt{mCS}), and \textit{Llama Inclusion} (\texttt{LI}) compared to the \textit{Zero-Shot} baseline.
For instance, with Qwen2.5-VL, our 16-shot configuration improves \texttt{SS} on \textit{Prototypical} datasets from 47.9 to 67.7, \texttt{mCS} from 31.1 to 67.2, and \texttt{LI} from 82.9 to 94.9.
In contrast, \textit{Random Context} acts as a distractor, causing performance degradation.
This is particularly evident in LLaVa OneVision (\cref{supp:tab:ow:main_llava}), where 16-shot \textit{Random} context causes \texttt{SS} to lower from 56.2 (\textit{Zero-Shot}) to 29.3, \texttt{mCS} from 53.4 to 29.6, and \texttt{LI} from 53.2 to 14.0, whereas \ours recovers and boosts performance to 74.0, 74.0, and 72.2, respectively.

Furthermore, \ours proves effective in handling fine-grained tasks, a setting where standard models typically struggle.
In the \textit{Very fine-grained} category, \ours achieves consistent gains.
Notably, Phi-3.5-Vision (\cref{supp:tab:ow:main_phi}) sees its \texttt{LI} score nearly double, jumping from 54.2 in \textit{Zero-Shot} to 99.6 with our \ours in the default 16-shot setting.
Similarly, Qwen2.5-VL improves from 69.0 to 93.6 in the same category.
While \textit{Pseudo ICL} generally offers some improvement over \textit{Random}, it lacks the stability of our method, often underperforming the zero-shot baseline.

Finally, the results highlight the importance of scaling the number of shots when using relevant context.
While 4-shot performance with \ours already surpasses the \textit{Zero-Shot} baseline in most metrics, such as the \textit{Fine-grained} \texttt{bCS} score for Qwen2-VL rising from 62.9 to 66.4, the performance gap widens significantly at 16 shots (reaching 61.1 \texttt{SS}, 72.0 \texttt{bCS}, and 57.3 \texttt{mCS}).
This behavior suggests that \ours successfully leverages the additional diverse unlabeled images to construct a more informative context, a capability that is absent when using \textit{Random} or \textit{Pseudo}-labeled contexts.

\begin{table*}

\newcommand\random{\multirow{3}{*}{\makecell{\textit{Random} \\ \textit{Ctx}}}}
\newcommand\pseudo{\multirow{3}{*}{\makecell{\textit{Pseudo} \\ \textit{ICL}}}}
\newcommand\ourr{\multirow{-3}{*}{\ours}}

\caption{
\textbf{Open-world results.}
We report results for \textit{Llama Inclusion} (\texttt{LI}), \textit{Semantic Similarity} (\texttt{SS}), \textit{Concept Similarity} (\texttt{bCS}), and \textit{Median Concept Similarity} (\texttt{mCS}).
\inlineColorbox{DrawioPurple}{Purple} indicates our \ours.
Higher is better on all metrics.
For each LMM, \textbf{bold} indicates the best result.
Results are split in \cref{supp:tab:ow:main_qwen,supp:tab:ow:main_llava,supp:tab:ow:main_phi} for readability.
}
\begin{subtable}{\textwidth}
\caption{\textbf{Qwen2-VL and Qwen2.5-VL.}}
  \centering
  \resizebox{\linewidth}{!}{
  \begin{tabular}{rc c cccc c cccc c cccc c cccc}
    \toprule
    \multirow[c]{2.5}{*}{Method} & \multirow[c]{2.5}{*}{Shots} && \multicolumn{4}{c}{Prototypical} && \multicolumn{4}{c}{Non-prototypical} && \multicolumn{4}{c}{Fine-grained} && \multicolumn{4}{c}{Very fine-grained} \\
    \cmidrule{4-7} \cmidrule{9-12} \cmidrule{14-17} \cmidrule{19-22}
    &&&
        \texttt{LI} & \texttt{SS} & \texttt{bCS} & \texttt{mCS}
    &&
        \texttt{LI} & \texttt{SS} & \texttt{bCS} & \texttt{mCS}
    &&
        \texttt{LI} & \texttt{SS} & \texttt{bCS} & \texttt{mCS}
    &&
        \texttt{LI} & \texttt{SS} & \texttt{bCS} & \texttt{mCS} \\
    \midrule

    \rowcolor{DrawioYellow}
    \multicolumn{22}{c}{\emph{Qwen2-VL 7B}} \\ [-2ex] \\

    \textit{Zero-Shot} & - &&
        78.7 & 51.9 & 76.0 & 43.7 &&
        42.6 & 30.8 & 49.8 & 29.2 &&
        64.0 & 39.2 & 62.9 & 31.9 &&
        63.0 & 34.5 & 43.4 & 33.1
        \\
    \cmidrule(lr){2-22}
    \random & 4 &&
         48.0 & 48.9 & 67.7 & 39.5 &&
         30.0 & 31.2 & 49.4 & 29.0 &&
         42.9 & 40.5 & 55.9 & 37.3 &&
         41.0 & 37.0 & 46.3 & 34.7
        \\
    & 8 &&
         48.2 & 49.2 & 66.0 & 41.9 &&
         24.0 & 27.6 & 45.7 & 26.6 &&
         37.6 & 36.8 & 49.7 & 35.7 &&
         37.0 & 31.4 & 39.2 & 29.7
        \\
    & 16 &&
        24.4 & 41.4 & 52.7 & 39.7 &&
        17.1 & 23.4 & 41.3 & 23.0 &&
        31.7 & 34.4 & 44.8 & 34.6 &&
        31.1 & 29.2 & 34.1 & 27.9
        \\
    \cmidrule(lr){2-22}
    \pseudo & 4 &&
        81.1 & 53.4 & \textbf{76.2} & 44.4 &&
        42.8 & 31.2 & {50.1} & 26.9 &&
        53.1 & 40.2 & 64.4 & 30.7 &&
        49.1 & 38.9 & 49.1 & 38.6
        \\
    & 8 &&
        76.8 & 52.9 & 75.2 & 44.6 &&
        37.7 & 34.0 & \textbf{52.1} & 31.9 &&
        47.3 & 36.6 & 58.9 & 32.4 &&
        47.0 & 37.1 & 46.4 & 36.3
        \\
    & 16 &&
        73.7 & 52.6 & 74.0 & 46.3 &&
        35.1 & 30.3 & 47.9 & 27.3 &&
        48.1 & 37.9 & 59.4 & 33.3 &&
        49.8 & 36.9 & 45.9 & 35.7
        \\
    \cmidrule(lr){2-22}
    \rowcolor{DrawioPurple} \cellcolor{white}
    & 4 &&
        90.8 & 59.8 & 73.7 & 60.5 &&
        \textbf{65.9} & 36.9 & 46.7 & 37.5 &&
        \textbf{88.8} & 55.1 & 66.4 & 52.1 &&
        93.4 & 36.1 & 44.9 & 35.8
        \\
    \rowcolor{DrawioPurple} \cellcolor{white}
    & 8 &&
        \textbf{92.1} & 61.4 & 72.0 & 59.6 &&
        62.1 & 37.9 & 45.3 & 36.6 &&
        86.6 & 58.9 & 69.8 & 54.0 &&
        \textbf{95.6} & 35.2 & 43.0 & 35.4
        \\
    \rowcolor{DrawioPurple}
    \cellcolor{white} \ourr & 16 &&
        91.5 & \textbf{65.6} & 74.3 & \textbf{63.5} &&
        61.6 & \textbf{41.9} & 49.4 & \textbf{40.5} &&
        87.3 & \textbf{61.1} & \textbf{72.0} & \textbf{57.3} &&
        91.5 & \textbf{42.5} & \textbf{50.2} & \textbf{39.8}
        \\

    \cmidrule{1-22}
    \rowcolor{DrawioYellow}
    \multicolumn{22}{c}{\emph{Qwen2.5-VL 7B}} \\ [-2ex] \\
    
    \textit{Zero-Shot} & - &&
        82.9 & 47.9 & \textbf{79.9} & 31.1 &&
        45.9 & 30.5 & 54.0 & 24.8 &&
        73.8 & 47.0 & \textbf{78.9} & 29.5 &&
        69.0 & 45.8 & 68.6 & 27.1
        \\
    \cmidrule(lr){2-22}
    \random & 4 &&
         81.9 & 52.3 & 77.8 & 41.6 &&
         48.5 & 34.5 & 57.5 & 26.6 &&
         72.3 & 48.5 & 76.9 & 34.4 &&
         35.6 & 50.3 & 68.4 & 38.1
        \\
    & 8 &&
         82.1 & 53.0 & 78.0 & 42.8 &&
         48.5 & 34.9 & 58.2 & 26.8 &&
         72.1 & 49.0 & 76.8 & 36.2 &&
         34.4 & 52.0 & 69.8 & 40.8
        \\
    & 16 &&
        82.5 & 53.5 & 78.2 & 43.1 &&
        48.2 & 35.7 & \textbf{58.9} & 27.3 &&
        70.6 & 49.0  & 76.7 & 36.3 &&
        34.7 & \textbf{52.4} & \textbf{70.2} & \textbf{42.0}
        \\
    \cmidrule(lr){2-22}
    \pseudo & 4 &&
        80.6 & 49.3 & 78.6 & 31.0 &&
        42.7 & 31.6 & 53.2 & 24.2 &&
        63.8 & 45.1 & 74.8 & 28.4 &&
        39.7 & 46.0 & 63.4 & 25.5
        \\
    & 8 &&
         80.5 & 49.3 & 78.8 & 30.6 &&
         43.7 & 31.9 & 53.3 & 24.4 &&
         63.6 & 44.9 & 75.2 & 27.9 &&
         40.5 & 46.3 & 64.9 & 24.0
        \\
    & 16 &&
         80.7 & 49.3 & 79.0 & 30.9 &&
         42.9 & 32.3 & 53.7 & 24.5 &&
         65.5 & 45.7 & 75.5 & 28.5 &&
         35.9 & 47.6 & 65.5 & 25.0
        \\
    \cmidrule(lr){2-22}
    \rowcolor{DrawioPurple} \cellcolor{white}
    & 4 &&
         90.3 & 68.0 & 69.7 & 67.0 &&
         61.0 & 40.1 & 44.1 & 39.5 &&
         82.3 & 60.4 & 63.2 & 59.2 &&
         88.7 & 39.6 & 39.6 & 39.6
        \\
    \rowcolor{DrawioPurple} \cellcolor{white}
    & 8 &&
         89.7 & \textbf{70.0} & 71.2 & \textbf{69.4} &&
         58.5 & 39.0 & 39.6 & 39.1 &&
         80.5 & \textbf{61.4} & 62.6 & \textbf{60.9} &&
         87.6 & 37.9 & 37.9 & 37.9
        \\
    \rowcolor{DrawioPurple}
    \cellcolor{white} \ourr & 16 &&
        \textbf{94.9} & {67.7} & 68.1 & {67.2} &&
        \textbf{67.6} & \textbf{42.6} & 45.1 & \textbf{42.3} &&
        \textbf{86.3} & {60.1} & 60.9 & {59.7} &&
        \textbf{93.6} & 36.4 & 36.6 & 36.5
        \\
    
    \bottomrule
  \end{tabular}}
  \label{supp:tab:ow:main_qwen}
\end{subtable}
\end{table*}

\begin{table*}

\newcommand\random{\multirow{3}{*}{\makecell{\textit{Random} \\ \textit{Ctx}}}}
\newcommand\pseudo{\multirow{3}{*}{\makecell{\textit{Pseudo} \\ \textit{ICL}}}}
\newcommand\ourr{\multirow{-3}{*}{\ours}}

\ContinuedFloat 
\begin{subtable}{\textwidth}
\caption{\textbf{LLaVa OneVision}.}
  \centering
  \resizebox{\linewidth}{!}{
  \begin{tabular}{rc c cccc c cccc c cccc c cccc}
    \toprule
    \multirow[c]{2.5}{*}{Method} & \multirow[c]{2.5}{*}{Shots} && \multicolumn{4}{c}{Prototypical} && \multicolumn{4}{c}{Non-prototypical} && \multicolumn{4}{c}{Fine-grained} && \multicolumn{4}{c}{Very fine-grained} \\
    \cmidrule{4-7} \cmidrule{9-12} \cmidrule{14-17} \cmidrule{19-22}
    &&&
        \texttt{LI} & \texttt{SS} & \texttt{bCS} & \texttt{mCS}
    &&
        \texttt{LI} & \texttt{SS} & \texttt{bCS} & \texttt{mCS}
    &&
        \texttt{LI} & \texttt{SS} & \texttt{bCS} & \texttt{mCS}
    &&
        \texttt{LI} & \texttt{SS} & \texttt{bCS} & \texttt{mCS} \\
    \midrule

    \rowcolor{DrawioYellow}
    \multicolumn{22}{c}{\emph{LLaVa OneVision 7B}} \\ [-2ex] \\
    
    \textit{Zero-Shot} & - &&
        53.2 & 56.2 & 62.0 & 53.4 &&
        28.1 & 31.6 & 43.8 & 30.2 &&
        40.4 & 39.0 & 43.9 & 37.2 &&
        \textbf{76.7} & 31.8 & 32.3 & 30.9
        \\
    \cmidrule(lr){2-22}
    \random & 4 &&
         13.9 & 33.7 & 35.1 & 33.9 &&
          7.3 & 25.3 & 35.7 & 23.7 &&
         20.8 & 34.2 & 34.5 & 34.2 &&
         74.1 & 30.8 & 30.8 & 30.8
        \\
    & 8 &&
         14.2 & 33.1 & 35.5 & 33.2 &&
          6.8 & 25.9 & 37.9 & 24.0 &&
         20.7 & 34.3 & 34.6 & 34.3 &&
         74.1 & 30.8 & 30.8 & 30.8
        \\
    & 16 &&
        14.0 & 29.3 & 36.7 & 29.6 &&
         8.6 & 26.0 & 39.3 & 24.1 &&
        21.0 & 33.2 & 35.8 & 33.4 &&
        75.8 & 30.8 & 30.8 & 30.8
        \\
    \cmidrule(lr){2-22}
    \pseudo & 4 &&
        19.7 & 31.2 & 41.1 & 30.5 &&
         3.3 & 18.3 & 31.5 & 19.7 &&
        20.3 & 35.5 & 40.0 & 34.7 &&
        70.4 & 30.6 & 31.1 & 29.8
        \\
    & 8 &&
         55.4 & 54.1 & 63.9 & 48.9 &&
         29.9 & 27.5 & 45.6 & 24.0 &&
         33.9 & 37.1 & 45.7 & 34.2 &&
         74.4 & 31.3 & 31.6 & 30.7
        \\
    & 16 &&
         58.1 & 55.0 & 65.3 & 49.3 &&
         31.8 & 28.1 & 46.3 & 25.1 &&
         33.6 & 37.1 & 45.7 & 34.5 &&
         73.3 & 31.7 & 32.3 & 30.5
        \\
    \cmidrule(lr){2-22}
    \rowcolor{DrawioPurple} \cellcolor{white}
    & 4 &&
         84.9 & 71.8 & 71.8 & 71.8 &&
         52.5 & 41.7 & 41.8 & 41.8 &&
         70.2 & 42.7 & 42.7 & 42.6 &&
         67.4 & \textbf{35.1} & \textbf{35.3} & \textbf{34.6}
        \\
    \rowcolor{DrawioPurple} \cellcolor{white}
    & 8 &&
         \textbf{87.4} & 73.8 & 73.8 & 73.8 &&
         55.3 & 46.3 & 46.3 & 46.3 &&
         \textbf{73.9} & 41.3 & 41.3 & 41.3 &&
         54.9 & 32.3 & 32.3 & 32.3
        \\
    \rowcolor{DrawioPurple}
    \cellcolor{white} \ourr & 16 &&
        {72.2} & \textbf{74.0} & \textbf{74.0} & \textbf{74.0} &&
        \textbf{61.7} & \textbf{55.3} & \textbf{55.3} & \textbf{55.3} &&
        {55.1} & \textbf{46.0} & \textbf{46.0} & \textbf{46.0} &&
        74.2 & {32.9} & {32.8} & {32.9}
        \\
    
    \bottomrule
  \end{tabular}}
  \label{supp:tab:ow:main_llava}
\end{subtable}
\end{table*}

\begin{table*}

\newcommand\random{\multirow{3}{*}{\makecell{\textit{Random} \\ \textit{Ctx}}}}
\newcommand\pseudo{\multirow{3}{*}{\makecell{\textit{Pseudo} \\ \textit{ICL}}}}
\newcommand\ourr{\multirow{-3}{*}{\ours}}

\ContinuedFloat 
\begin{subtable}{\textwidth}
\caption{\textbf{Phi-3.5-Vision and Phi-4-Multimodal.}}
  \centering
  \resizebox{\linewidth}{!}{
  \begin{tabular}{rc c cccc c cccc c cccc c cccc}
    \toprule
    \multirow[c]{2.5}{*}{Method} & \multirow[c]{2.5}{*}{Shots} && \multicolumn{4}{c}{Prototypical} && \multicolumn{4}{c}{Non-prototypical} && \multicolumn{4}{c}{Fine-grained} && \multicolumn{4}{c}{Very fine-grained} \\
    \cmidrule{4-7} \cmidrule{9-12} \cmidrule{14-17} \cmidrule{19-22}
    &&&
        \texttt{LI} & \texttt{SS} & \texttt{bCS} & \texttt{mCS}
    &&
        \texttt{LI} & \texttt{SS} & \texttt{bCS} & \texttt{mCS}
    &&
        \texttt{LI} & \texttt{SS} & \texttt{bCS} & \texttt{mCS}
    &&
        \texttt{LI} & \texttt{SS} & \texttt{bCS} & \texttt{mCS} \\
    \midrule

    \rowcolor{DrawioYellow}
    \multicolumn{22}{c}{\emph{Phi-3.5-Vision}} \\ [-2ex] \\
    
    \textit{Zero-Shot} & - &&
        60.7 & 48.2 & {65.6} & 46.1 &&
        28.7 & 24.9 & {36.7} & 24.1 &&
        50.7 & 32.1 & 47.2 & 31.3 &&
        54.2 & 29.5 & 36.3 & 29.8
        \\
    \cmidrule(lr){2-22}
    \random & 4 &&
         49.2 & 57.5 & 60.9 & 57.3 &&
         14.7 & 27.2 & 35.4 & 27.2 &&
         29.1 & 35.9 & 43.1 & 35.6 &&
         51.7 & 27.7 & 32.3 & 26.5
        \\
    & 8 &&
         45.0 & 53.4 & 57.6 & 53.3 &&
         11.3 & 27.1 & 37.5 & 27.5 &&
         26.7 & 34.7 & 41.5 & 34.7 &&
         56.7 & 28.2 & 31.7 & 27.3
        \\
    & 16 &&
        40.3 & 48.3 & 54.4 & 48.4 &&
        10.4 & 26.3 & 36.4 & 27.4 &&
        24.1 & 26.3 & 36.4 & 27.4 &&
        58.2 & 27.1 & 31.7 & 26.3
        \\
    \cmidrule(lr){2-22}
    \pseudo & 4 &&
        54.1 & 44.1 & 61.7 & 40.1 &&
        23.7 & 22.8 & 35.1 & 21.5 &&
        43.1 & 33.0 & {48.6} & 29.7 &&
        24.9 & 32.4 & \textbf{41.8} & 32.5
        \\
    & 8 &&
         59.3 & 46.7 & 65.3 & 41.1 &&
         24.1 & 22.3 & 34.6 & 21.3 &&
         34.5 & 32.3 & 50.0 & 28.2 &&
         25.8 & 32.8 & 41.0 & 34.0
        \\
    & 16 &&
         55.7 & 46.1 & 63.7 & 43.1 &&
         25.2 & 22.9 & 34.9 & 21.4 &&
         35.0 & 33.8 & \textbf{51.9} & 29.5 &&
         30.0 & 33.3 & 40.0 & 33.6
        \\
    \cmidrule(lr){2-22}
    \rowcolor{DrawioPurple} \cellcolor{white}
    & 4 &&
         80.8 & 63.3 & 64.6 & 63.1 &&
         52.3 & \textbf{36.5} & \textbf{37.6} & \textbf{36.7} &&
         84.8 & 39.1 & 44.9 & 39.2 &&
         88.8 & \textbf{36.6} & 38.1 & \textbf{36.1}
        \\
    \rowcolor{DrawioPurple} \cellcolor{white}
    & 8 &&
         84.7 & \textbf{64.7} & \textbf{66.2} & \textbf{65.0}  &&
         \textbf{59.3} & 31.6 & 36.0 & 32.9 &&
         84.7 & \textbf{42.1} & 48.9 & \textbf{42.6} &&
         92.9 & 32.0 & 37.3 & 32.4
        \\
    \rowcolor{DrawioPurple}
    \cellcolor{white} \ourr & 16 &&
        \textbf{92.1} & {59.7} & 63.2 & {60.3} &&
        {58.3} & {30.0} & 35.2 & {31.7} &&
        \textbf{88.1} & {39.2} & 45.7 & {42.1} &&
        \textbf{99.6} & {33.0} & 33.6 & {33.1}
        \\
    
    \cmidrule{1-22}
    \rowcolor{DrawioYellow}
    \multicolumn{22}{c}{\emph{Phi-4-Multimodal}} \\ [-2ex] \\
    
    \textit{Zero-Shot} & - &&
        49.8 & 57.4 & 58.7 & 57.2 &&
        21.2 & 29.2 & 32.7 & 29.2 &&
        37.7 & 39.2 & 39.2 & 39.1 &&
        73.6 & 31.6 & 31.7 & 31.6
        \\
    \cmidrule(lr){2-22}
    \random & 4 &&
         10.9 & 32.6 & 32.6 & 32.6 &&
          9.6 & 30.1 & 30.2 & 30.1 &&
         29.9 & 38.1 & 38.1 & 38.1 &&
         70.0 & 30.6 & 30.6 & 30.6
        \\
    & 8 &&
         13.0 & 34.2 & 34.2 & 34.2 &&
          7.7 & 29.6 & 29.7 & 29.6 &&
         31.2 & 38.5 & 38.5 & 38.5 &&
         71.7 & 30.7 & 30.7 & 30.7
        \\
    & 16 &&
         11.3 & 32.8 & 32.8 & 32.8 &&
          5.8 & 28.5 & 28.6 & 28.5 &&
         30.9 & 37.9 & 37.9 & 37.9 &&
         72.3 & 30.8 & 30.7 & 30.6
        \\
    \cmidrule(lr){2-22}
    \pseudo & 4 &&
        51.9 & 61.5 & 62.0 & 61.5 &&
        15.1 & 26.6 & 31.2 & 26.7 &&
        37.7 & 41.6 & 41.7 & 41.5 &&
        72.8 & 31.9 & 31.9 & 31.9
        \\
    & 8 &&
         49.2 & 59.9 & 60.0 & 59.9 &&
         14.9 & 32.0 & 33.3 & 32.0 &&
         36.4 & 41.4 & 41.4 & 41.4 &&
         73.2 & 31.5 & 31.5 & 31.5
        \\
    & 16 &&
         51.6 & 61.5 & 61.5 & 61.5 &&
         19.1 & 24.7 & 31.1 & 24.5 &&
         35.7 & 41.1 & 41.1 & 41.0 &&
         71.6 & 32.0 & 32.0 & 32.0
        \\
    \cmidrule(lr){2-22}
    \rowcolor{DrawioPurple} \cellcolor{white}
    & 4 &&
         87.0 & \textbf{68.5} & 69.4 & \textbf{68.6} &&
         40.0 & 35.9 & 36.6 & 35.6 &&
         \textbf{80.7} & 49.7 & 53.2 & 49.2 &&
         \textbf{92.4} & 34.6 & 36.3 & 34.4
        \\
    \rowcolor{DrawioPurple} \cellcolor{white}
    & 8 &&
         85.3 & 65.9 & 68.4 & 65.7 &&
         61.5 & 37.4 & 43.8 & 34.7 &&
         79.3 & \textbf{57.0} & \textbf{60.3} & \textbf{56.8} &&
         91.5 & 38.2 & 38.9 & 37.7
        \\
    \rowcolor{DrawioPurple}
    \cellcolor{white} \ourr & 16 &&
        \textbf{91.5} & {65.5} & \textbf{70.1} & {66.4} &&
        \textbf{67.6} & \textbf{43.2} & \textbf{46.1} & \textbf{43.4} &&
        {79.1} & {53.3} & {55.5} & {53.0} &&
        {75.2} & \textbf{40.2} & \textbf{42.5} & \textbf{37.9}
        \\
    
    \bottomrule
  \end{tabular}}
  \label{supp:tab:ow:main_phi}
\end{subtable}
\end{table*}

\begin{table*}

\newcommand\random{\multirow{3}{*}{\makecell{\textit{Random} \\ \textit{Ctx}}}}
\newcommand\pseudo{\multirow{3}{*}{\makecell{\textit{Pseudo} \\ \textit{ICL}}}}
\newcommand\ourr{\multirow{-3}{*}{\ours}}

\caption{
\textbf{Open-world results. \textit{Semantic Similarity}} (\texttt{SS}) on the ten datasets.
\inlineColorbox{DrawioPurple}{Purple} indicates our \ours.
Higher is better.
For each LMM, \textbf{bold} indicates the best result.
$\Delta$ computed \wrt the average (Avg.) of \textit{Zero-Shot}.
Results are split in \cref{supp:tab:ow-ss-qwen,supp:tab:ow-ss-llava,supp:tab:ow-ss-phi} for readability.
}
\begin{subtable}{\textwidth}
\caption{\textbf{Qwen2-VL and Qwen2.5-VL.}}
  \centering
  \resizebox{\linewidth}{!}{
  \begin{tabular}{rc c cccccccccccc}
    \toprule
    Method & Shots && C101 & DTD & ESAT & FGVC & FLWR & FOOD & PETS & CARS & S397 & U101 & Avg. & $\Delta$ \\
    \midrule

    \rowcolor{DrawioYellow}
    \multicolumn{15}{c}{\emph{Qwen2-VL 7B}} \\ [-2ex] \\

    \textit{Zero-Shot} & - &&
        55.8 & 28.6 & 20.7 & 20.6 & 41.7 & 50.7 & 25.1 & 48.3 & 48.1 & 43.1 & 38.3 \\
    \cmidrule(lr){2-15}
    \random & 4 &&
        54.8 & 25.5 & 27.3 & 20.8 & 35.7 & 52.5 & 33.1 & 53.3 & 42.9 & 41.0 & 38.7 & \ccc{-0.4} \\
    & 8 &&
       56.1 & 21.5 & 27.9 & 20.3 & 30.8 & 50.9 & 28.7 & 42.5 & 42.3 & 33.3 & 35.4 & \ccc{-2.9} \\
    & 16 &&
       50.8 & 18.8 & 26.4 & 22.8 & 26.7 & 46.5 & 30.0 & 35.5 & 32.0 & 25.1 & 31.5 & \ccc{-6.8} \\
    \cmidrule(lr){2-15}
    \pseudo & 4 &&
       57.6 & 29.7 & 18.6 & 21.0 & 40.5 & 52.6 & 27.3 & \textbf{56.8} & 49.2 & 45.5 & 39.9 & \ccc{+1.6} \\
    & 8 &&
       57.1 & 28.4 & 29.1 & 20.5 & 35.6 & 50.1 & 24.1 & 53.6 & 48.6 & 44.5 & 39.2 & \ccc{+0.9} \\
    & 16 &&
       58.2 & 29.3 & 16.6 & 20.3 & 39.0 & 49.4 & 25.4 & 53.5 & 46.9 & 44.9 & 38.4 & \ccc{+0.1} \\
    \cmidrule(lr){2-15}
    \rowcolor{DrawioPurple} \cellcolor{white}
    & 4 &&
       62.5 & 29.0 & 32.2 & 26.3 & 49.3 & 66.1 & 49.7 & 45.9 & 57.0 & 49.6 & 46.8 & \ccc{+8.5} \\
    \rowcolor{DrawioPurple} \cellcolor{white}
    & 8 &&
       61.3 & 35.7 & 31.3 & 25.0 & \textbf{57.3} & 66.4 & 53.1 & 45.4 & 61.5 & 46.9 & 48.4 & \ccc{+10.5} \\
    \rowcolor{DrawioPurple}
    \cellcolor{white} \ourr & 16 &&
       \textbf{68.4} & \textbf{36.5} & \textbf{37.0} & \textbf{31.4} & 54.5 & \textbf{67.9} & \textbf{61.0} & {53.7} & \textbf{62.7} & \textbf{52.3} & \textbf{52.5} & \ccc{+14.2} \\

    \cmidrule{1-15}
    \rowcolor{DrawioYellow}
    \multicolumn{15}{c}{\emph{Qwen2.5-VL 7B}} \\ [-2ex] \\
    
    \textit{Zero-Shot} & - &&
        48.8 & 28.3 & 19.0 & 36.7 & 47.4 & 52.4 & 41.1 & 54.9 & 47.0 & 44.2 & 42.0 \\
    \cmidrule(lr){2-15}
    \random & 4 &&
        55.8 & 29.1 & 28.3 & 35.4 & 49.6 & 47.7 & 48.1 & 65.2 & 48.8 & 46.2 & 45.4 & \ccc{+3.4} \\
    & 8 &&
        56.7 & 29.9 & 29.5 & \textbf{36.0} & 50.1 & 47.7 & 49.2 & 68.1 & 49.3 & 45.3 & 46.2 & \ccc{+4.2} \\
    & 16 &&
        57.6 & 30.2 & 31.7 & 35.8 & 51.0 & 47.8 & 48.2 & \textbf{69.0} & 49.5 & 45.1 & 46.6 & \ccc{+4.6} \\
    \cmidrule(lr){2-15}
    \pseudo & 4 &&
        50.8 & 29.5 & 19.1 & 33.0 & 47.5 & 49.0 & 38.9 & 58.9 & 47.9 & 46.0 & 42.1 & \ccc{+0.1} \\
    & 8 &&
        50.3 & 29.8 & 19.2 & 34.0 & 48.4 & 49.2 & 37.1 & 58.6 & 48.3 & 46.6 & 42.2 & \ccc{+0.2} \\
    & 16 &&
        50.2 & 30.4 & 20.6 & 34.7 & 47.3 & 50.1 & 39.5 & 60.6 & 48.5 & 46.1 & 42.8 & \ccc{+0.8} \\
    \cmidrule(lr){2-15}
    \rowcolor{DrawioPurple} \cellcolor{white}
    & 4 &&
        68.0 & 39.8 & 30.9 & 29.6 & 64.5 & \textbf{58.4} & 58.4 & 49.5 & \textbf{68.0} & 49.5 & \textbf{51.7} & \ccc{+9.7} \\
    \rowcolor{DrawioPurple} \cellcolor{white}
    & 8 &&
        \textbf{73.6} & \textbf{42.3} & 34.4 & 29.0 & \textbf{66.9} & \textbf{58.4} & \textbf{58.9} & 46.7 & 66.3 & 40.4 & \textbf{51.7} & \ccc{+9.7} \\
    \rowcolor{DrawioPurple}
    \cellcolor{white} \ourr & 16 &&
        70.3 & 41.7 & \textbf{35.1} & 29.6 & 66.4 & 56.8 & 57.1 & 43.3 & 65.0 & \textbf{51.0} & 51.6 & \ccc{+9.6} \\
    
    \bottomrule
  \end{tabular}}
  \label{supp:tab:ow-ss-qwen}
\end{subtable}
\end{table*}

\begin{table*}

\newcommand\random{\multirow{3}{*}{\makecell{\textit{Random} \\ \textit{Ctx}}}}
\newcommand\pseudo{\multirow{3}{*}{\makecell{\textit{Pseudo} \\ \textit{ICL}}}}
\newcommand\ourr{\multirow{-3}{*}{\ours}}

\ContinuedFloat
\begin{subtable}{\textwidth}
\caption{\textbf{LLaVa OneVision}.}
  \centering
  \resizebox{\linewidth}{!}{
  \begin{tabular}{rc c cccccccccccc}
    \toprule
    Method & Shots && C101 & DTD & ESAT & FGVC & FLWR & FOOD & PETS & CARS & S397 & U101 & Avg. & $\Delta$ \\
    \midrule

    \rowcolor{DrawioYellow}
    \multicolumn{15}{c}{\emph{LLaVa OneVision 7B}} \\ [-2ex] \\

    \textit{Zero-Shot} & - &&
        68.9 & 32.0 & 19.4 & 29.4 & 37.5 & 41.6 & 37.8 & 34.3 & 43.4 & 43.4 & 38.8 \\
    \cmidrule(lr){2-15}
    \random & 4 &&
        41.2 & 24.7 & 24.9 & 29.4 & 32.1 & 33.0 & 37.6 & 32.2 & 26.3 & 26.3 & 30.8 & \ccc{-8.0} \\
    & 8 &&
        40.1 & 24.1 & 27.1 & 29.4 & 32.0 & 33.3 & 37.7 & 32.2 & 26.1 & 26.4 & 30.8 & \ccc{-8.0} \\
    & 16 &&
        31.5 & 25.3 & 26.3 & 29.4 & 27.8 & 34.0 & 37.8 & 32.2 & 27.2 & 26.3 & 29.8 & \ccc{-9.0} \\
    \cmidrule(lr){2-15}
    \pseudo & 4 &&
        19.9 & 17.2 & 17.6 & 29.4 & 26.3 & 45.1 & 35.1 & 31.8 & 42.5 & 20.0 & 28.5 & \ccc{-10.3} \\
    & 8 &&
        67.0 & 31.0 & 23.5 & 29.4 & 33.8 & 39.5 & 38.1 & 33.2 & 41.2 & 27.9 & 36.4 & \ccc{-2.4} \\
    & 16 &&
        66.5 & 30.1 & 23.2 & 29.4 & 32.7 & 40.6 & 38.2 & 34.0 & 43.5 & 31.0 & 36.9 & \ccc{-1.9} \\
    \cmidrule(lr){2-15}
    \rowcolor{DrawioPurple} \cellcolor{white}
    & 4 &&
        79.4 & 61.7 & 23.8 & 29.4 & \textbf{48.7} & 41.0 & 38.4 & \textbf{40.8} & 64.2 & 39.7 & 46.7 & \ccc{+7.9} \\
    \rowcolor{DrawioPurple} \cellcolor{white}
    & 8 &&
        81.1 & 52.2 & 32.9 & \textbf{29.8} & 40.6 & 43.9 & 39.3 & 34.8 & \textbf{66.5} & 53.9 & 47.5 & \ccc{+8.7} \\
    \rowcolor{DrawioPurple}
    \cellcolor{white} \ourr & 16 &&
        \textbf{82.7} & \textbf{71.8} & \textbf{36.7} & \textbf{29.8} & 41.7 & \textbf{53.8} & \textbf{42.5} & 35.9 & 65.3 & \textbf{57.4} & \textbf{51.8} & \ccc{+13.0} \\
    
    \bottomrule
  \end{tabular}}
  \label{supp:tab:ow-ss-llava}
\end{subtable}
\end{table*}

\begin{table*}

\newcommand\random{\multirow{3}{*}{\makecell{\textit{Random} \\ \textit{Ctx}}}}
\newcommand\pseudo{\multirow{3}{*}{\makecell{\textit{Pseudo} \\ \textit{ICL}}}}
\newcommand\ourr{\multirow{-3}{*}{\ours}}

\ContinuedFloat
\begin{subtable}{\textwidth}
\caption{\textbf{Phi-3.5-Vision and Phi-4-Multimodal.}}
  \centering
  \resizebox{\linewidth}{!}{
  \begin{tabular}{rc c cccccccccccc}
    \toprule
    Method & Shots && C101 & DTD & ESAT & FGVC & FLWR & FOOD & PETS & CARS & S397 & U101 & Avg. & $\Delta$ \\
    \midrule

    \rowcolor{DrawioYellow}
    \multicolumn{15}{c}{\emph{Phi-3.5-Vision}} \\ [-2ex] \\

    \textit{Zero-Shot} & - &&
        53.2 & 29.1 & 7.4 & 19.9 & 31.6 & 40.2 & 24.6 & 39.1 & 43.2 & 38.3 & 32.6 \\
    \cmidrule(lr){2-15}
    \random & 4 &&
        67.0 & 26.9 & 16.9 & 20.5 & 36.3 & 41.8 & 29.6 & 34.9 & 48.1 & 37.8 & 36.0 & \ccc{+3.4} \\
    & 8 &&
        62.1 & 24.7 & 24.1 & 22.9 & 36.6 & 37.5 & 30.1 & 33.5 & 44.8 & 32.6 & 34.9 & \ccc{+2.3} \\
    & 16 &&
        57.4 & 24.2 & 25.1 & 21.0 & 35.3 & 34.2 & 29.4 & 33.2 & 39.3 & 29.7 & 32.9 & \ccc{+2.3} \\
    \cmidrule(lr){2-15}
    \pseudo & 4 &&
        49.6 & 26.0 & 6.6 & 18.8 & 31.7 & 39.3 & 28.0 & 46.0 & 38.5 & 35.8 & 32.1 & \ccc{-0.5} \\
    & 8 &&
        50.6 & 27.3 & 5.5 & 19.0 & 32.8 & 40.3 & 23.7 & 46.6 & 42.9 & 34.0 & 32.3 & \ccc{-0.3} \\
    & 16 &&
        52.7 & 26.4 & 5.8 & 19.8 & 31.0 & 40.6 & 29.9 & \textbf{46.8} & 39.6 & 36.4 & 32.9 & \ccc{+0.3} \\
    \cmidrule(lr){2-15}
    \rowcolor{DrawioPurple} \cellcolor{white}
    & 4 &&
        \textbf{69.7} & {37.8} & \textbf{32.0} & 28.8 & \textbf{45.2} & 38.2 & 34.0 & 44.4 & 56.8 & \textbf{39.6} & 42.7 & \ccc{+10.1} \\
    \rowcolor{DrawioPurple} \cellcolor{white}
    & 8 &&
        72.4 & \textbf{37.9} & 30.0 & 25.5 & 42.8 & \textbf{49.0} & \textbf{34.4} & 38.5 & \textbf{56.9} & 26.9 & 41.4 & \ccc{+8.8} \\
    \rowcolor{DrawioPurple}
    \cellcolor{white} \ourr & 16 &&
        67.9 & 32.6 & 31.0 & \textbf{29.8} & 42.2 & 44.0 & 31.5 & 36.3 & 51.4 & 26.6 & \textbf{39.3} & \ccc{+6.7} \\

    \cmidrule{1-15}
    \rowcolor{DrawioYellow}
    \multicolumn{15}{c}{\emph{Phi-4-Multimodal}} \\ [-2ex] \\

    \textit{Zero-Shot} & - &&
        73.5 & 33.9 & 13.8 & 29.2 & 42.2 & 37.8 & 37.5 & 34.1 & 41.2 & 40.0 & 38.3 \\
    \cmidrule(lr){2-15}
    \random & 4 &&
        36.7 & 29.7 & 36.1 & 28.8 & 40.9 & 36.5 & 36.8 & 32.3 & 28.5 & 24.4 & 33.1 & \ccc{-5.2} \\
    & 8 &&
        39.5 & 27.8 & 36.9 & 29.1 & 41.4 & 36.4 & 37.6 & 32.3 & 28.9 & 24.2 & 33.4 & \ccc{-4.9} \\
    & 16 &&
        38.2 & 26.4 & 36.0 & 29.3 & 41.0 & 35.2 & 37.6 & 32.2 & 27.4 & 23.2 & 32.6 & \ccc{-5.7} \\
    \cmidrule(lr){2-15}
    \pseudo & 4 &&
        75.7 & 32.2 & 14.9 & 29.3 & 42.8 & 43.8 & 38.2 & 34.5 & 47.4 & 32.7 & 39.1 & \ccc{+0.8} \\
    & 8 &&
        74.4 & 36.9 & 25.1 & \textbf{29.4} & 42.0 & 43.9 & 38.2 & 33.6 & 45.4 & 34.0 & 40.3 & \ccc{+2.0} \\
    & 16 &&
        76.4 & 29.4 & 13.2 & \textbf{29.4} & 41.9 & 43.1 & 38.2 & 34.7 & 46.7 & 31.5 & 38.4 & \ccc{+0.1} \\
    \cmidrule(lr){2-15}
    \rowcolor{DrawioPurple} \cellcolor{white}
    & 4 &&
        \textbf{77.2} & 28.8 & 27.5 & 28.7 & 49.2 & 53.3 & 46.7 & 40.5 & 59.7 & \textbf{51.5} & 46.3 & \ccc{+8.2} \\
    \rowcolor{DrawioPurple} \cellcolor{white}
    & 8 &&
        71.9 & 38.3 & 36.8 & 28.4 & \textbf{54.6} & \textbf{62.1} & 54.3 & 48.1 & \textbf{59.9} & 37.0 & 49.1 & \ccc{+10.8} \\
    \rowcolor{DrawioPurple}
    \cellcolor{white} \ourr & 16 &&
        72.8 & \textbf{39.6} & \textbf{42.1} & 28.4 & 47.8 & 57.3 & \textbf{54.9} & \textbf{51.9} & 58.3 & 47.9 & \textbf{50.1} & \ccc{+11.8} \\
    
    \bottomrule
  \end{tabular}}
  \label{supp:tab:ow-ss-phi}
\end{subtable}
\end{table*}

\begin{table*}

\newcommand\random{\multirow{3}{*}{\makecell{\textit{Random} \\ \textit{Ctx}}}}
\newcommand\pseudo{\multirow{3}{*}{\makecell{\textit{Pseudo} \\ \textit{ICL}}}}
\newcommand\ourr{\multirow{-3}{*}{\ours}}

\caption{
\textbf{Open-world results. \textit{Concept Similarity}} (\texttt{bCS}) on the ten datasets.
\inlineColorbox{DrawioPurple}{Purple} indicates our \ours.
Higher is better.
For each LMM, \textbf{bold} indicates the best result.
For each LMM, \textbf{bold} indicates the best result.
$\Delta$ computed \wrt the average (Avg.) of \textit{Zero-Shot}.
Results are split in \cref{supp:tab:ow-bcs-qwen,supp:tab:ow-bcs-llava,supp:tab:ow-bcs-phi} for readability.
}
\begin{subtable}{\textwidth}
\caption{\textbf{Qwen2-VL and Qwen2.5-VL.}}
  \centering
  \resizebox{\linewidth}{!}{
  \begin{tabular}{rc c cccccccccccc}
    \toprule
    Method & Shots && C101 & DTD & ESAT & FGVC & FLWR & FOOD & PETS & CARS & S397 & U101 & Avg. & $\Delta$ \\
    \midrule

    \rowcolor{DrawioYellow}
    \multicolumn{15}{c}{\emph{Qwen2-VL 7B}} \\ [-2ex] \\

    \textit{Zero-Shot} & - &&
         \textbf{81.3} & 50.3 & 39.8 & 30.7 & \textbf{68.7} & 77.0 & 43.2 & 55.7 & 70.7 & 59.0 & 57.6 \\
    \cmidrule(lr){2-15}
    \random & 4 &&
        75.9 & 42.7 & 52.0 & 30.7 & 49.4 & 68.7 & 49.7 & 61.9 & 59.4 & 53.5 & 54.4 & \ccc{-3.2} \\
    & 8 &&
        75.4 & 38.3 & 53.7 & 29.6 & 46.1 & 64.2 & 38.7 & 48.8 & 56.5 & 45.2 & 49.7 & \ccc{-7.9} \\
    & 16 &&
        63.3 & 33.7 & \textbf{54.2} & 29.4 & 41.0 & 57.6 & 35.9 & 38.8 & 42.0 & 35.9 & 43.2 & \ccc{-14.4} \\
    \cmidrule(lr){2-15}
    \pseudo & 4 &&
        79.9 & \textbf{51.1} & 37.2 & 31.3 & 63.7 & 78.4 & 51.0 & \textbf{67.0} & \textbf{72.4} & \textbf{62.1} & 59.4 & \ccc{+1.8} \\
    & 8 &&
        80.2 & 44.9 & 53.2 & 30.6 & 56.8 & 73.2 & 46.8 & 62.1 & 70.1 & 58.3 & 57.6 & \ccc{+0.0} \\
    & 16 &&
        80.2 & 48.5 & 34.0 & 30.2 & 58.4 & 72.5 & 47.4 & 61.6 & 67.7 & 61.1 & 56.2 & \ccc{-1.4} \\
    \cmidrule(lr){2-15}
    \rowcolor{DrawioPurple} \cellcolor{white}
    & 4 &&
        79.6 & 42.6 & 35.3 & 38.1 & 54.3 & \textbf{80.4} & 64.7 & 51.8 & 67.8 & \textbf{62.1} & 57.7 & \ccc{+0.1} \\
    \rowcolor{DrawioPurple} \cellcolor{white}
    & 8 &&
        76.5 & 44.8 & 37.4 & 35.6 & 64.8 & 74.6 & 69.9 & 50.4 & 67.6 & 53.8 & 57.5 & \ccc{-0.1} \\
    \rowcolor{DrawioPurple}
    \cellcolor{white} \ourr & 16 &&
        79.0 & 46.2 & 40.2 & \textbf{42.2} & 64.4 & 77.5 & \textbf{73.9} & 58.1 & 69.5 & 61.8 & \textbf{61.3} & \ccc{+3.7} \\

    \cmidrule{1-15}
    \rowcolor{DrawioYellow}
    \multicolumn{15}{c}{\emph{Qwen2.5-VL 7B}} \\ [-2ex] \\

    \textit{Zero-Shot} & - &&
        \textbf{85.6} & 53.4 & 41.3 & \textbf{68.7} & \textbf{79.7} & \textbf{79.6} & 77.3 & 68.5 & \textbf{74.2} & \textbf{67.2} & 69.5 \\
    \cmidrule(lr){2-15}
    \random & 4 &&
        82.9 & 52.6 & 54.8 & 60.0 & 77.0 & 72.5 & 81.3 & 76.9 & 72.6 & 65.2 & 69.6 & \ccc{+0.1} \\
    & 8 &&
        82.9 & 52.7 & 56.7 & 60.6 & 76.4 & 72.6 & \textbf{81.4} & 79.0 & 73.0 & 65.1 & 70.0 & \ccc{+0.5} \\
    & 16 &&
        83.5 & 52.9 & \textbf{59.0} & 60.3 & 76.6 & 72.6 & 80.8 & \textbf{80.1} & 73.0 & 64.7 & \textbf{70.3} & \ccc{+0.8} \\
    \cmidrule(lr){2-15}
    \pseudo & 4 &&
        85.2 & 53.3 & 41.3 & 57.1 & 77.2 & 72.9 & 74.3 & 69.6 & 72.1 & 65.1 & 66.8 & \ccc{-2.7} \\
    & 8 &&
        85.0 & 53.3 & 41.0 & 60.0 & 78.6 & 73.4 & 73.7 & 69.9 & 72.7 & 65.4 & 67.3 & \ccc{-2.2} \\
    & 16 &&
        85.0 & \textbf{54.1} & 42.1 & 59.9 & 77.1 & 74.6 & 74.7 & 71.1 & 73.0 & 65.0 & 67.7 & \ccc{-1.8} \\
    \cmidrule(lr){2-15}
    \rowcolor{DrawioPurple} \cellcolor{white}
    & 4 &&
        71.1 & 47.6 & 30.9 & 29.6 & 64.5 & 60.6 & 64.5 & 49.5 & 68.2 & 53.8 & 54.0 & \ccc{-15.5} \\
    \rowcolor{DrawioPurple} \cellcolor{white}
    & 8 &&
        75.7 & 43.7 & 34.4 & 29.1 & 67.7 & 60.3 & 60.0 & 46.7 & 66.7 & 40.6 & 52.5 & \ccc{-17.0} \\
    \rowcolor{DrawioPurple}
    \cellcolor{white} \ourr & 16 &&
        71.1 & 45.3 & 35.6 & 29.9 & 66.6 & 57.8 & 58.5 & 43.3 & 65.1 & 54.4 & 52.8 & \ccc{-16.7} \\
    
    \bottomrule
  \end{tabular}}
  \label{supp:tab:ow-bcs-qwen}
\end{subtable}
\end{table*}

\begin{table*}

\newcommand\random{\multirow{3}{*}{\makecell{\textit{Random} \\ \textit{Ctx}}}}
\newcommand\pseudo{\multirow{3}{*}{\makecell{\textit{Pseudo} \\ \textit{ICL}}}}
\newcommand\ourr{\multirow{-3}{*}{\ours}}

\ContinuedFloat
\begin{subtable}{\textwidth}
\caption{\textbf{LLaVa OneVision.}}
  \centering
  \resizebox{\linewidth}{!}{
  \begin{tabular}{rc c cccccccccccc}
    \toprule
    Method & Shots && C101 & DTD & ESAT & FGVC & FLWR & FOOD & PETS & CARS & S397 & U101 & Avg. & $\Delta$ \\
    \midrule

    \rowcolor{DrawioYellow}
    \multicolumn{15}{c}{\emph{LLaVa OneVision 7B}} \\ [-2ex] \\

    \textit{Zero-Shot} & - &&
        79.1 & 46.9 & 41.0 & 29.4 & 51.9 & 41.9 & 37.9 & 35.3 & 44.8 & 43.5 & 45.2 \\
    \cmidrule(lr){2-15}
    \random & 4 &&
        44.0 & 30.3 & 50.4 & 29.4 & 32.7 & 33.0 & 37.6 & 32.2 & 26.3 & 26.3 & 34.2 & \ccc{-11.0} \\
    & 8 &&
        44.9 & 30.9 & \textbf{56.4} & 29.4 & 32.7 & 33.3 & 37.7 & 32.2 & 26.1 & 26.4 & 35.0 & \ccc{-10.2} \\
    & 16 &&
        46.2 & 36.4 & 55.2 & 29.4 & 35.5 & 34.0 & 37.8 & 32.2 & 27.2 & 26.3 & 36.0 & \ccc{-9.2} \\
    \cmidrule(lr){2-15}
    \pseudo & 4 &&
        33.0 & 35.0 & 38.6 & 29.4 & 35.1 & 49.7 & 35.3 & 32.9 & 49.2 & 20.9 & 35.9 & \ccc{-9.3} \\
    & 8 &&
        \textbf{84.3} & 53.7 & 54.6 & 29.4 & \textbf{58.2} & 40.5 & 38.4 & 33.8 & 43.5 & 28.5 & 46.5 & \ccc{+1.3} \\
    & 16 &&
        84.1 & 53.5 & 53.9 & 29.4 & 57.0 & 41.6 & 38.4 & 35.2 & 46.6 & 31.5 & 47.1 & \ccc{+1.9} \\
    \cmidrule(lr){2-15}
    \rowcolor{DrawioPurple} \cellcolor{white}
    & 4 &&
        79.4 & 61.7 & 23.8 & 29.4 & 48.7 & 41.0 & 38.4 & \textbf{41.3} & 64.2 & 39.8 & 46.8 & \ccc{+1.6} \\
    \rowcolor{DrawioPurple} \cellcolor{white}
    & 8 &&
        81.1 & 52.3 & 32.9 & \textbf{29.8} & 40.6 & 43.9 & 39.3 & 34.8 & \textbf{66.5} & 53.9 & 47.5 & \ccc{+2.3} \\
    \rowcolor{DrawioPurple}
    \cellcolor{white} \ourr & 16 &&
        82.7 & \textbf{71.8} & 36.7 & \textbf{29.8} & 41.7 & \textbf{53.9} & \textbf{42.5} & 35.9 & 65.3 & \textbf{57.4} & \textbf{51.8} & \ccc{+6.6} \\
    
    \bottomrule
  \end{tabular}}
  \label{supp:tab:ow-bcs-llava}
\end{subtable}
\end{table*}

\begin{table*}

\newcommand\random{\multirow{3}{*}{\makecell{\textit{Random} \\ \textit{Ctx}}}}
\newcommand\pseudo{\multirow{3}{*}{\makecell{\textit{Pseudo} \\ \textit{ICL}}}}
\newcommand\ourr{\multirow{-3}{*}{\ours}}

\ContinuedFloat 
\begin{subtable}{\textwidth}
\caption{\textbf{Phi-3.5-Vision and Phi-4-Multimodal.}}
  \centering
  \resizebox{\linewidth}{!}{
  \begin{tabular}{rc c cccccccccccc}
    \toprule
    Method & Shots && C101 & DTD & ESAT & FGVC & FLWR & FOOD & PETS & CARS & S397 & U101 & Avg. & $\Delta$ \\
    \midrule

    \rowcolor{DrawioYellow}
    \multicolumn{15}{c}{\emph{Phi-3.5-Vision}} \\ [-2ex] \\

    \textit{Zero-Shot} & - &&
        \textbf{73.6} & \textbf{43.5} & 16.4 & 29.6 & 44.9 & 58.1 & 38.7 & 43.0 & 57.6 & \textbf{50.1} & 45.5 \\
    \cmidrule(lr){2-15}
    \random & 4 &&
        71.1 & 34.1 & 29.2 & 29.3 & 42.5 & 50.2 & 36.7 & 35.3 & 50.6 & 42.7 & 42.2 & \ccc{-3.3} \\
    & 8 &&
        67.4 & 32.0 & \textbf{41.6} & 29.3 & 41.2 & 46.3 & 36.9 & 34.1 & 47.7 & 38.9 & 41.5 & \ccc{-4.0} \\
    & 16 &&
        64.7 & 31.9 & 40.1 & 29.3 & 41.0 & 42.7 & 36.6 & 34.1 & 44.2 & 37.2 & 40.2 & \ccc{-5.3} \\
    \cmidrule(lr){2-15}
    \pseudo & 4 &&
        69.3 & 40.9 & 16.1 & 29.2 & 41.9 & 58.9 & 45.1 & \textbf{54.4} & 54.2 & 48.3 & 45.8 & \ccc{+0.3} \\
    & 8 &&
        70.2 & 41.7 & 14.6 & 28.2 & 45.3 & 59.4 & 45.3 & 53.8 & \textbf{60.4} & 47.4 & 46.6 & \ccc{+1.1} \\
    & 16 &&
        71.5 & 40.8 & 14.9 & 25.8 & 42.9 & \textbf{60.4} & \textbf{52.5} & 54.2 & 55.9 & 49.1 & \textbf{46.8} & \ccc{+1.3} \\
    \cmidrule(lr){2-15}
    \rowcolor{DrawioPurple} \cellcolor{white}
    & 4 &&
        71.2 & 40.2 & 32.0 & \textbf{31.4} & \textbf{53.8} & 40.5 & 40.3 & 44.8 & 58.0 & 40.5 & 45.3 & \ccc{-0.2} \\
    \rowcolor{DrawioPurple} \cellcolor{white}
    & 8 &&
        \textbf{73.6} & 42.6 & 30.2 & 30.8 & 52.0 & 52.6 & 42.1 & 43.8 & 58.8 & 35.3 & 46.2 & \ccc{+0.7} \\
    \rowcolor{DrawioPurple}
    \cellcolor{white} \ourr & 16 &&
        71.0 & 39.0 & 32.6 & {30.9} & 50.5 & 48.2 & 38.4 & 36.3 & 55.3 & 34.1 & 43.6 & \ccc{-1.9} \\

    \cmidrule{1-15}
    \rowcolor{DrawioYellow}
    \multicolumn{15}{c}{\emph{Phi-4-Multimodal}} \\ [-2ex] \\

    \textit{Zero-Shot} & - &&
        75.8 & 40.6 & 17.4 & \textbf{29.3} & 42.2 & 37.9 & 37.5 & 34.1 & 41.7 & 40.1 & 39.6 \\
    \cmidrule(lr){2-15}
    \random & 4 &&
        36.8 & 29.7 & 36.3 & 28.8 & 40.9 & 36.5 & 36.8 & 32.3 & 28.5 & 24.8 & 33.1 & \ccc{-6.5} \\
    & 8 &&
        39.5 & 27.8 & 37.2 & 29.1 & 41.4 & 36.4 & 37.6 & 32.3 & 28.9 & 24.2 & 33.4 & \ccc{-6.2} \\
    & 16 &&
        38.2 & 26.4 & 36.2 & 29.2 & 41.0 & 35.2 & 37.6 & 32.2 & 27.4 & 23.2 & 32.7 & \ccc{-6.9} \\
    \cmidrule(lr){2-15}
    \pseudo & 4 &&
        76.7 & 38.5 & 22.4 & \textbf{29.3} & 42.9 & 43.9 & 38.2 & 34.5 & 47.4 & 32.7 & 40.6 & \ccc{+1.0} \\
    & 8 &&
        74.5 & 37.0 & 28.9 & \textbf{29.3} & 42.0 & 43.9 & 38.2 & 33.6 & 45.4 & 34.0 & 40.7 & \ccc{+1.1} \\
    & 16 &&
        76.4 & 41.0 & 21.0 & \textbf{29.3} & 41.9 & 43.1 & 38.2 & 34.7 & 46.7 & 31.5 & 40.4 & \ccc{+0.8} \\
    \cmidrule(lr){2-15}
    \rowcolor{DrawioPurple} \cellcolor{white}
    & 4 &&
        77.8 & 29.3 & 27.5 & \textbf{29.3} & 52.7 & 59.0 & 47.9 & 43.3 & \textbf{61.1} & 53.1 & 48.1 & \ccc{+8.5} \\
    \rowcolor{DrawioPurple} \cellcolor{white}
    & 8 &&
        76.7 & 42.0 & 37.7 & 28.4 & \textbf{57.6} & \textbf{65.3} & \textbf{57.9} & 49.4 & {60.2} & 51.9 & 52.7 & \ccc{+13.1} \\
    \rowcolor{DrawioPurple}
    \cellcolor{white} \ourr & 16 &&
        \textbf{81.0} & \textbf{42.3} & \textbf{42.2} & 29.1 & 50.3 & 59.4 & 56.8 & \textbf{55.9} & 59.2 & \textbf{53.7} & \textbf{53.0} & \ccc{+13.4} \\
    
    \bottomrule
  \end{tabular}}
  \label{supp:tab:ow-bcs-phi}
\end{subtable}
\end{table*}

\begin{table*}

\newcommand\random{\multirow{3}{*}{\makecell{\textit{Random} \\ \textit{Ctx}}}}
\newcommand\pseudo{\multirow{3}{*}{\makecell{\textit{Pseudo} \\ \textit{ICL}}}}
\newcommand\ourr{\multirow{-3}{*}{\ours}}

\caption{
\textbf{Open-world results. \textit{Median Concept Similarity}} (\texttt{mCS}) on the ten datasets.
\inlineColorbox{DrawioPurple}{Purple} indicates our \ours.
Higher is better.
For each LMM, \textbf{bold} indicates the best result.
For each LMM, \textbf{bold} indicates the best result.
$\Delta$ computed \wrt the average (Avg.) of \textit{Zero-Shot}.
Results are split in \cref{supp:tab:ow-mcs-qwen,supp:tab:ow-mcs-llava,supp:tab:ow-mcs-phi} for readability.
}
\begin{subtable}{\textwidth}
\caption{\textbf{Qwen2-VL and Qwen2.5-VL.}}
  \centering
  \resizebox{\linewidth}{!}{
  \begin{tabular}{rc c cccccccccccc}
    \toprule
    Method & Shots && C101 & DTD & ESAT & FGVC & FLWR & FOOD & PETS & CARS & S397 & U101 & Avg. & $\Delta$ \\
    \midrule

    \rowcolor{DrawioYellow}
    \multicolumn{15}{c}{\emph{Qwen2-VL 7B}} \\ [-2ex] \\

    \textit{Zero-Shot} & - &&
        51.3 & 27.9 & 20.6 & 20.4 & 33.3 & 37.3 & 25.0 & 45.8 & 36.1 & 38.9 & 33.7 \\
    \cmidrule(lr){2-15}
    \random & 4 &&
        46.3 & 26.0 & 24.0 & 20.8 & 34.9 & 45.7 & 31.3 & 48.6 & 32.7 & 37.1 & 34.7 & \ccc{+1.0} \\
    & 8 &&
        49.0 & 23.5 & 23.3 & 20.3 & 30.7 & 47.6 & 28.9 & 39.0 & 34.8 & 33.0 & 33.0 & \ccc{-0.7} \\
    & 16 &&
        48.6 & 20.8 & 23.1 & 22.8 & 26.8 & 46.8 & 30.0 & 32.9 & 30.8 & 25.2 & 30.8 & \ccc{-2.9} \\
    \cmidrule(lr){2-15}
    \pseudo & 4 &&
        56.1 & 27.9 & 18.3 & 20.7 & 30.1 & 35.0 & 27.1 & 56.5 & 32.8 & 34.4 & 33.9 & \ccc{+0.2} \\
    & 8 &&
        55.2 & 27.7 & 26.3 & 20.3 & 32.2 & 39.9 & 25.1 & 52.3 & 34.1 & 41.9 & 35.5 & \ccc{+1.8} \\
    & 16 &&
        57.5 & 28.2 & 16.0 & 20.1 & 34.8 & 38.7 & 26.5 & 51.3 & 35.2 & 37.7 & 34.6 & \ccc{+0.9} \\
    \cmidrule(lr){2-15}
    \rowcolor{DrawioPurple} \cellcolor{white}
    & 4 &&
        \textbf{66.6} & 29.0 & 34.6 & 26.2 & 52.2 & 62.9 & 41.3 & 45.5 & 54.5 & 48.8 & 46.1 & \ccc{+12.4} \\
    \rowcolor{DrawioPurple} \cellcolor{white}
    & 8 &&
        58.3 & 34.3 & 27.4 & 25.7 & \textbf{57.3} & 62.9 & 41.8 & 45.1 & 61.0 & 48.2 & 46.2 & \ccc{+12.5} \\
    \rowcolor{DrawioPurple}
    \cellcolor{white} \ourr & 16 &&
        65.2 & \textbf{34.8} & \textbf{35.9} & \textbf{30.2} & 53.5 & \textbf{67.2} & \textbf{51.2} & \textbf{49.4} & \textbf{61.8} & \textbf{50.9} & \textbf{50.0} & \ccc{+16.3} \\

    \cmidrule{1-15}
    \rowcolor{DrawioYellow}
    \multicolumn{15}{c}{\emph{Qwen2.5-VL 7B}} \\ [-2ex] \\

    \textit{Zero-Shot} & - &&
        33.0 & 26.3 & 18.6 & 24.7 & 29.5 & 31.8 & 27.3 & 29.4 & 29.3 & 29.6 & 27.9 \\
    \cmidrule(lr){2-15}
    \random & 4 &&
        51.4 & 26.7 & 21.4 & 26.1 & 33.3 & 31.1 & 38.9 & 50.1 & 31.9 & 31.6 & 34.3 & \ccc{+6.4} \\
    & 8 &&
        53.1 & 26.9 & 21.9 & 27.2 & 37.4 & 31.0 & 40.1 & 54.3 & 32.4 & 31.6 & 35.6 & \ccc{+7.7} \\
    & 16 &&
        54.0 & 26.8 & 23.2 & 28.6 & 39.3 & 31.2 & 38.5 & 55.4 & 32.2 & 32.1 & 36.1 & \ccc{+8.2} \\
    \cmidrule(lr){2-15}
    \pseudo & 4 &&
        32.6 & 26.7 & 18.9 & 24.3 & 29.4 & 31.1 & 24.8 & 26.8 & 29.5 & 26.9 & 27.1 & \ccc{-0.8} \\
    & 8 &&
        32.1 & 26.7 & 18.9 & 24.1 & 29.1 & 30.6 & 23.8 & 23.9 & 29.1 & 27.7 & 26.6 & \ccc{-1.3} \\
    & 16 &&
        32.5 & 27.1 & 20.3 & 24.7 & 28.8 & 31.1 & 25.6 & 25.4 & 29.3 & 26.2 & 27.1 & \ccc{-0.8} \\
    \cmidrule(lr){2-15}
    \rowcolor{DrawioPurple} \cellcolor{white}
    & 4 &&
        66.0 & 37.9 & 30.9 & \textbf{29.6} & 64.2 & \textbf{57.9} & 55.6 & \textbf{49.5} & \textbf{67.9} & 49.9 & 50.9 & \ccc{+23.0} \\
    \rowcolor{DrawioPurple} \cellcolor{white}
    & 8 &&
        \textbf{72.4} & \textbf{42.5} & \textbf{34.4} & 29.0 & 66.4 & 57.5 & \textbf{58.9} & 46.7 & 66.3 & 40.4 & \textbf{51.5} & \ccc{+23.6} \\
    \rowcolor{DrawioPurple}
    \cellcolor{white} \ourr & 16 &&
        69.5 & 41.8 & \textbf{34.4} & \textbf{29.6} & \textbf{66.0} & 56.3 & 56.8 & 43.3 & 65.0 & \textbf{50.5} & 51.3 & \ccc{+23.4} \\
    
    \bottomrule
  \end{tabular}}
  \label{supp:tab:ow-mcs-qwen}
\end{subtable}
\end{table*}

\begin{table*}

\newcommand\random{\multirow{3}{*}{\makecell{\textit{Random} \\ \textit{Ctx}}}}
\newcommand\pseudo{\multirow{3}{*}{\makecell{\textit{Pseudo} \\ \textit{ICL}}}}
\newcommand\ourr{\multirow{-3}{*}{\ours}}

\ContinuedFloat
\begin{subtable}{\textwidth}
\caption{\textbf{LLaVa OneVision.}}
  \centering
  \resizebox{\linewidth}{!}{
  \begin{tabular}{rc c cccccccccccc}
    \toprule
    Method & Shots && C101 & DTD & ESAT & FGVC & FLWR & FOOD & PETS & CARS & S397 & U101 & Avg. & $\Delta$ \\
    \midrule

    \rowcolor{DrawioYellow}
    \multicolumn{15}{c}{\emph{LLaVa OneVision 7B}} \\ [-2ex] \\

    \textit{Zero-Shot} & - &&
        64.1 & 30.0 & 17.6 & 29.4 & 32.5 & 41.4 & 37.8 & 32.5 & 42.6 & 43.1 & 37.1 \\
    \cmidrule(lr){2-15}
    \random & 4 &&
        41.6 & 25.6 & 19.3 & 29.4 & 32.1 & 33.0 & 37.6 & 32.2 & 26.3 & 26.3 & 30.3 & \ccc{-6.8} \\
    & 8 &&
        40.4 & 25.2 & 20.4 & 29.4 & 32.0 & 33.3 & 37.7 & 32.2 & 26.1 & 26.4 & 30.3 & \ccc{-6.8} \\
    & 16 &&
        32.1 & 25.5 & 20.3 & 29.4 & 28.3 & 34.0 & 37.8 & 32.2 & 27.2 & 26.3 & 29.3 & \ccc{-7.8} \\
    \cmidrule(lr){2-15}
    \pseudo & 4 &&
        23.3 & 22.5 & 16.3 & 29.4 & 27.1 & 41.9 & 35.1 & 30.2 & 37.8 & 20.3 & 28.4 & \ccc{-8.7} \\
    & 8 &&
        58.1 & 27.6 & 16.9 & 29.4 & 25.6 & 38.7 & 38.2 & 32.1 & 39.6 & 27.5 & 33.4 & \ccc{-3.7} \\
    & 16 &&
        57.3 & 27.0 & 17.9 & 29.4 & 25.3 & 39.8 & 38.2 & 31.6 & 41.4 & 30.6 & 33.8 & \ccc{-3.3} \\
    \cmidrule(lr){2-15}
    \rowcolor{DrawioPurple} \cellcolor{white}
    & 4 &&
        79.4 & 61.7 & 23.8 & 29.4 & \textbf{48.4} & 41.0 & 38.4 & \textbf{39.8} & 64.2 & 39.8 & 46.6 & \ccc{+9.5} \\
    \rowcolor{DrawioPurple} \cellcolor{white}
    & 8 &&
        81.1 & 52.3 & 32.9 & \textbf{29.8} & 40.6 & 43.9 & 39.3 & 34.8 & \textbf{66.5} & 53.8 & 47.5 & \ccc{+10.4} \\
    \rowcolor{DrawioPurple}
    \cellcolor{white} \ourr & 16 &&
        \textbf{82.7} & \textbf{71.8} & \textbf{36.7} & \textbf{29.8} & 41.7 & \textbf{53.8} & \textbf{42.5} & 35.9 & 65.3 & \textbf{57.3} & \textbf{51.8} & \ccc{+14.7} \\
    
    \bottomrule
  \end{tabular}}
  \label{supp:tab:ow-mcs-llava}
\end{subtable}
\end{table*}

\begin{table*}

\newcommand\random{\multirow{3}{*}{\makecell{\textit{Random} \\ \textit{Ctx}}}}
\newcommand\pseudo{\multirow{3}{*}{\makecell{\textit{Pseudo} \\ \textit{ICL}}}}
\newcommand\ourr{\multirow{-3}{*}{\ours}}

\ContinuedFloat 
\begin{subtable}{\textwidth}
\caption{\textbf{Phi-3.5-Vision and Phi-4-Multimodal.}}
  \centering
  \resizebox{\linewidth}{!}{
  \begin{tabular}{rc c cccccccccccc}
    \toprule
    Method & Shots && C101 & DTD & ESAT & FGVC & FLWR & FOOD & PETS & CARS & S397 & U101 & Avg. & $\Delta$ \\
    \midrule

    \rowcolor{DrawioYellow}
    \multicolumn{15}{c}{\emph{Phi-3.5-Vision}} \\ [-2ex] \\

    \textit{Zero-Shot} & - &&
        53.1 & 28.8 & 7.4 & 21.2 & 32.3 & 36.6 & 25.0 & 38.3 & 39.2 & 36.1 & 31.8 \\
    \cmidrule(lr){2-15}
    \random & 4 &&
        67.0 & 27.5 & 16.6 & 20.5 & 36.3 & 41.3 & 29.4 & 32.4 & 47.6 & 37.3 & 35.6 & \ccc{+3.8} \\
    & 8 &&
        62.0 & 25.3 & 24.3 & 22.9 & 36.6 & 37.6 & 29.8 & 31.7 & 44.5 & 32.8 & 34.8 & \ccc{+3.0} \\
    & 16 &&
        57.6 & 24.9 & 26.5 & 21.0 & 35.3 & 35.3 & 29.2 & 31.6 & 39.2 & 30.9 & 33.1 & \ccc{+1.3} \\
    \cmidrule(lr){2-15}
    \pseudo & 4 &&
        47.4 & 25.3 & 6.7 & 20.8 & 27.3 & 32.7 & 29.2 & 44.3 & 32.8 & 32.4 & 29.9 & \ccc{-1.9} \\
    & 8 &&
        49.2 & 26.1 & 5.5 & 21.1 & 27.0 & 32.9 & 24.7 & 46.8 & 32.9 & 32.3 & 29.9 & \ccc{-1.9} \\
    & 16 &&
        52.1 & 25.5 & 5.8 & 20.4 & 27.0 & 32.9 & 28.7 & \textbf{46.9} & 34.0 & 32.9 & 30.6 & \ccc{-0.2} \\
    \cmidrule(lr){2-15}
    \rowcolor{DrawioPurple} \cellcolor{white}
    & 4 &&
        69.5 & 38.3 & \textbf{32.0} & 29.0 & \textbf{44.1} & 37.9 & 35.8 & 43.2 & 56.7 & \textbf{39.8} & \textbf{42.6} & \ccc{+10.8} \\
    \rowcolor{DrawioPurple} \cellcolor{white}
    & 8 &&
        \textbf{72.7} & \textbf{39.4} & 30.0 & 28.4 & 42.5 & \textbf{49.4} & 35.9 & 36.3 & \textbf{57.2} & 29.3 & 42.1 & \ccc{+10.3} \\
    \rowcolor{DrawioPurple}
    \cellcolor{white} \ourr & 16 &&
        68.3 & 34.7 & 30.8 & \textbf{30.0} & 41.8 & 46.3 & \textbf{38.2} & 36.3 & 52.3 & 29.5 & 40.8 & \ccc{+9.0} \\

    \cmidrule{1-15}
    \rowcolor{DrawioYellow}
    \multicolumn{15}{c}{\emph{Phi-4-Multimodal}} \\ [-2ex] \\

    \textit{Zero-Shot} & - &&
        73.3 & 34.4 & 13.6 & 29.2 & 42.2 & 37.8 & 37.5 & 34.1 & 41.2 & 39.7 & 38.3 \\
    \cmidrule(lr){2-15}
    \random & 4 &&
        36.7 & 29.7 & 36.1 & 28.8 & 40.9 & 36.5 & 36.8 & 32.3 & 28.5 & 24.5 & 33.1 & \ccc{-5.2} \\
    & 8 &&
        39.5 & 27.8 & 36.9 & 29.1 & 41.4 & 36.4 & 37.6 & 32.3 & 28.9 & 24.1 & 33.4 & \ccc{-4.9} \\
    & 16 &&
        38.2 & 26.4 & 36.0 & 29.2 & 41.0 & 35.2 & 37.6 & 32.2 & 27.4 & 23.1 & 32.6 & \ccc{-5.7} \\
    \cmidrule(lr){2-15}
    \pseudo & 4 &&
        75.5 & 32.8 & 14.9 & \textbf{29.3} & 42.7 & 43.8 & 38.2 & 34.5 & 47.4 & 32.5 & 39.2 & \ccc{+0.9} \\
    & 8 &&
        74.4 & 36.9 & 25.1 & \textbf{29.3} & 42.0 & 43.8 & 38.2 & 33.6 & 45.4 & 34.0 & 40.3 & \ccc{+2.0} \\
    & 16 &&
        76.4 & 28.9 & 13.2 & \textbf{29.3} & 41.9 & 43.0 & 38.2 & 34.7 & 46.7 & 31.5 & 38.4 & \ccc{+0.1} \\
    \cmidrule(lr){2-15}
    \rowcolor{DrawioPurple} \cellcolor{white}
    & 4 &&
        \textbf{77.6} & 28.8 & 27.5 & 28.7 & 49.9 & 50.9 & 46.9 & 40.1 & 59.6 & \textbf{50.6} & 46.1 & \ccc{+7.8} \\
    \rowcolor{DrawioPurple} \cellcolor{white}
    & 8 &&
        71.7 & 38.7 & 36.6 & 28.4 & \textbf{53.8} & \textbf{61.7} & {54.7} & 47.0 & \textbf{59.8} & 28.7 & 48.1 & \ccc{+9.8} \\
    \rowcolor{DrawioPurple}
    \cellcolor{white} \ourr & 16 &&
        74.1 & \textbf{40.2} & \textbf{41.7} & 27.8 & 46.7 & 57.3 & \textbf{55.0} & \textbf{48.0} & 58.7 & 48.2 & \textbf{49.8} & \ccc{+11.5} \\
    
    \bottomrule
  \end{tabular}}
  \label{supp:tab:ow-mcs-phi}
\end{subtable}
\end{table*}

\begin{table*}

\newcommand\random{\multirow{3}{*}{\makecell{\textit{Random} \\ \textit{Ctx}}}}
\newcommand\pseudo{\multirow{3}{*}{\makecell{\textit{Pseudo} \\ \textit{ICL}}}}
\newcommand\ourr{\multirow{-3}{*}{\ours}}

\caption{
\textbf{Open-world results. \textit{Llama Inclusion}} (\texttt{LI}) on the ten datasets.
\inlineColorbox{DrawioPurple}{Purple} indicates our \ours.
Higher is better.
For each LMM, \textbf{bold} indicates the best result.
$\Delta$ computed \wrt the average (Avg.) of \textit{Zero-Shot}.
For each LMM, \textbf{bold} indicates the best result.
Results are split in \cref{supp:tab:ow-li-qwen,supp:tab:ow-li-llava,supp:tab:ow-li-phi} for readability.
}
\begin{subtable}{\textwidth}
\caption{\textbf{Qwen2-VL and Qwen2.5-VL.}}
  \centering
  \resizebox{\linewidth}{!}{
  \begin{tabular}{rc c cccccccccccc}
    \toprule
    Method & Shots && C101 & DTD & ESAT & FGVC & FLWR & FOOD & PETS & CARS & S397 & U101 & Avg. & $\Delta$ \\
    \midrule

    \rowcolor{DrawioYellow}
    \multicolumn{15}{c}{\emph{Qwen2-VL 7B}} \\ [-2ex] \\

    \textit{Zero-Shot} & - &&
        84.0 & 59.5 & 17.7 & 55.5 & 68.9 & 74.3 & 46.0 & 63.5 & 72.2 & 47.7 & 58.9 \\
    \cmidrule(lr){2-15}
    \random & 4 &&
         64.8 & 36.5 & 19.5 & 47.1 & 38.9 & 45.4 & 44.3 & 34.9 & 31.3 & 33.9 & 39.7 & \ccc{-19.2} \\
    & 8 &&
        63.2 & 24.9 & 20.7 & 45.9 & 31.7 & 39.7 & 41.5 & 28.0 & 33.1 & 26.2 & 35.5 & \ccc{-23.4} \\
    & 16 &&
        31.1 & 17.8 & 17.6 & 31.1 & 59.7 & 25.3 & 10.0 & 31.0 & 17.8 & 15.9 & 25.7 & \ccc{-33.2} \\
    \cmidrule(lr){2-15}
    \pseudo & 4 &&
        85.8 & 60.5 & 9.9 & 53.6 & 52.3 & 70.9 & 36.2 & 44.6 & 76.4 & 58.1 & 54.8 & \ccc{-4.1} \\
    & 8 &&
        83.0 & 45.1 & 18.6 & 52.9 & 50.0 & 64.4 & 27.3 & 41.0 & 70.6 & 49.4 & 50.2 & \ccc{-8.7} \\
    & 16 &&
        82.6 & 47.6 & 6.4 & 55.5 & 52.1 & 62.5 & 29.7 & 44.1 & 64.8 & 51.3 & 49.7 & \ccc{-9.2} \\
    \cmidrule(lr){2-15}
    \rowcolor{DrawioPurple} \cellcolor{white}
    & 4 &&
        92.7 & \textbf{69.4} & \textbf{54.9} & 88.6 & \textbf{91.5} & 88.9 & \textbf{86.2} & 98.1 & 88.8 & 73.5 & \textbf{83.3} & \ccc{+24.4} \\
    \rowcolor{DrawioPurple} \cellcolor{white}
    & 8 &&
        93.5 & 68.3 & 32.2 & \textbf{92.4} & 91.0 & \textbf{93.8} & 75.0 & \textbf{98.8} & \textbf{90.6} & \textbf{85.9} & 82.2 & \ccc{+23.3} \\
    \rowcolor{DrawioPurple}
    \cellcolor{white} \ourr & 16 &&
        \textbf{94.0} & 66.5 & 39.1 & 84.8 & 90.7 & 91.0 & 80.3 & 98.1 & 89.0 & 79.2 & 81.3 & \ccc{+22.4} \\

    \cmidrule{1-15}
    \rowcolor{DrawioYellow}
    \multicolumn{15}{c}{\emph{Qwen2.5-VL 7B}} \\ [-2ex] \\

    \textit{Zero-Shot} & - &&
        84.3 & 58.9 & 12.5 & 68.8 & 74.7 & 76.1 & 70.7 & 69.3 & 81.6 & 66.3 & 66.3 \\
    \cmidrule(lr){2-15}
    \random & 4 &&
        85.4 & 60.3 & 22.0 & 39.5 & 75.7 & 64.9 & 76.3 & 31.7 & 78.5 & 63.3 & 59.7 & \ccc{-6.6} \\
    & 8 &&
        86.1 & 61.4 & 21.7 & 37.6 & 73.7 & 65.4 & \textbf{77.1} & 31.2 & 78.1 & 62.5 & 59.5 & \ccc{-6.8} \\
    & 16 &&
        87.3 & 61.5 & 21.3 & 34.6 & 74.4 & 61.3 & 75.9 & 34.8 & 77.6 & 61.7 & 59.1 & \ccc{-7.2} \\
    \cmidrule(lr){2-15}
    \pseudo & 4 &&
        84.3 & 55.3 & 10.7 & 38.1 & 65.9 & 61.0 & 64.4 & 41.3 & 76.9 & 61.9 & 56.0 & \ccc{-10.3} \\
    & 8 &&
        83.1 & 56.4 & 13.4 & 33.3 & 69.8 & 59.3 & 61.7 & 47.6 & 77.9 & 61.4 & 56.4 & \ccc{-9.9} \\
    & 16 &&
        83.0 & 56.3 & 11.3 & 30.5 & 69.3 & 61.1 & 66.1 & 41.3 & 78.4 & 61.2 & 55.9 & \ccc{-10.4} \\
    \cmidrule(lr){2-15}
    \rowcolor{DrawioPurple} \cellcolor{white}
    & 4 &&
        90.1 & 68.8 & 42.1 & 95.6 & 94.4 & 80.2 & 72.4 & 81.8 & 90.6 & 72.2 & 78.8 & \ccc{+12.5} \\
    \rowcolor{DrawioPurple} \cellcolor{white}
    & 8 &&
        89.3 & \textbf{77.4} & 40.1 & \textbf{96.3} & \textbf{95.4} & 78.6 & 67.4 & 78.8 & 90.0 & 58.0 & 77.1 & \ccc{+10.8} \\
    \rowcolor{DrawioPurple}
    \cellcolor{white} \ourr & 16 &&
        \textbf{95.9} & 71.7 & \textbf{50.2} & 93.7 & 93.9 & \textbf{90.8} & 74.1 & \textbf{93.5} & \textbf{93.9} & \textbf{81.1} & \textbf{83.9} & \ccc{+17.6} \\
    
    \bottomrule
  \end{tabular}}
  \label{supp:tab:ow-li-qwen}
\end{subtable}
\end{table*}

\begin{table*}

\newcommand\random{\multirow{3}{*}{\makecell{\textit{Random} \\ \textit{Ctx}}}}
\newcommand\pseudo{\multirow{3}{*}{\makecell{\textit{Pseudo} \\ \textit{ICL}}}}
\newcommand\ourr{\multirow{-3}{*}{\ours}}

\ContinuedFloat
\begin{subtable}{\textwidth}
\caption{\textbf{LLaVa OneVision.}}
  \centering
  \resizebox{\linewidth}{!}{
  \begin{tabular}{rc c cccccccccccc}
    \toprule
    Method & Shots && C101 & DTD & ESAT & FGVC & FLWR & FOOD & PETS & CARS & S397 & U101 & Avg. & $\Delta$ \\
    \midrule

    \rowcolor{DrawioYellow}
    \multicolumn{15}{c}{\emph{LLaVa OneVision 7B}} \\ [-2ex] \\

    \textit{Zero-Shot} & - &&
        81.3 & 45.6 & 11.8 & 68.9 & 48.9 & 22.0 & 50.2 & \textbf{84.4} & 25.0 & 27.0 & 46.5 \\
    \cmidrule(lr){2-15}
    \random & 4 &&
        24.3 & 6.6 & 9.9 & 67.9 & 18.0 & 4.5 & 40.0 & 80.4 & 3.6 & 5.3 & 26.0 & \ccc{-20.5} \\
    & 8 &&
        25.1 & 7.0 & 9.0 & 67.9 & 17.3 & 4.5 & 40.4 & 80.4 & 3.3 & 4.3 & 25.9 & \ccc{-20.6} \\
    & 16 &&
        21.8 & 16.3 & 7.8 & 0.0 & 6.5 & 2.6 & 0.0 & 0.0 & 0.7 & 0.3 & 5.6 & \ccc{-40.9} \\
    \cmidrule(lr){2-15}
    \pseudo & 4 &&
        3.1 & 2.8 & 5.1 & 68.9 & 7.7 & 24.4 & 28.9 & 71.8 & 36.3 & 2.0 & 25.1 & \ccc{-21.4} \\
    & 8 &&
        87.7 & 61.3 & 15.1 & 67.9 & 43.9 & 11.8 & 46.1 & 80.8 & 23.2 & 13.4 & 45.1 & \ccc{-1.4} \\
    & 16 &&
        87.9 & 61.2 & 18.0 & 67.9 & 42.1 & 12.7 & 46.1 & 78.9 & 28.3 & 16.3 & 45.9 & \ccc{-0.6} \\
    \cmidrule(lr){2-15}
    \rowcolor{DrawioPurple} \cellcolor{white}
    & 4 &&
        85.5 & 84.8 & 34.3 & 96.0 & 93.2 & 46.9 & 70.4 & 38.8 & 84.2 & 38.4 & 67.2 & \ccc{+20.7} \\
    \rowcolor{DrawioPurple} \cellcolor{white}
    & 8 &&
        88.0 & 66.6 & 22.3 & \textbf{96.6} & 93.1 & 58.1 & 70.4 & 13.2 & 86.8 & 76.9 & 67.2 & \ccc{+20.7} \\
    \rowcolor{DrawioPurple}
    \cellcolor{white} \ourr & 16 &&
        \textbf{96.5} & \textbf{88.9} & \textbf{66.8} & 65.4 & \textbf{95.1} & \textbf{73.0} & \textbf{72.6} & 64.4 & \textbf{89.8} & \textbf{80.9} & \textbf{79.3} & \ccc{+32.8} \\
    
    \bottomrule
  \end{tabular}}
  \label{supp:tab:ow-li-llava}
\end{subtable}
\end{table*}

\begin{table*}

\newcommand\random{\multirow{3}{*}{\makecell{\textit{Random} \\ \textit{Ctx}}}}
\newcommand\pseudo{\multirow{3}{*}{\makecell{\textit{Pseudo} \\ \textit{ICL}}}}
\newcommand\ourr{\multirow{-3}{*}{\ours}}

\ContinuedFloat
\begin{subtable}{\textwidth}
\caption{\textbf{Phi-3.5-Vision and Phi-4-Multimodal.}}
  \centering
  \resizebox{\linewidth}{!}{
  \begin{tabular}{rc c cccccccccccc}
    \toprule
    Method & Shots && C101 & DTD & ESAT & FGVC & FLWR & FOOD & PETS & CARS & S397 & U101 & Avg. & $\Delta$ \\
    \midrule

    \rowcolor{DrawioYellow}
    \multicolumn{15}{c}{\emph{Phi-3.5-Vision}} \\ [-2ex] \\

    \textit{Zero-Shot} & - &&
        75.0 & 45.6 & 1.7 & 60.7 & 61.8 & 42.9 & 47.4 & 47.7 & 46.4 & 38.8 & 46.8 \\
    \cmidrule(lr){2-15}
    \random & 4 &&
        66.6 & 19.9 & 5.0 & 39.2 & 46.5 & 20.0 & 20.7 & 64.2 & 31.9 & 19.2 & 33.3 & \ccc{-13.5} \\
    & 8 &&
        61.7 & 15.0 & 2.8 & 47.9 & 44.0 & 14.5 & 21.7 & 65.6 & 28.4 & 16.1 & 31.8 & \ccc{-15.0} \\
    & 16 &&
        56.5 & 15.4 & 0.8 & 55.6 & 43.4 & 8.6 & 20.2 & 60.9 & 24.1 & 15.0 & 30.0 & \ccc{-16.8} \\
    \cmidrule(lr){2-15}
    \pseudo & 4 &&
        66.9 & 34.5 & 1.0 & 35.7 & 40.4 & 44.0 & 45.1 & 14.1 & 41.1 & 35.5 & 35.8 & \ccc{-11.0} \\
    & 8 &&
        68.4 & 39.8 & 0.0 & 35.8 & 31.7 & 40.0 & 31.8 & 15.8 & 50.3 & 32.5 & 34.6 & \ccc{-12.2} \\
    & 16 &&
        66.7 & 40.2 & 0.0 & 44.0 & 26.1 & 42.4 & 36.5 & 16.0 & 44.7 & 35.3 & 35.2 & \ccc{-11.6} \\
    \cmidrule(lr){2-15}
    \rowcolor{DrawioPurple} \cellcolor{white}
    & 4 &&
        79.2 & 66.0 & 29.5 & 98.5 & \textbf{89.9} & 78.1 & \textbf{86.4} & 79.2 & 82.4 & 61.4 & 75.1 & \ccc{+28.3} \\
    \rowcolor{DrawioPurple} \cellcolor{white}
    & 8 &&
        84.9 & \textbf{75.4} & 36.0 & 94.7 & 88.6 & 80.7 & 84.8 & 91.1 & 84.6 & \textbf{66.5} & 78.7 & \ccc{+31.9} \\
    \rowcolor{DrawioPurple}
    \cellcolor{white} \ourr & 16 &&
        \textbf{93.5} & 75.1 & \textbf{47.8} & \textbf{99.1} & 84.1 & \textbf{94.6} & 85.7 & \textbf{100.0} & \textbf{90.6} & 52.1 & \textbf{82.3} & \ccc{+35.5} \\

    \cmidrule{1-15}
    \rowcolor{DrawioYellow}
    \multicolumn{15}{c}{\emph{Phi-4-Multimodal}} \\ [-2ex] \\

    \textit{Zero-Shot} & - &&
        76.6 & 32.4 & 10.0 & 67.1 & 54.2 & 15.0 & 43.9 & 80.1 & 23.0 & 21.3 & 42.4 \\
    \cmidrule(lr){2-15}
    \random & 4 &&
        15.5 & 5.9 & 17.7 & 62.7 & 44.7 & 9.1 & 36.0 & 77.3 & 6.3 & 5.3 & 28.0 & \ccc{-14.4} \\
    & 8 &&
        19.7 & 3.3 & 15.4 & 64.1 & 45.3 & 8.1 & 40.3 & 79.6 & 6.2 & 4.5 & 28.6 & \ccc{-13.8} \\
    & 16 &&
        18.3 & 1.7 & 12.1 & 65.7 & 46.3 & 6.7 & 39.7 & 78.9 & 4.3 & 3.6 & 27.7 & \ccc{-14.7} \\
    \cmidrule(lr){2-15}
    \pseudo & 4 &&
        75.7 & 26.8 & 5.4 & 67.7 & 50.6 & 16.6 & 45.9 & 77.8 & 28.1 & 13.0 & 40.8 & \ccc{-1.6} \\
    & 8 &&
        74.1 & 19.7 & 10.1 & 67.7 & 47.3 & 15.9 & 45.9 & 78.6 & 24.4 & 14.9 & 39.8 & \ccc{-2.6} \\
    & 16 &&
        76.6 & 42.7 & 2.4 & 67.8 & 45.5 & 15.6 & 45.9 & 75.5 & 26.6 & 12.2 & 41.1 & \ccc{-1.3} \\
    \cmidrule(lr){2-15}
    \rowcolor{DrawioPurple} \cellcolor{white}
    & 4 &&
        88.3 & 31.1 & 21.0 & \textbf{95.6} & 88.6 & \textbf{76.4} & \textbf{77.0} & 89.1 & 85.7 & 68.0 & 72.1 & \ccc{+29.7} \\
    \rowcolor{DrawioPurple} \cellcolor{white}
    & 8 &&
        87.6 & 72.1 & 38.7 & 88.5 & 89.0 & 75.5 & 73.3 & \textbf{94.4} & 83.1 & \textbf{73.7} & 77.6 & \ccc{+35.2} \\
    \rowcolor{DrawioPurple}
    \cellcolor{white} \ourr & 16 &&
        \textbf{93.7} & \textbf{72.9} & \textbf{61.9} & 70.7 & \textbf{92.5} & 70.2 & 74.7 & 79.7 & \textbf{89.3} & 68.1 & \textbf{77.4} & \ccc{+35.0} \\
    
    \bottomrule
  \end{tabular}}
  \label{supp:tab:ow-li-phi}
\end{subtable}
\end{table*}

\section{Streaming results}
\label{supp:sec:ow_streaming}

We provide detailed results for the Open-World \textit{Streaming} experiments in this section.
We evaluate all five models on the ten datasets, testing the \textit{Pseudo-label} ICL variant and \ours.

We organize the results as follows:
\begin{itemize}
    \item \textbf{Main results.} \cref{supp:tab:ow-streaming} presents the aggregated performance for the Qwen series, LLaVa OneVision, and the Phi series, categorizing results by dataset group: \textit{Prototypical}, \textit{Non-prototypical}, \textit{Fine-grained}, and \textit{Very fine-grained}.

    \item \textbf{Metric-specific results.} \cref{supp:tab:ow-streaming-ss,supp:tab:ow-streaming-mcs,supp:tab:ow-streaming-bcs,supp:tab:ow-streaming-li} provide granular breakdowns across the ten datasets for \texttt{SS}, \texttt{bCS}, \texttt{mCS}, and \texttt{LI}, respectively.
    We report the average score and the improvement ($\Delta$) relative to the \textit{Zero-Shot} baseline.
\end{itemize}

\myparagraph{Results discussion.}
The streaming results in \cref{supp:tab:ow-streaming} confirm the robustness of \ours in a streaming scenario.
Across the majority of models and dataset groups, \ours consistently outperforms both the \textit{Zero-Shot} baseline and the \textit{Pseudo ICL} approach.
For example, on \textit{Prototypical} datasets, \ours achieves a \texttt{LI} of 90.4 with Qwen2-VL (vs. 78.7 \textit{Zero-Shot}) and 84.9 with Phi-4-Multimodal (vs. 49.8 \textit{Zero-Shot}).
Similarly, \texttt{SS} increases from 51.9 to 60.9 for Qwen2-VL and from 57.4 to 62.7 for Phi-4-Multimodal, while \texttt{mCS} from 43.7 to 59.5 and from 57.2 to 64.6, respectively.
While \textit{Pseudo ICL} occasionally shows strength in specific similarity metrics (\eg, \texttt{SS} for Phi-3.5-Vision), it is inconsistent across models, metrics, and datasets.
In contrast, \ours exhibits superior overall stability.

\begin{table*}

\newcommand\pseudo{\makecell{\textit{Pseudo} \textit{ICL}}}
\newcommand\ourr{\ours}

\caption{\textbf{Streaming results.}
  We report results for \textit{Llama Inclusion} (\texttt{LI}), \textit{Semantic Similarity} (\texttt{SS}), \textit{Concept Similarity} ($\texttt{bCS}$), and \textit{Median Concept Similarity} ($\texttt{mCS}$).
  \inlineColorbox{DrawioPurple}{Purple} indicates our \ours.
  Higher is better on all metrics.
  For each LMM, \textbf{bold} indicates the best result.
  }
  \centering
  \resizebox{\linewidth}{!}{
  \begin{tabular}{r c cccc c cccc c cccc c cccc}
    \toprule
    \multirow[c]{2.5}{*}{Method} && \multicolumn{4}{c}{Prototypical} && \multicolumn{4}{c}{Non-prototypical} && \multicolumn{4}{c}{Fine-grained} && \multicolumn{4}{c}{Very fine-grained} \\
    \cmidrule{3-6} \cmidrule{8-11} \cmidrule{13-16} \cmidrule{18-21}
    &&
        \texttt{LI} & \texttt{SS} & \texttt{bCS} & \texttt{mCS}
    &&
        \texttt{LI} & \texttt{SS} & \texttt{bCS} & \texttt{mCS}
    &&
        \texttt{LI} & \texttt{SS} & \texttt{bCS} & \texttt{mCS}
    &&
        \texttt{LI} & \texttt{SS} & \texttt{bCS} & \texttt{mCS} \\
    \midrule

    \rowcolor{DrawioYellow}
    \multicolumn{21}{c}{\emph{Qwen2-VL 7B}} \\ [-2ex] \\
    \textit{Zero-Shot} &&
        78.7 & 51.9 & \textbf{76.0} & 43.7 &&
        42.6 & 30.8 & \textbf{49.8} & 29.2 &&
        64.0 & 39.2 & 62.9 & 31.9 &&
        63.0 & 34.5 & 43.4 & 33.1
        \\
    \pseudo &&
        65.9 & \textbf{62.9} & 70.5 & \textbf{61.4} &&
        36.5 & 36.5 & 44.3 & 36.3 &&
        54.0 & 54.0 & 59.3 & 52.6 &&
        40.0 & 40.0 & 46.5 & \textbf{39.9}
        \\
    \rowcolor{DrawioPurple}
    \cellcolor{white} \ourr &&
       \textbf{90.4} & 60.9 & 71.7 & 59.5 &&
       \textbf{58.8} & \textbf{38.8} & 46.1 & \textbf{38.4} &&
       \textbf{83.4} & \textbf{56.0} & \textbf{66.9} & \textbf{55.2} &&
       \textbf{83.5} & \textbf{42.2} & \textbf{51.4} & 38.7
        \\

    \cmidrule(lr){1-21}
    \rowcolor{DrawioYellow}
    \multicolumn{21}{c}{\emph{Qwen2.5-VL 7B}} \\ [-2ex] \\
    \textit{Zero-Shot} &&
        82.9 & 47.9 & \textbf{79.9} & 31.1 &&
        45.9 & 30.5 & \textbf{54.0} & 24.8 &&
        73.8 & 47.0 & \textbf{78.9} & 29.5 &&
        69.0 & 45.8 & \textbf{68.6} & 27.1
        \\
    \pseudo &&
        63.7 & \textbf{70.4} & 70.6 & \textbf{70.4} &&
        21.7 & \textbf{40.2} & 40.3 & \textbf{40.1} &&
        50.2 & \textbf{58.2} & 58.3 & \textbf{58.0} &&
        73.6 & \textbf{45.9} & 46.2 & \textbf{44.1}
        \\
    \rowcolor{DrawioPurple}
    \cellcolor{white} \ourr &&
        \textbf{87.3} & 66.0 & 66.5 & 66.0 &&
        \textbf{60.2} & 39.5 & 40.6 & 39.5 &&
        \textbf{81.0} & 55.0 & 56.7 & 54.9 &&
        \textbf{86.7} & 37.3 & 37.3 & 37.3
        \\

    \cmidrule(lr){1-21}
    \rowcolor{DrawioYellow}
    \multicolumn{21}{c}{\emph{LLaVa OneVision 7B}} \\ [-2ex] \\
    \textit{Zero-Shot} &&
        53.2 & 56.2 & 62.0 & 53.4 &&
        28.1 & 31.6 & 43.8 & 30.2 &&
        \textbf{40.4} & 39.0 & 43.9 & 37.2 &&
        \textbf{76.7} & 31.8 & 32.3 & 30.9
        \\
    \pseudo &&
        44.8 & 45.0 & 56.0 & 39.0 &&
        25.2 & 25.8 & 41.1 & 23.1 &&
        25.7 & 34.1 & 42.3 & 33.3 &&
        72.4 & 30.8 & 30.8 & 30.8
        \\
    \rowcolor{DrawioPurple}
    \cellcolor{white} \ourr &&
        \textbf{61.7} & \textbf{70.9} & \textbf{70.9} & \textbf{70.9} &&
        \textbf{33.5} & \textbf{49.7} & \textbf{49.8} & \textbf{49.7} &&
        38.9 & \textbf{45.9} & \textbf{45.9} & \textbf{45.9} &&
        73.8 & \textbf{34.2} & \textbf{34.2} & \textbf{34.2}
        \\

    \cmidrule(lr){1-21}
    \rowcolor{DrawioYellow}
    \multicolumn{21}{c}{\emph{Phi-3.5-Vision}} \\ [-2ex] \\
    \textit{Zero-Shot} &&
        60.7 & 48.2 & \textbf{65.6} & 46.1 &&
        28.7 & 24.9 & \textbf{36.7} & 24.1 &&
        50.7 & 32.1 & \textbf{47.2} & 31.3 &&
        54.2 & 29.5 & \textbf{36.3} & 29.8
        \\
    \pseudo &&
        49.5 & \textbf{58.8} & 61.0 & \textbf{58.5} &&
        19.3 & \textbf{29.1} & 35.5 & 28.9 &&
        41.3 & \textbf{42.7} & 45.6 & 42.0 &&
        \textbf{68.8} & \textbf{32.5} & 32.5 & 32.5
        \\
    \rowcolor{DrawioPurple}
    \cellcolor{white} \ourr &&
        \textbf{78.6} & 50.3 & 55.3 & 52.5 &&
        \textbf{40.6} & 28.5 & 33.4 & \textbf{31.2} &&
        \textbf{84.6} & 38.7 & 46.2 & \textbf{42.2} &&
        68.2 & 31.4 & 33.8 & \textbf{32.6}
        \\

    \cmidrule(lr){1-21}
    \rowcolor{DrawioYellow}
    \multicolumn{21}{c}{\emph{Phi-4-Multimodal}} \\ [-2ex] \\
    \textit{Zero-Shot} &&
        49.8 & 57.4 & 58.7 & 57.2 &&
        21.2 & 29.2 & 32.7 & 29.2 &&
        37.7 & 39.2 & 39.2 & 39.1 &&
        73.6 & 31.6 & 31.7 & 31.6
        \\
    \pseudo &&
        50.6 & 62.3 & 62.3 & 62.3 &&
        18.8 & 36.1 & 36.7 & 36.0 &&
        36.7 & 41.8 & 41.8 & 41.8 &&
        71.3 & 31.7 & 31.7 & 31.7
        \\
    \rowcolor{DrawioPurple}
    \cellcolor{white} \ourr &&
        \textbf{84.9} & \textbf{62.7} & \textbf{69.4} & \textbf{64.6} &&
        \textbf{64.8} & \textbf{40.7} & \textbf{44.5} & \textbf{40.7} &&
        \textbf{76.7} & \textbf{53.2} & \textbf{57.4} & \textbf{53.5} &&
        \textbf{77.8} & \textbf{40.0} & \textbf{45.0} & \textbf{39.1}
        \\
    
    \bottomrule
  \end{tabular}}
  \label{supp:tab:ow-streaming}
\end{table*}

\begin{table*}
\caption{
\textbf{Streaming results. \textit{Semantic Similarity}} (\texttt{SS}) on the ten datasets.
\inlineColorbox{DrawioPurple}{Purple} indicates our \ours.
Higher is better.
For each LMM, \textbf{bold} indicates the best result.
$\Delta$ computed \wrt the average (Avg.) of \textit{Zero-Shot}.
For each LMM, \textbf{bold} indicates the best result.
}
  \centering
  \resizebox{\linewidth}{!}{
  \begin{tabular}{rc cccccccccccc}
    \toprule
    Method && C101 & DTD & ESAT & FGVC & FLWR & FOOD & PETS & CARS & S397 & U101 & Avg. & $\Delta$ \\
    \midrule

    \rowcolor{DrawioYellow}
    \multicolumn{14}{c}{\emph{Qwen2-VL 7B}} \\ [-2ex] \\
    \textit{Zero-Shot} &&
         55.8 & 28.6 & 20.7 & 20.6 & 41.7 & 50.7 & 25.1 & 48.3 & 48.1 & 43.1 & 38.3 \\
    \textit{Pseudo ICL} &&
        \textbf{70.0} & 31.9 & 24.3 & 25.8 & 51.6 & \textbf{70.0} & 40.5 & \textbf{54.2} & 55.7 & \textbf{53.4} & 47.7 & \ccc{+9.4} \\
    \rowcolor{DrawioPurple} \cellcolor{white}
    \ours &&
        63.2 & \textbf{34.0} & \textbf{32.1} & \textbf{33.8} & \textbf{54.1} & 62.0 & \textbf{51.9} & 50.6 & \textbf{58.5} & 50.2 & \textbf{49.0} & \ccc{+10.7} \\
    \cmidrule(lr){2-14}

    \rowcolor{DrawioYellow}
    \multicolumn{14}{c}{\emph{Qwen2.5-VL 7B}} \\ [-2ex] \\
    \textit{Zero-Shot} &&
         48.8 & 28.3 & 19.0 & 36.7 & 47.4 & 52.4 & 41.1 & 54.9 & 47.0 & 44.2 & 42.0 \\
    \textit{Pseudo ICL} &&
        \textbf{78.7} & \textbf{37.1} & 27.4 & 29.8 & \textbf{66.5} & \textbf{66.8} & 41.2 & \textbf{61.9} & \textbf{62.2} & \textbf{56.2} & \textbf{52.8} & \ccc{+10.8} \\
    \rowcolor{DrawioPurple} \cellcolor{white}
    \ours &&
        70.5 & 35.5 & \textbf{32.0} & \textbf{30.1} & 63.6 & 51.9 & \textbf{49.5} & 44.4 & 61.6 & 51.0 & 49.0 & \ccc{+7.0} \\
    \cmidrule(lr){2-14}

    \rowcolor{DrawioYellow}
    \multicolumn{14}{c}{\emph{LLaVa OneVision 7B}} \\ [-2ex] \\
    \textit{Zero-Shot} &&
        68.9 & 32.0 & 19.4 & \textbf{29.4} & 37.5 & 41.6 & 37.8 & 34.3 & 43.4 & 43.4 & 38.8 \\
    \textit{Pseudo ICL} &&
        59.7 & 29.0 & 17.0 & \textbf{29.4} & 27.6 & 36.8 & 37.9 & 32.2 & 30.4 & 31.5 & 33.1 & \ccc{-5.7} \\
    \rowcolor{DrawioPurple} \cellcolor{white}
    \ours &&
         \textbf{77.4} & \textbf{57.8} & \textbf{39.7} & \textbf{29.4} & \textbf{42.5} & \textbf{54.2} & \textbf{40.9} & \textbf{39.1} & \textbf{64.3} & \textbf{51.7} & \textbf{49.7} & \ccc{+10.9} \\
    \cmidrule(lr){2-14}

    \rowcolor{DrawioYellow}
    \multicolumn{14}{c}{\emph{Phi-3.5-Vision}} \\ [-2ex] \\
    \textit{Zero-Shot} &&
         53.2 & 29.1 & 7.4 & 19.9 & 31.6 & 40.2 & 24.6 & \textbf{39.1} & 43.2 & 38.3 & 32.6 \\
    \textit{Pseudo ICL} &&
        \textbf{71.7} & \textbf{32.8} & 11.8 & \textbf{29.3} & 44.1 & \textbf{50.7} & \textbf{33.2} & 35.7 & \textbf{45.8} & \textbf{42.7} & \textbf{39.8} & \ccc{+7.2} \\
    \rowcolor{DrawioPurple} \cellcolor{white}
    \ours &&
        55.4 & 28.1 & \textbf{24.4} & 26.6 & \textbf{44.2} & 39.7 & 32.1 & 36.3 & 45.1 & 33.0 & 36.5 & \ccc{+3.9} \\
    \cmidrule(lr){2-14}

    \rowcolor{DrawioYellow}
    \multicolumn{14}{c}{\emph{Phi-4-Multimodal}} \\ [-2ex] \\
    \textit{Zero-Shot} &&
         73.5 & 33.9 & 13.8 & 29.2 & 42.2 & 37.8 & 37.5 & 34.1 & 41.2 & 40.0 & 38.3 \\
    \textit{Pseudo ICL} &&
        \textbf{78.1} & 40.5 & 28.6 & \textbf{29.4} & 42.4 & 44.9 & 38.1 & 34.0 & 46.5 & 39.3 & 42.2 & \ccc{+3.9} \\
    \rowcolor{DrawioPurple} \cellcolor{white}
    \ours &&
        69.1 & \textbf{41.4} & \textbf{34.3} & 27.7 & \textbf{49.0} & \textbf{60.5} & \textbf{50.2} & \textbf{52.3} & \textbf{56.3} & \textbf{46.5} & \textbf{48.7} & \ccc{+10.4} \\
    
    \bottomrule
  \end{tabular}}
  \label{supp:tab:ow-streaming-ss}
\end{table*}

\begin{table*}
\caption{
\textbf{Streaming results. \textit{Concept Similarity}} (\texttt{bCS}) on the ten datasets.
\inlineColorbox{DrawioPurple}{Purple} indicates our \ours.
Higher is better.
For each LMM, \textbf{bold} indicates the best result.
$\Delta$ computed \wrt the average (Avg.) of \textit{Zero-Shot}.
For each LMM, \textbf{bold} indicates the best result.
}
  \centering
  \resizebox{\linewidth}{!}{
  \begin{tabular}{rc cccccccccccc}
    \toprule
    Method && C101 & DTD & ESAT & FGVC & FLWR & FOOD & PETS & CARS & S397 & U101 & Avg. & $\Delta$ \\
    \midrule

    \rowcolor{DrawioYellow}
    \multicolumn{14}{c}{\emph{Qwen2-VL 7B}} \\ [-2ex] \\
    \textit{Zero-Shot} &&
         \textbf{81.3} & \textbf{50.3} & \textbf{39.8} & 30.7 & \textbf{68.7} & 77.0 & 43.2 & 55.7 & \textbf{70.7} & 59.0 & 57.6 \\
    \textit{Pseudo ICL} &&
        79.9 & 44.2 & 32.8 & 30.7 & 59.1 & \textbf{77.3} & 41.4 & \textbf{62.3} & 61.0 & 55.8 & 54.5 & \ccc{-3.1} \\
    \rowcolor{DrawioPurple} \cellcolor{white}
    \ours &&
        77.0 & 42.7 & 36.2 & \textbf{46.9} & 62.6 & 71.7 & \textbf{66.5} & 56.0 & 66.3 & \textbf{59.3} & \textbf{58.5} & \ccc{+0.9} \\
    \cmidrule(lr){2-14}

    \rowcolor{DrawioYellow}
    \multicolumn{14}{c}{\emph{Qwen2.5-VL 7B}} \\ [-2ex] \\
    \textit{Zero-Shot} &&
         \textbf{85.6} & \textbf{53.4} & \textbf{41.3} & \textbf{68.7} & \textbf{79.7} & \textbf{79.6} & \textbf{77.3} & \textbf{68.5} & \textbf{74.2} & \textbf{67.2} & \textbf{69.5} \\
    \textit{Pseudo ICL} &&
        79.1 & 37.2 & 27.5 & 30.3 & 66.9 & 66.9 & 41.2 & 62.2 & 62.2 & 56.2 & 53.0 & \ccc{-16.5} \\
    \rowcolor{DrawioPurple} \cellcolor{white}
    \ours &&
        71.4 & 37.5 & 32.1 & 30.2 & 64.7 & 52.5 & 52.9 & 44.4 & 61.7 & 52.2 & 50.0 & \ccc{-19.5} \\
    \cmidrule(lr){2-14}

    \rowcolor{DrawioYellow}
    \multicolumn{14}{c}{\emph{LLaVa OneVision 7B}} \\ [-2ex] \\
    \textit{Zero-Shot} &&
        79.1 & 46.9 & 41.0 & \textbf{29.4} & \textbf{51.9} & 41.9 & 37.9 & 35.3 & 44.8 & 43.5 & 45.2 \\
    \textit{Pseudo ICL} &&
        \textbf{81.6} & 49.4 & 36.9 & \textbf{29.4} & 48.6 & 40.4 & 37.9 & 32.2 & 31.5 & 37.0 & 42.5 & \ccc{-2.7} \\
    \rowcolor{DrawioPurple} \cellcolor{white}
    \ours &&
        77.4 & \textbf{57.8} & \textbf{39.7} & \textbf{29.4} & 42.5 & 54.2 & \textbf{40.9} & \textbf{39.1} & \textbf{64.3} & \textbf{51.8} & \textbf{49.7} & \ccc{+4.5} \\
    \cmidrule(lr){2-14}

    \rowcolor{DrawioYellow}
    \multicolumn{14}{c}{\emph{Phi-3.5-Vision}} \\ [-2ex] \\
    \textit{Zero-Shot} &&
         73.6 & \textbf{43.5} & 16.4 & 29.6 & 44.9 & \textbf{58.1} & \textbf{38.7} & \textbf{43.0} & \textbf{57.6} & \textbf{50.1} & \textbf{45.5} \\
    \textit{Pseudo ICL} &&
        \textbf{74.1} & 39.9 & 20.2 & 29.4 & 44.7 & 56.2 & 35.9 & 35.7 & 47.9 & 46.3 & 43.0 & \ccc{-2.5} \\
    \rowcolor{DrawioPurple} \cellcolor{white}
    \ours &&
        59.3 & 34.8 & \textbf{24.9} & \textbf{30.9} & \textbf{53.6} & 46.7 & 38.3 & 36.6 & 51.3 & 40.5 & 41.7 & \ccc{-3.8} \\
    \cmidrule(lr){2-14}

    \rowcolor{DrawioYellow}
    \multicolumn{14}{c}{\emph{Phi-4-Multimodal}} \\ [-2ex] \\
    \textit{Zero-Shot} &&
         75.8 & 40.6 & 17.4 & 29.3 & 42.2 & 37.9 & 37.5 & 34.1 & 41.7 & 40.1 & 39.6 \\
    \textit{Pseudo ICL} &&
        \textbf{78.2} & 40.6 & 30.0 & 29.4 & 42.4 & 44.9 & 38.1 & 34.0 & 46.5 & 39.4 & 42.4 & \ccc{+2.8} \\
    \rowcolor{DrawioPurple} \cellcolor{white}
    \ours &&
        77.9 & \textbf{44.0} & \textbf{35.3} & \textbf{34.5} & \textbf{53.7} & \textbf{64.3} & \textbf{54.2} & \textbf{55.5} & \textbf{61.0} & \textbf{54.1} & \textbf{53.4} & \ccc{+13.8} \\
    
    \bottomrule
  \end{tabular}}
  \label{supp:tab:ow-streaming-bcs}
\end{table*}

\begin{table*}
\caption{
\textbf{Streaming results. \textit{Median Concept Similarity}} (\texttt{mCS}) on the ten datasets.
\inlineColorbox{DrawioPurple}{Purple} indicates our \ours.
Higher is better.
For each LMM, \textbf{bold} indicates the best result.
$\Delta$ computed \wrt the average (Avg.) of \textit{Zero-Shot}.
For each LMM, \textbf{bold} indicates the best result.
}
  \centering
  \resizebox{\linewidth}{!}{
  \begin{tabular}{rc cccccccccccc}
    \toprule
    Method && C101 & DTD & ESAT & FGVC & FLWR & FOOD & PETS & CARS & S397 & U101 & Avg. & $\Delta$ \\
    \midrule

    \rowcolor{DrawioYellow}
    \multicolumn{14}{c}{\emph{Qwen2-VL 7B}} \\ [-2ex] \\
    \textit{Zero-Shot} &&
         51.3 & 27.9 & 20.6 & 20.4 & 33.3 & 37.3 & 25.0 & 45.8 & 36.1 & 38.9 & 33.7 \\
    \textit{Pseudo ICL} &&
        \textbf{69.4} & 31.2 & 24.9 & 25.7 & 50.1 & \textbf{67.2} & 40.5 & \textbf{54.2} & 53.3 & \textbf{52.9} & 46.9 & \ccc{+13.2} \\
    \rowcolor{DrawioPurple} \cellcolor{white}
    \ours &&
        61.4 & \textbf{34.8} & \textbf{31.1} & \textbf{30.0} & \textbf{54.9} & 61.4 & \textbf{49.3} & 47.5 & \textbf{57.5} & 49.2 & \textbf{47.7} & \ccc{+14.0} \\
    \cmidrule(lr){2-14}

    \rowcolor{DrawioYellow}
    \multicolumn{14}{c}{\emph{Qwen2.5-VL 7B}} \\ [-2ex] \\
    \textit{Zero-Shot} &&
         33.0 & 26.3 & 18.6 & 24.7 & 29.5 & 31.8 & 27.3 & 29.4 & 29.3 & 29.6 & 27.9 \\
    \textit{Pseudo ICL} &&
        \textbf{78.7} & \textbf{37.0} & 27.2 & 29.8 & \textbf{66.2} & \textbf{66.5} & 41.2 & \textbf{58.4} & \textbf{62.1} & \textbf{55.9} & \textbf{52.3} & \ccc{+24.4} \\
    \rowcolor{DrawioPurple} \cellcolor{white}
    \ours &&
        70.5 & 35.8 & \textbf{31.8} & \textbf{30.2} & 63.1 & 52.1 & \textbf{49.6} & 44.4 & 61.6 & 51.0 & 49.0 & \ccc{+21.1} \\
    \cmidrule(lr){2-14}

    \rowcolor{DrawioYellow}
    \multicolumn{14}{c}{\emph{LLaVa OneVision 7B}} \\ [-2ex] \\
    \textit{Zero-Shot} &&
        64.1 & 30.0 & 17.6 & \textbf{29.4} & 32.5 & 41.4 & 37.8 & 32.5 & 42.6 & 43.1 & 37.1 \\
    \textit{Pseudo ICL} &&
        48.3 & 26.8 & 16.5 & \textbf{29.4} & 26.5 & 35.5 & 37.9 & 32.2 & 29.7 & 26.1 & 30.9 & \ccc{-6.2} \\
    \rowcolor{DrawioPurple} \cellcolor{white}
    \ours &&
        \textbf{77.4} & \textbf{57.8} & \textbf{39.7} & \textbf{29.4} & \textbf{42.5} & \textbf{54.2} & \textbf{40.9} & \textbf{39.0} & \textbf{64.3} & \textbf{51.5} & \textbf{49.7} & \ccc{+12.6} \\
    \cmidrule(lr){2-14}

    \rowcolor{DrawioYellow}
    \multicolumn{14}{c}{\emph{Phi-3.5-Vision}} \\ [-2ex] \\
    \textit{Zero-Shot} &&
         53.1 & 28.8 & 7.4 & 21.2 & 32.3 & 36.6 & 25.0 & \textbf{38.3} & 39.2 & 36.1 & 31.8 \\
    \textit{Pseudo ICL} &&
        \textbf{71.7} & \textbf{32.6} & 11.8 & \textbf{29.4} & 44.0 & \textbf{48.9} & 33.1 & 35.6 & 45.3 & \textbf{42.3} & \textbf{39.5} & \ccc{+7.7} \\
    \rowcolor{DrawioPurple} \cellcolor{white}
    \ours &&
        56.7 & 31.5 & \textbf{24.6} & 28.8 & \textbf{44.2} & 44.8 & \textbf{37.8} & 36.4 & \textbf{48.2} & 37.6 & 39.1 & \ccc{+7.3} \\
    \cmidrule(lr){2-14}

    \rowcolor{DrawioYellow}
    \multicolumn{14}{c}{\emph{Phi-4-Multimodal}} \\ [-2ex] \\
    \textit{Zero-Shot} &&
         73.3 & 34.4 & 13.6 & 29.2 & 42.2 & 37.8 & 37.5 & 34.1 & 41.2 & 39.7 & 38.3 \\
    \textit{Pseudo ICL} &&
        \textbf{78.1} & 40.6 & 28.6 & \textbf{29.4} & 42.3 & 44.9 & 38.1 & 34.0 & 46.5 & 39.0 & 42.1 & \ccc{+3.8} \\
    \rowcolor{DrawioPurple} \cellcolor{white}
    \ours &&
        71.1 & \textbf{42.2} & \textbf{33.6} & 27.4 & \textbf{48.9} & \textbf{60.2} & \textbf{51.5} & \textbf{50.7} & \textbf{58.1} & \textbf{46.3} & \textbf{49.0} & \ccc{+10.7} \\
    
    \bottomrule
  \end{tabular}}
  \label{supp:tab:ow-streaming-mcs}
\end{table*}

\begin{table*}
\caption{
\textbf{Streaming open-world results. \textit{Llama Inclusion}} (\texttt{LI}) on the ten datasets.
\inlineColorbox{DrawioPurple}{Purple} indicates our \ours.
Higher is better.
For each LMM, \textbf{bold} indicates the best result.
$\Delta$ computed \wrt the average (Avg.) of \textit{Zero-Shot}.
For each LMM, \textbf{bold} indicates the best result.
}
  \centering
  \resizebox{\linewidth}{!}{
  \begin{tabular}{rc cccccccccccc}
    \toprule
    Method && C101 & DTD & ESAT & FGVC & FLWR & FOOD & PETS & CARS & S397 & U101 & Avg. & $\Delta$ \\
    \midrule

    \rowcolor{DrawioYellow}
    \multicolumn{14}{c}{\emph{Qwen2-VL 7B}} \\ [-2ex] \\
    \textit{Zero-Shot} &&
         84.0 & 59.5 & 17.7 & 55.5 & 68.9 & 74.3 & 46.0 & 63.5 & 72.2 & 47.7 & 58.9 \\
    \textit{Pseudo ICL} &&
        83.9 & 41.8 & 12.6 & 63.6 & 55.8 & 60.3 & 49.0 & 42.8 & 47.8 & 33.7 & 49.1 & \ccc{-9.8} \\
    \rowcolor{DrawioPurple} \cellcolor{white}
    \ours &&
        \textbf{93.5} & \textbf{61.1} & \textbf{43.9} & \textbf{76.4} & \textbf{90.0} & \textbf{85.5} & \textbf{74.7} & \textbf{90.7} & \textbf{87.2} & \textbf{71.5 }& \textbf{77.4} & \ccc{+18.5} \\
    \cmidrule(lr){2-14}

    \rowcolor{DrawioYellow}
    \multicolumn{14}{c}{\emph{Qwen2.5-VL 7B}} \\ [-2ex] \\
    \textit{Zero-Shot} &&
         84.3 & 58.9 & 12.5 & 68.8 & 74.7 & \textbf{76.1} & 70.7 & 69.3 & 81.6 & 66.3 & 66.3 \\
    \textit{Pseudo ICL} &&
        80.9 & 19.1 & 12.7 & 72.8 & 62.8 & 40.2 & 47.8 & 74.4 & 46.5 & 33.5 & 49.0 & \ccc{-17.3} \\
    \rowcolor{DrawioPurple} \cellcolor{white}
    \ours &&
        \textbf{89.5} & \textbf{60.3} & \textbf{42.9} & \textbf{94.2} & \textbf{93.5} & 73.5 & \textbf{76.1} & \textbf{79.1} & \textbf{85.0} & \textbf{77.3} & \textbf{77.2} & \ccc{+10.9} \\
    \cmidrule(lr){2-14}

    \rowcolor{DrawioYellow}
    \multicolumn{14}{c}{\emph{LLaVa OneVision 7B}} \\ [-2ex] \\
    \textit{Zero-Shot} &&
        {81.3} & 45.6 & 11.8 & \textbf{68.9} & \textbf{48.9} & 22.0 & \textbf{50.2} & \textbf{84.4} & 25.0 & \textbf{27.0} & 46.5 \\
    \textit{Pseudo ICL} &&
        \textbf{81.5} & \textbf{47.5} & 3.9 & 65.9 & 25.7 & 10.1 & 41.4 & 79.0 & 8.1 & 24.3 & 38.7 & \ccc{-7.8} \\
    \rowcolor{DrawioPurple} \cellcolor{white}
    \ours &&
        77.0 & 40.3 & \textbf{37.7} & 65.7 & 45.4 & \textbf{23.5} & 47.8 & 81.9 & \textbf{46.4} & 22.4 & \textbf{48.8} & \ccc{+2.3} \\
    \cmidrule(lr){2-14}

    \rowcolor{DrawioYellow}
    \multicolumn{14}{c}{\emph{Phi-3.5-Vision}} \\ [-2ex] \\
    \textit{Zero-Shot} &&
         75.0 & 45.6 & 1.7 & 60.7 & 61.8 & 42.9 & 47.4 & 47.7 & 46.4 & 38.8 & 46.8 \\
    \textit{Pseudo ICL} &&
        69.4 & 31.7 & 4.8 & 60.9 & 56.9 & 31.3 & 35.8 & \textbf{76.7} & 29.5 & 21.5 & 41.9 & \ccc{-4.9} \\
    \rowcolor{DrawioPurple} \cellcolor{white}
    \ours &&
        \textbf{79.9} & \textbf{46.6} & \textbf{12.0} & \textbf{96.0} & \textbf{88.0} & \textbf{87.0} & \textbf{78.9} & 40.3 & \textbf{77.2} & \textbf{63.1} & \textbf{66.9} & \ccc{+20.1} \\
    \cmidrule(lr){2-14}

    \rowcolor{DrawioYellow}
    \multicolumn{14}{c}{\emph{Phi-4-Multimodal}} \\ [-2ex] \\
    \textit{Zero-Shot} &&
        76.6 & 32.4 & 10.0 & 67.1 & 54.2 & 15.0 & 43.9 & \textbf{80.1} & 23.0 & 21.3 & 42.4 \\
    \textit{Pseudo ICL} &&
        76.1 & 22.7 & 17.4 & 65.9 & 47.6 & 16.9 & 45.7 & 76.7 & 25.1 & 16.3 & 41.0 & \ccc{-1.4} \\
    \rowcolor{DrawioPurple} \cellcolor{white}
    \ours &&
        \textbf{86.4} & \textbf{65.1} & \textbf{60.5} & \textbf{84.1} & \textbf{86.4} & \textbf{78.6} & \textbf{65.0} & 71.6 & \textbf{83.5} & \textbf{68.6} & \textbf{75.0} & \ccc{+32.6} \\
    
    \bottomrule
  \end{tabular}}
  \label{supp:tab:ow-streaming-li}
\end{table*}

\section{Implementation details}
\label{supp:sec:impl-details}

For the closed-world experiments, we evaluate the three most common CLIP~\cite{radford2021learning} variants using ViT-B/32, ViT-B/16, and ViT-L/14~\cite{dosovitskiy2020image}.
For both the closed-world and open-world experiments, we evaluate five Large Multimodal Models (LMMs) from three publicly available model series: (i) Qwen2-VL 7B~\cite{wang2024qwen2} and Qwen2.5-VL 7B~\cite{bai2025qwen25vltechnicalreport}; (ii) LLaVa OneVision 7B~\cite{li2024llava-ov}; (iii) Phi-3.5-Vision~\cite{abdin2024phi} and Phi-4-Multimodal~\cite{microsoft2025phi4minitechnicalreportcompact}.

For LMMs, we use a varying batch size for generation.
For tasks with a small context (\eg, multi-choice experiments), our hardware (see \textit{Resources} below) supports up to batch size 32 for Qwen2-VL and Qwen2.5-VL.
For more complex and rich contexts, such as that of few-shot classification in the closed-world setting (see \cref{sec:closed-world-classification,tab:cw:icl_vs_models}) and the context we build for \ours (see \cref{sec:open-world-classification,tab:ow:main}), the batch size needs to be decreased to 8, 4, or 2 samples per GPU, depending on the LMM.
To reduce VRAM usage, we downscale the context images to $224 \times 224$ pixels.
For reproducibility, we always use greedy decoding up to 64 generated tokens.
For a fair comparison, we use the same experimental framework of~\cite{conti2025large} for all our experiments.
We note that the Llama Inclusion metric is sensitive, and provides incorrect scores for non-sentence responses.
Therefore, we encapsulate the comma-separated class options provided by \ours in the template
``The target object in the photo is one of these [output].''.

We run all our experiments on NVIDIA A100 GPUs with 40, 64, and 80 GB of VRAM.
Simpler experiments (\eg, \textit{Zero-shot} closed-world) can be run on a single GPU, while we run experiments with large contexts on multiple GPUs to reduce wait times, using up to 4 GPUs per experiment.
Evaluation time ranges from a few minutes for \textit{Zero-shot} closed-world experiments to 8-10 hours for the largest datasets (Food101 and SUN397, see \cref{supp:tab:dataset_details} for details on their sizes) in the \textit{Streaming} setting, where the context has to be re-generated many times.

\section{Qualitative results}
\label{supp:sec:qualitatives}

\begin{figure*}[p!]
    \centering
    \begin{subfigure}[b]{0.3\linewidth}
        \centering
        \includegraphics[width=\linewidth]{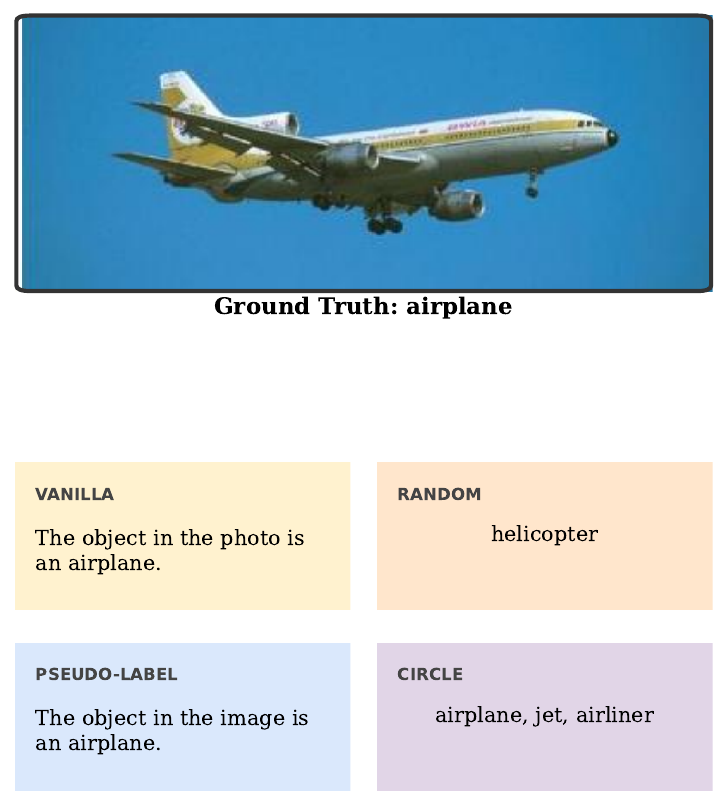}
        \label{supp:fig:c101:a}
    \end{subfigure}
    \hfill
    \begin{subfigure}[b]{0.3\linewidth}
        \centering
        \includegraphics[width=\linewidth]{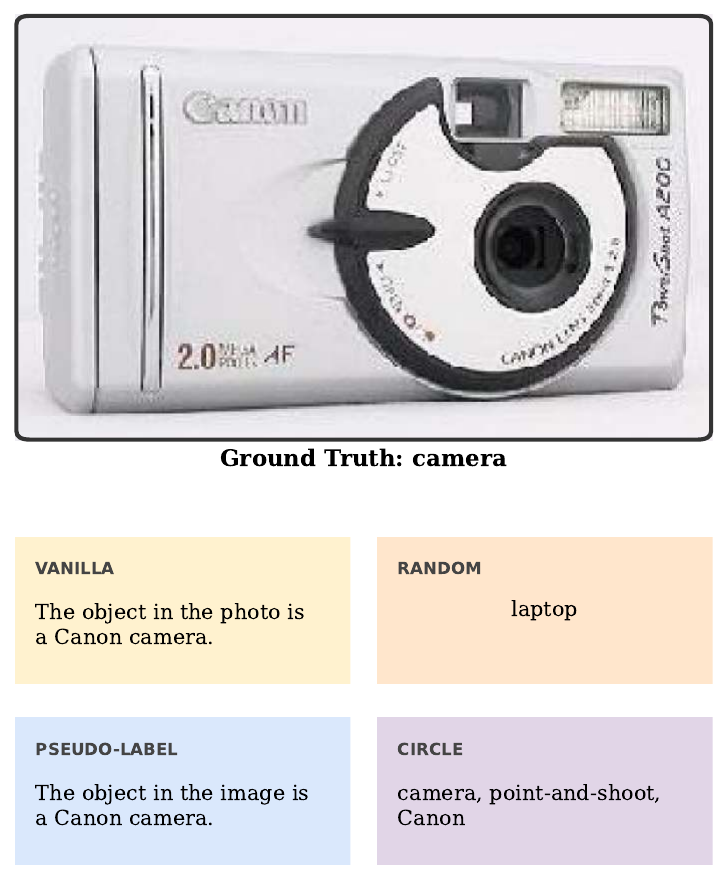}
        \label{supp:fig:c101:b}
    \end{subfigure}
    \hfill
    \begin{subfigure}[b]{0.3\linewidth}
        \centering
        \includegraphics[width=\linewidth]{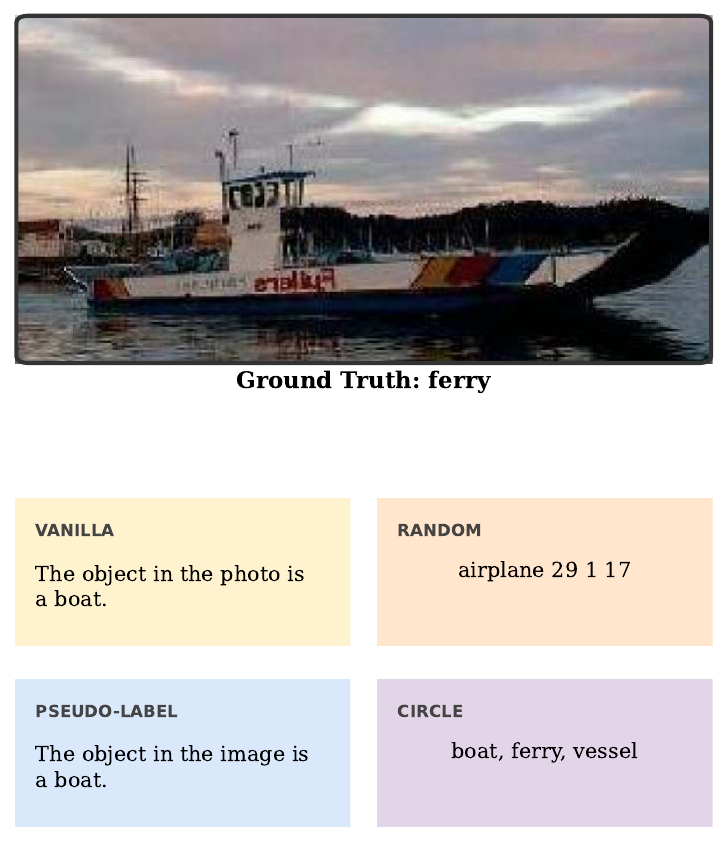}
        \label{supp:fig:c101:c}
    \end{subfigure}
    
    \caption{
    \textbf{Qualitative results from Caltech101~\cite{fei2004caltech101}.}
    We visualize three distinct samples, showing the response given by the \inlineColorbox{DrawioYellow}{\textit{Vanilla}} model, with \inlineColorbox{DrawioOrange}{\textit{Random}} context, with \inlineColorbox{DrawioBlue}{\textit{Pseudo-labeling}} examples, and using our \inlineColorbox{DrawioPurple}{\ours}.
    We use Qwen2-VL~\cite{wang2024qwen2} 7B as the LMM.
    }
    \label{supp:fig:caltech101}
\end{figure*}

\begin{figure*}[p!]
    \centering
    \begin{subfigure}[b]{0.3\linewidth}
        \centering
        \includegraphics[width=\linewidth]{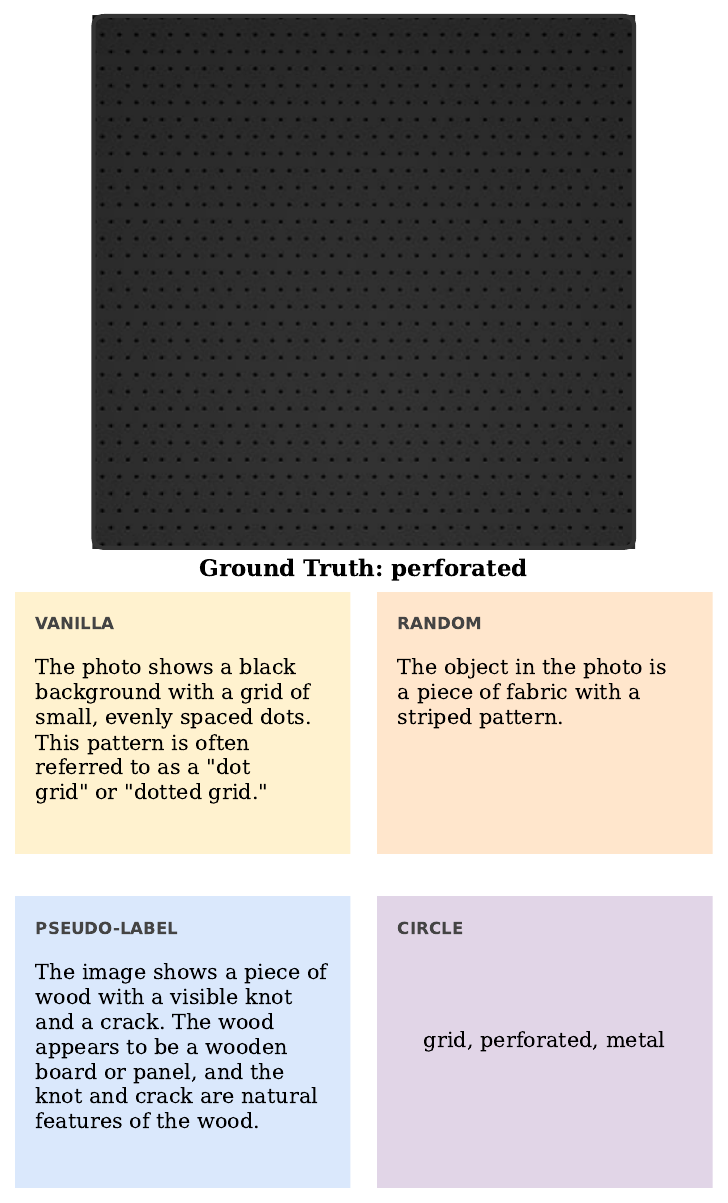}
        \label{supp:fig:dtd:a}
    \end{subfigure}
    \hfill
    \begin{subfigure}[b]{0.3\linewidth}
        \centering
        \includegraphics[width=\linewidth]{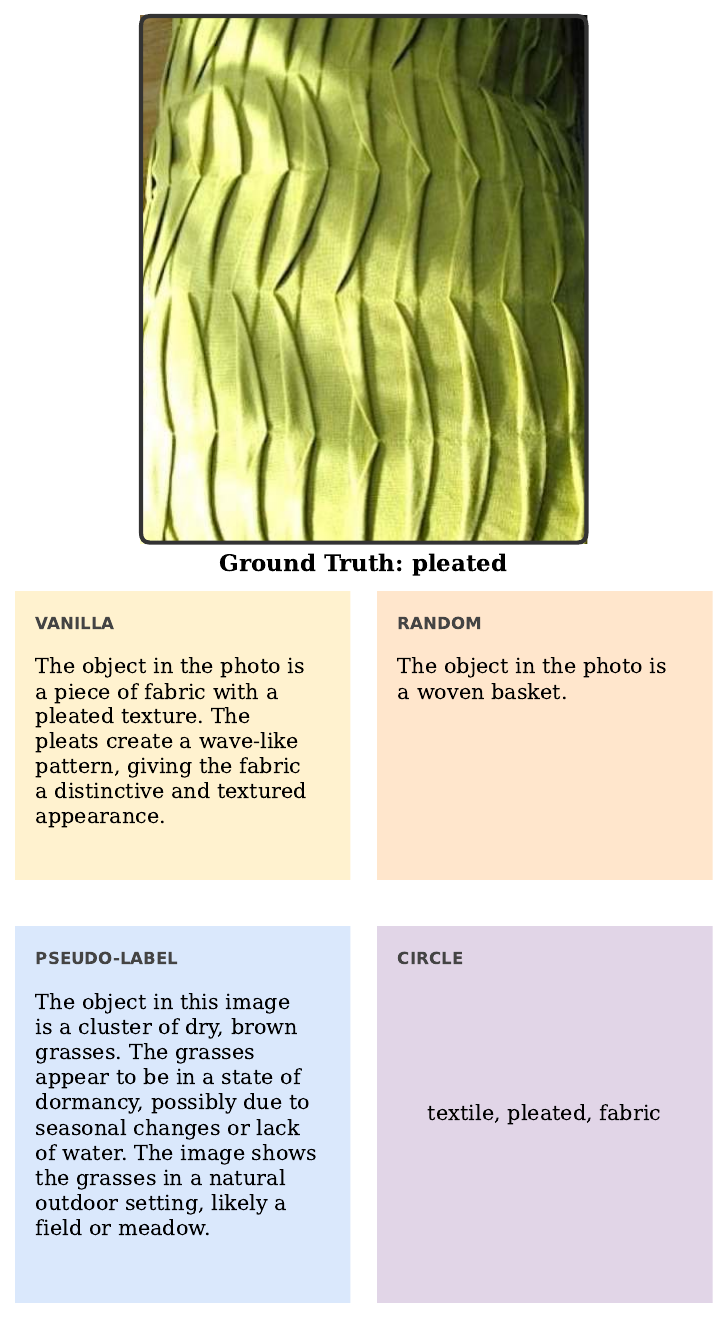}
        \label{supp:fig:dtd:b}
    \end{subfigure}
    \hfill
    \begin{subfigure}[b]{0.3\linewidth}
        \centering
        \includegraphics[width=\linewidth]{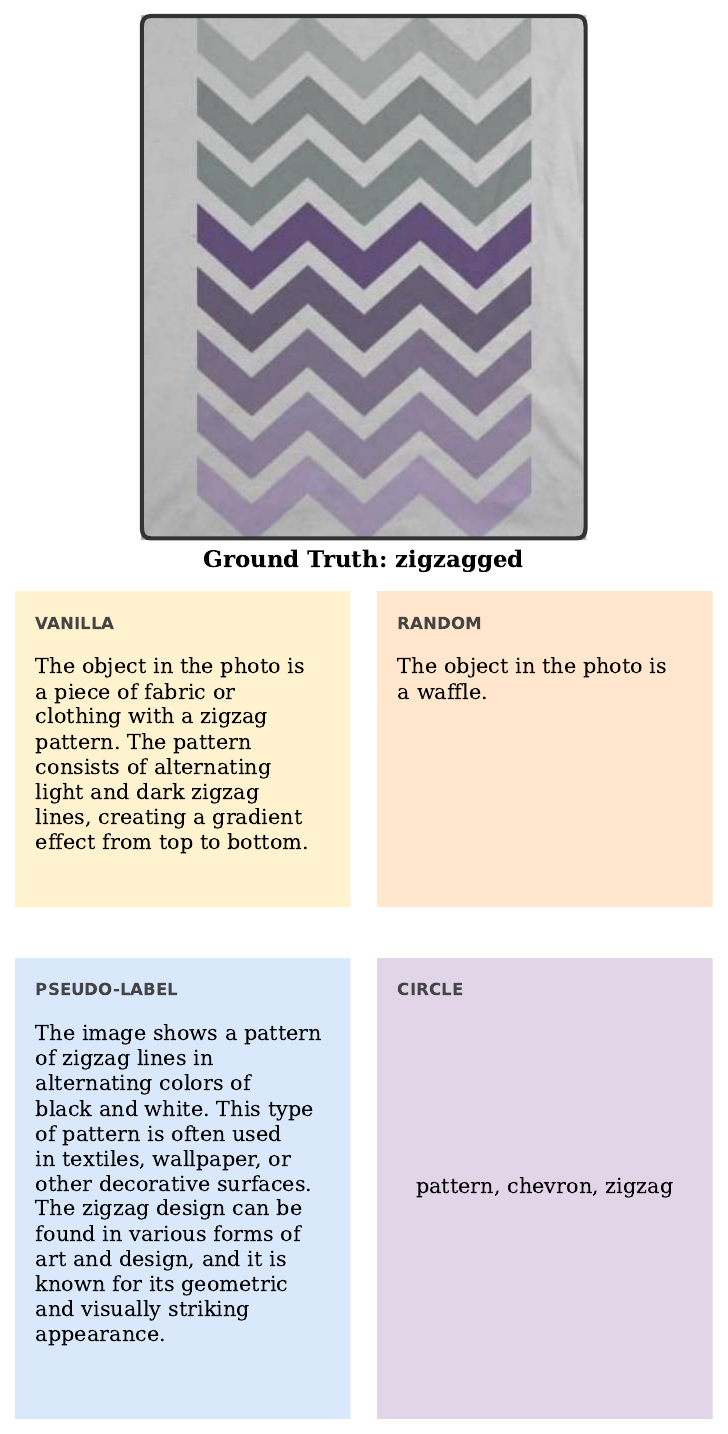}
        \label{supp:fig:dtd:c}
    \end{subfigure}
    
    \caption{
    \textbf{Qualitative results from DTD~\cite{cimpoi2014dtd}}
    We visualize three distinct samples, showing the response given by the \inlineColorbox{DrawioYellow}{\textit{Vanilla}} model, with \inlineColorbox{DrawioOrange}{\textit{Random}} context, with \inlineColorbox{DrawioBlue}{\textit{Pseudo-labeling}} examples, and using our \inlineColorbox{DrawioPurple}{\ours}.
    We use Qwen2-VL~\cite{wang2024qwen2} 7B as the LMM.
    }
    \label{supp:fig:dtd}
\end{figure*}

\begin{figure*}[p!]
    \centering
    \begin{subfigure}[b]{0.3\linewidth}
        \centering
        \includegraphics[width=\linewidth]{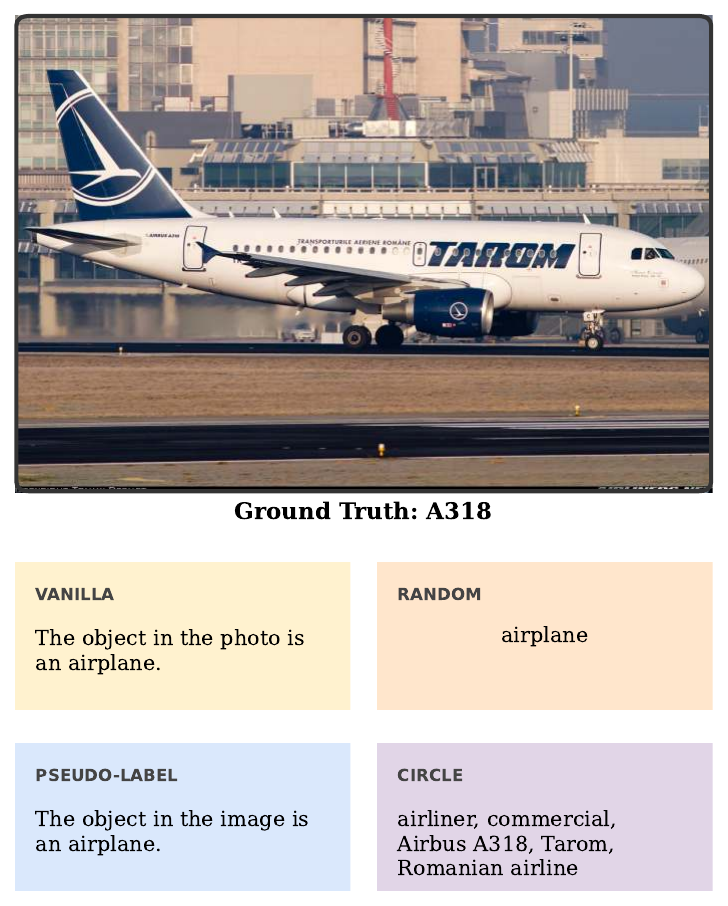}
        \label{supp:fig:fgvc:a}
    \end{subfigure}
    \hfill
    \begin{subfigure}[b]{0.3\linewidth}
        \centering
        \includegraphics[width=\linewidth]{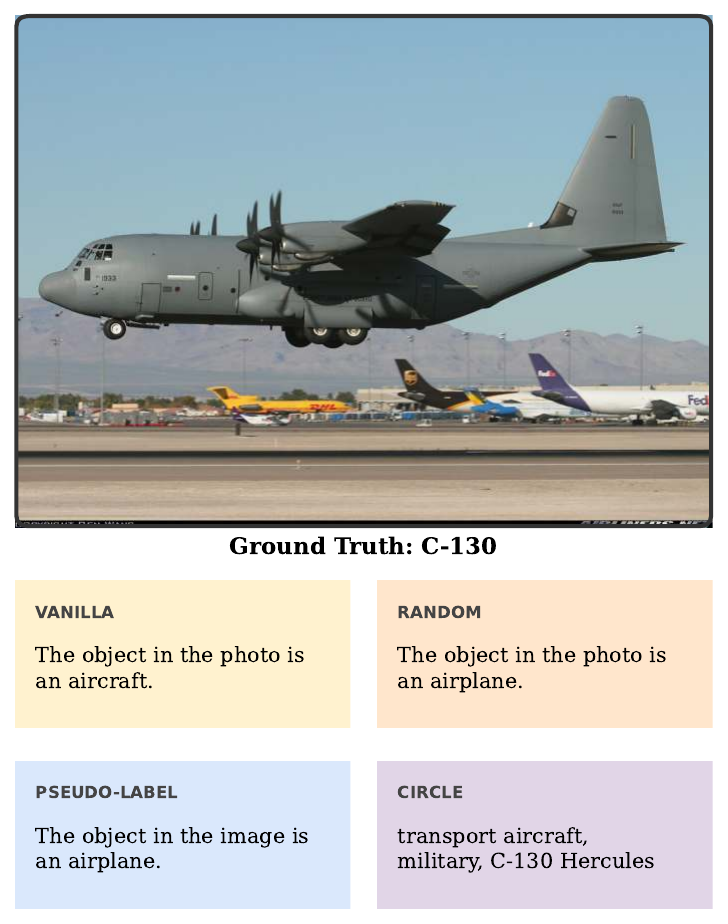}
        \label{supp:fig:fgvc:b}
    \end{subfigure}
    \hfill
    \begin{subfigure}[b]{0.3\linewidth}
        \centering
        \includegraphics[width=\linewidth]{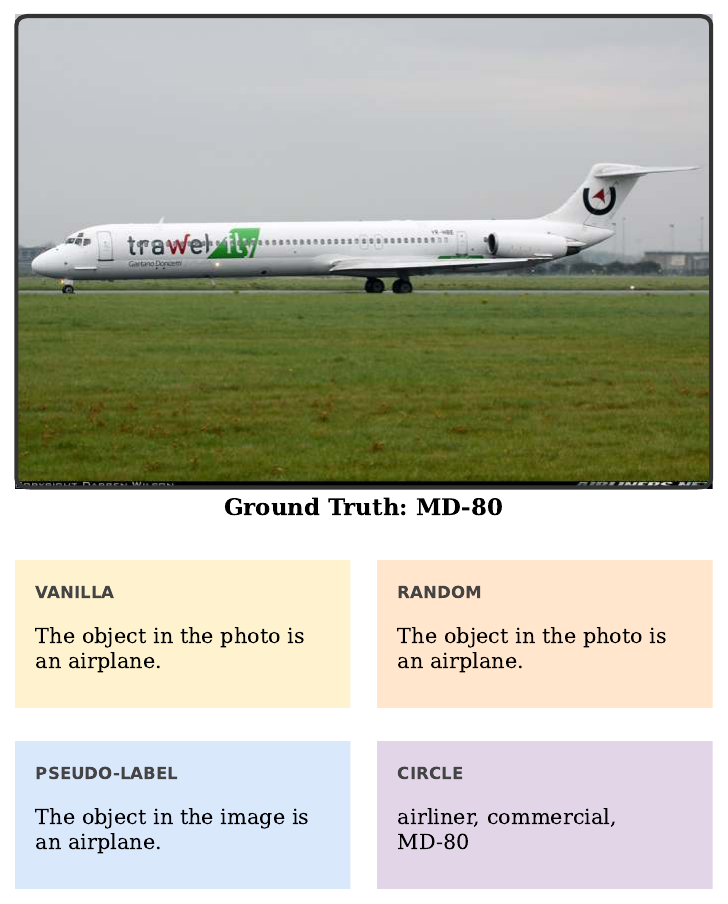}
        \label{supp:fig:fgvc:c}
    \end{subfigure}
    
    \caption{
    \textbf{Qualitative results from FGVC Aircraft~\cite{maji2013aircraft}.}
    We visualize three distinct samples, showing the response given by the \inlineColorbox{DrawioYellow}{\textit{Vanilla}} model, with \inlineColorbox{DrawioOrange}{\textit{Random}} context, with \inlineColorbox{DrawioBlue}{\textit{Pseudo-labeling}} examples, and using our \inlineColorbox{DrawioPurple}{\ours}.
    We use Qwen2-VL~\cite{wang2024qwen2} 7B as the LMM.
    }
    \label{supp:fig:fgvc}
\end{figure*}

\begin{figure*}[p!]
    \centering
    \begin{subfigure}[b]{0.3\linewidth}
        \centering
        \includegraphics[width=\linewidth]{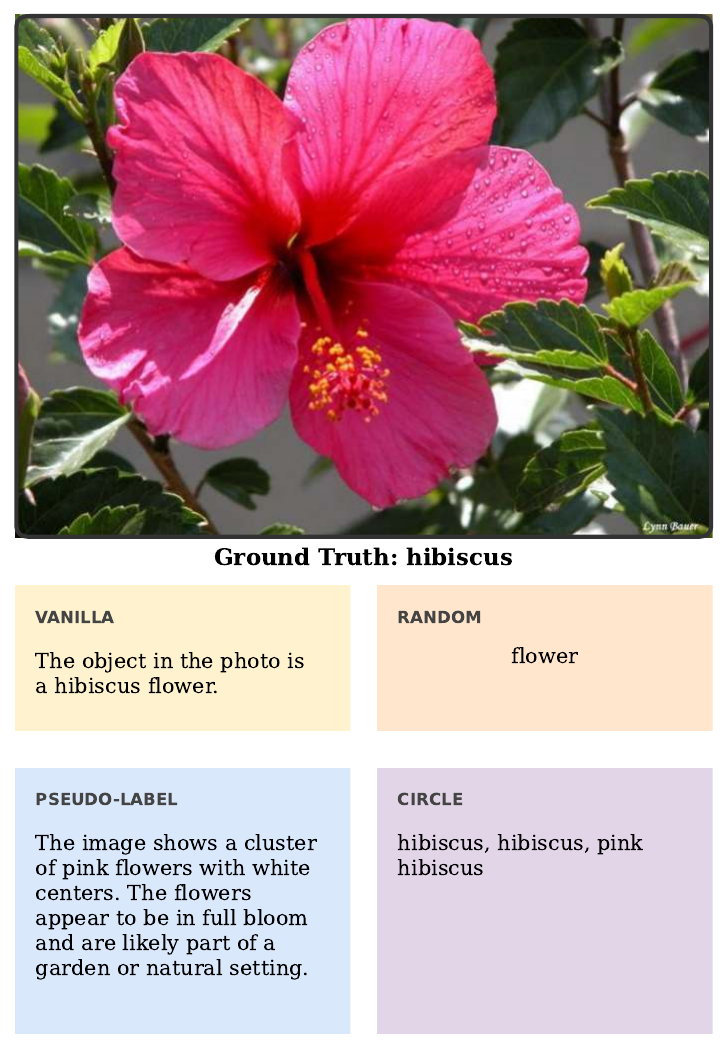}
        \label{supp:fig:flwr:a}
    \end{subfigure}
    \hfill
    \begin{subfigure}[b]{0.3\linewidth}
        \centering
        \includegraphics[width=\linewidth]{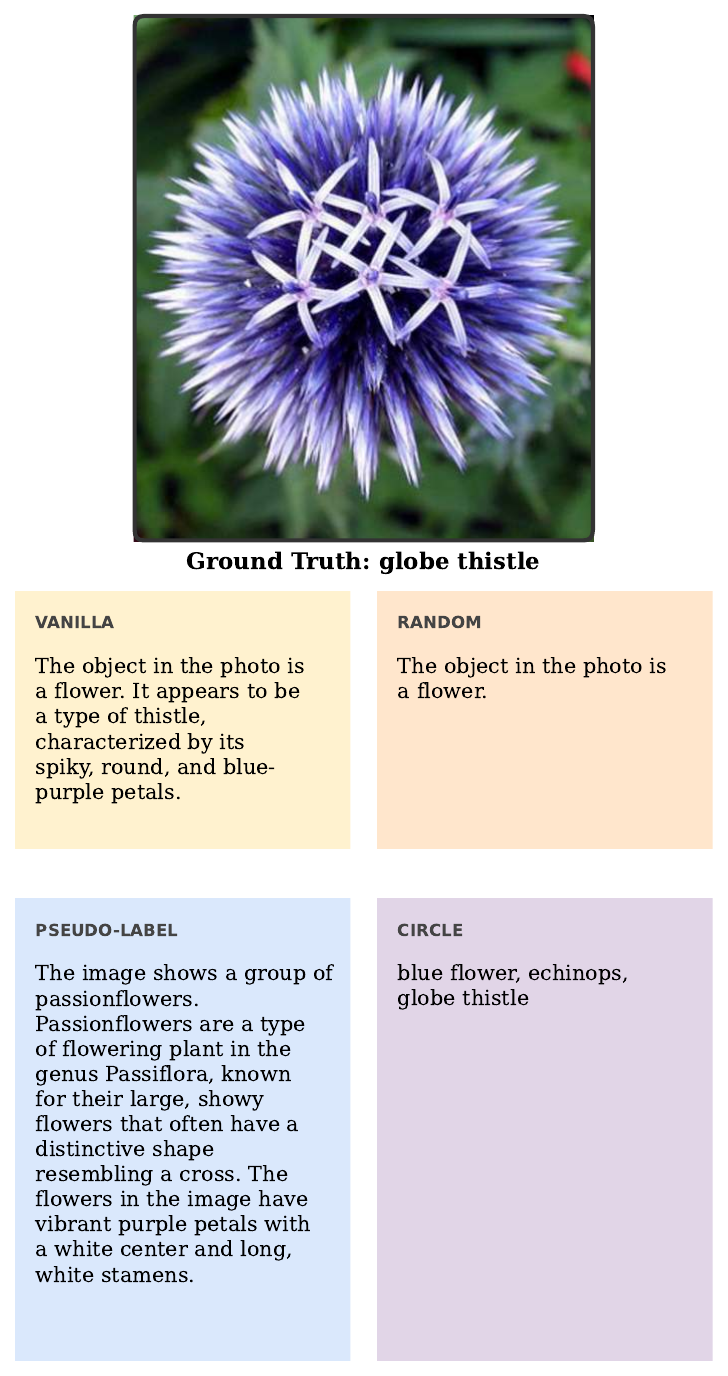}
        \label{supp:fig:flwr:b}
    \end{subfigure}
    \hfill
    \begin{subfigure}[b]{0.3\linewidth}
        \centering
        \includegraphics[width=\linewidth]{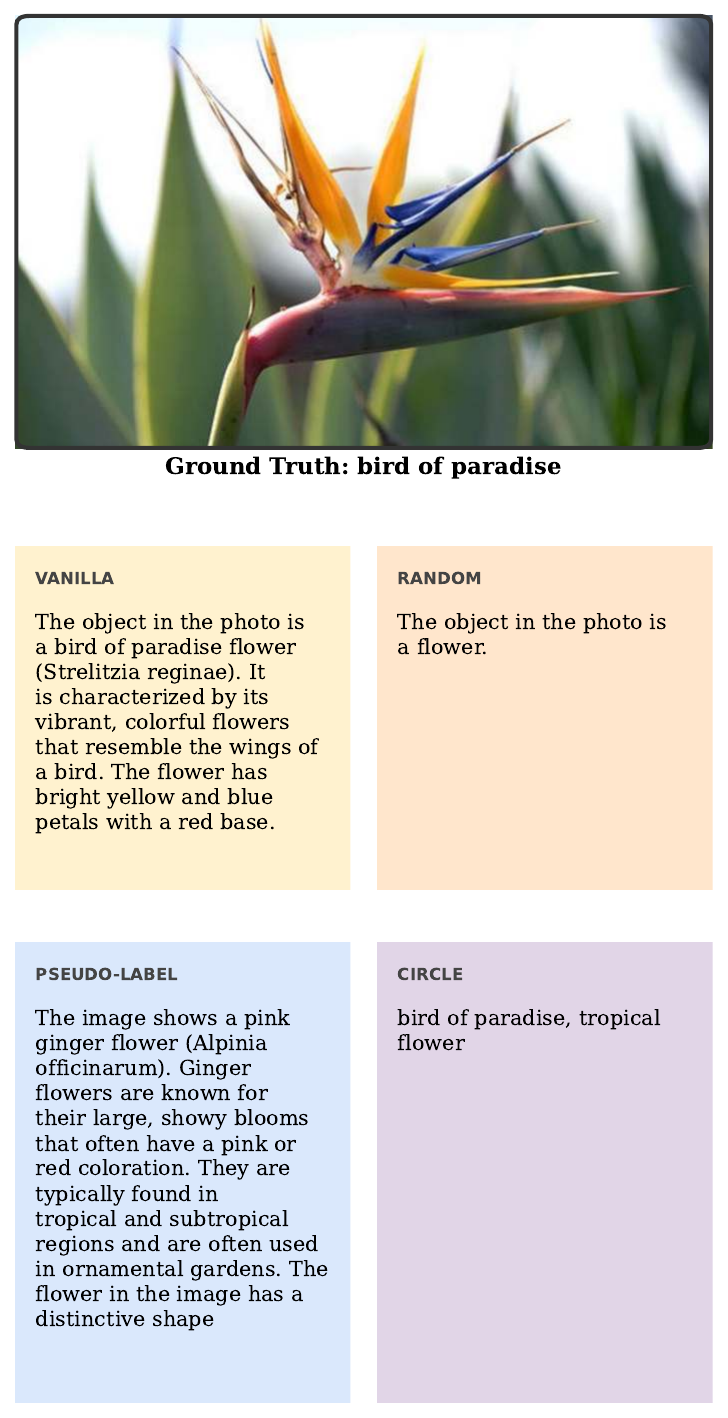}
        \label{supp:fig:flwr:c}
    \end{subfigure}
    
    \caption{
    \textbf{Qualitative results from Flowers102~\cite{nilsback2008flowers}.}
    We visualize three distinct samples, showing the response given by the \inlineColorbox{DrawioYellow}{\textit{Vanilla}} model, with \inlineColorbox{DrawioOrange}{\textit{Random}} context, with \inlineColorbox{DrawioBlue}{\textit{Pseudo-labeling}} examples, and using our \inlineColorbox{DrawioPurple}{\ours}.
    We use Qwen2-VL~\cite{wang2024qwen2} 7B as the LMM.
    }
    \label{supp:fig:flwr}
\end{figure*}

\begin{figure*}[p!]
    \centering
    \begin{subfigure}[b]{0.3\linewidth}
        \centering
        \includegraphics[width=\linewidth]{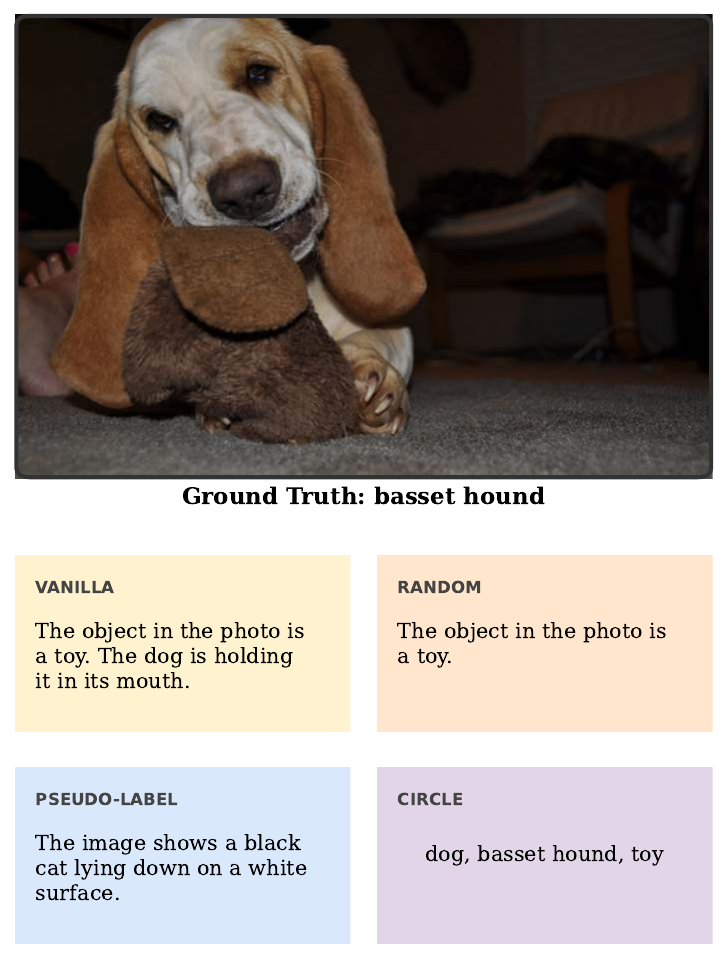}
        \label{supp:fig:pets:a}
    \end{subfigure}
    \hfill
    \begin{subfigure}[b]{0.3\linewidth}
        \centering
        \includegraphics[width=\linewidth]{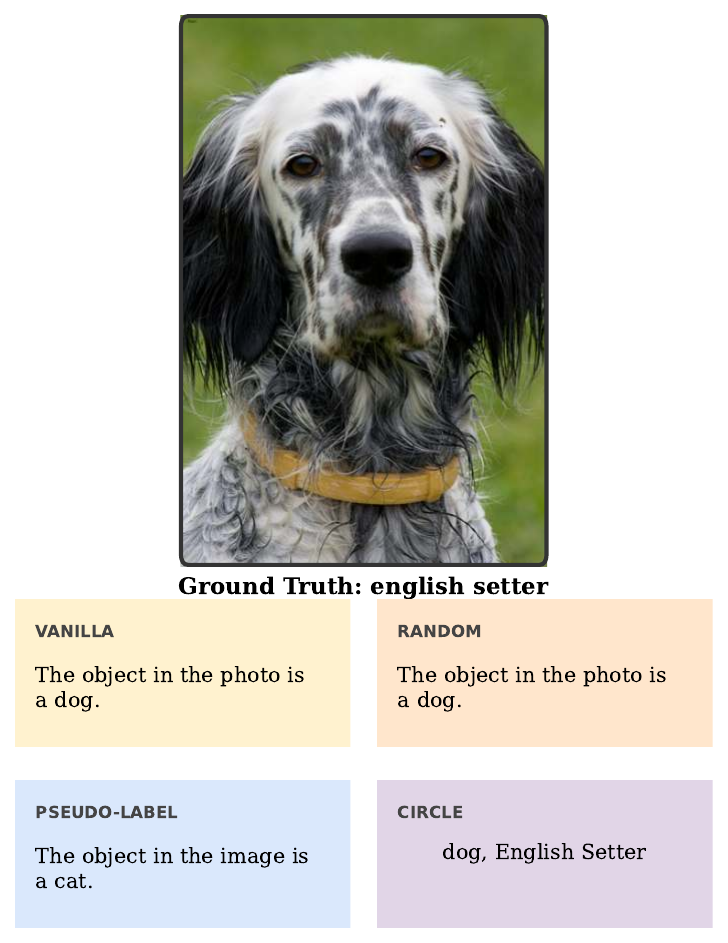}
        \label{supp:fig:pets:b}
    \end{subfigure}
    \hfill
    \begin{subfigure}[b]{0.3\linewidth}
        \centering
        \includegraphics[width=\linewidth]{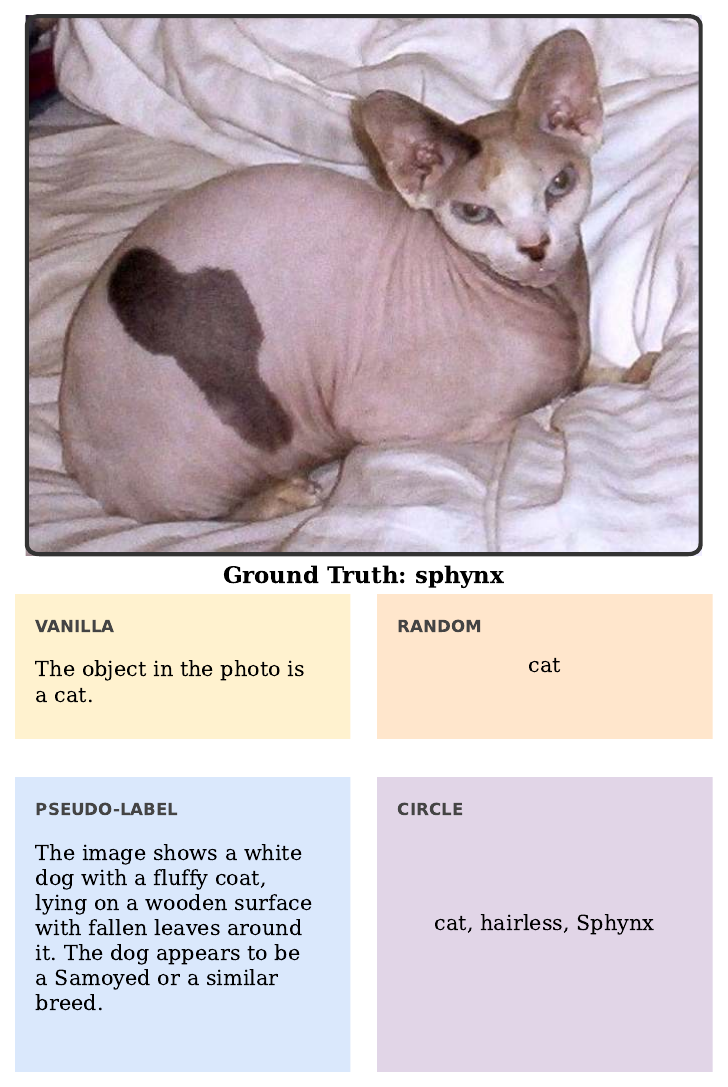}
        \label{supp:fig:pets:c}
    \end{subfigure}
    
    \caption{
    \textbf{Qualitative results from Oxford Pets~\cite{parkhi2012pets}.}
    We visualize three distinct samples, showing the response given by the \inlineColorbox{DrawioYellow}{\textit{Vanilla}} model, with \inlineColorbox{DrawioOrange}{\textit{Random}} context, with \inlineColorbox{DrawioBlue}{\textit{Pseudo-labeling}} examples, and using our \inlineColorbox{DrawioPurple}{\ours}.
    We use Qwen2-VL~\cite{wang2024qwen2} 7B as the LMM.
    }
    \label{supp:fig:pets}
\end{figure*}

\begin{figure*}[p!]
    \centering
    \begin{subfigure}[b]{0.3\linewidth}
        \centering
        \includegraphics[width=\linewidth]{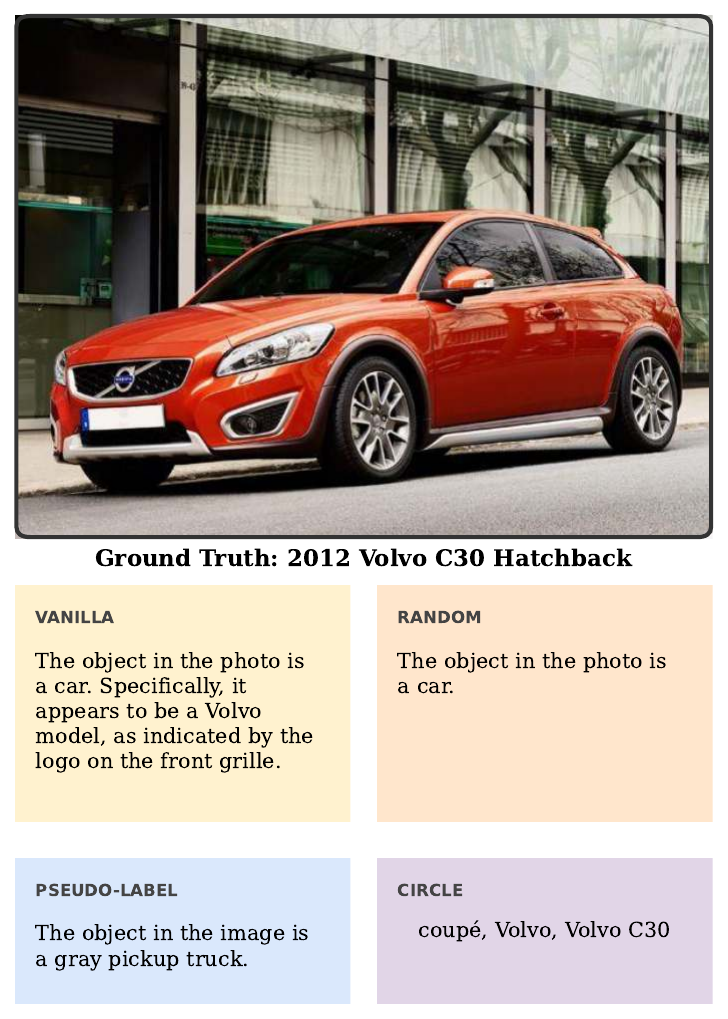}
        \label{supp:fig:cars:a}
    \end{subfigure}
    \hfill
    \begin{subfigure}[b]{0.3\linewidth}
        \centering
        \includegraphics[width=\linewidth]{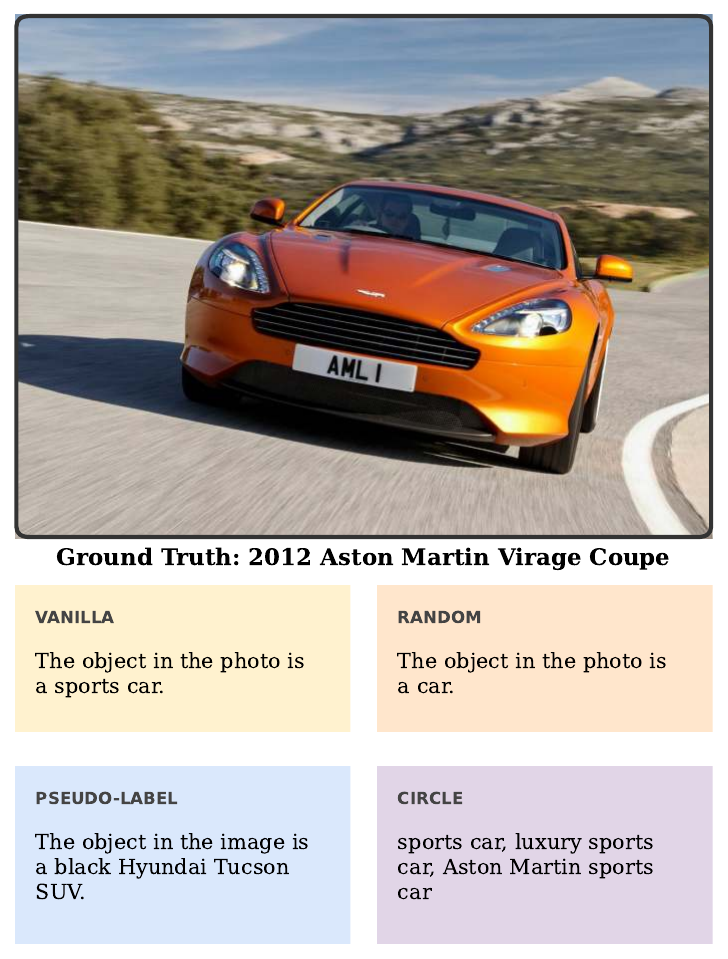}
        \label{supp:fig:cars:b}
    \end{subfigure}
    \hfill
    \begin{subfigure}[b]{0.3\linewidth}
        \centering
        \includegraphics[width=\linewidth]{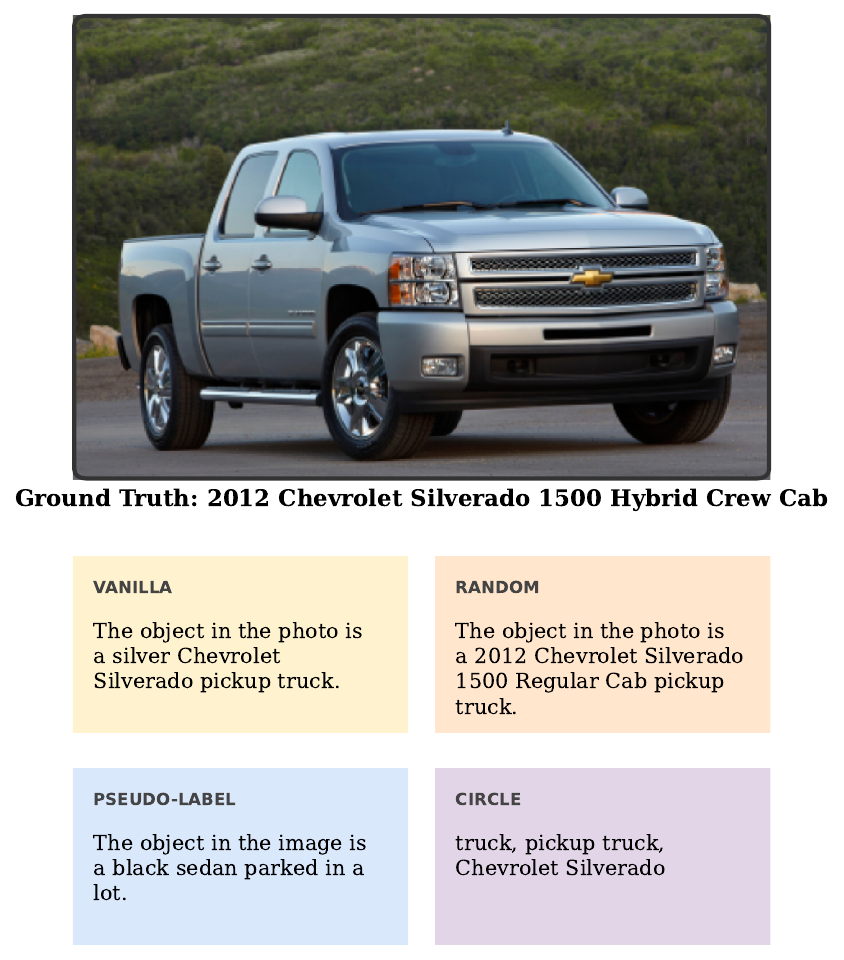}
        \label{supp:fig:cars:c}
    \end{subfigure}
    
    \caption{
    \textbf{Qualitative results from Stanford Cars~\cite{krause20133cars}.}
    We visualize three distinct samples, showing the response given by the \inlineColorbox{DrawioYellow}{\textit{Vanilla}} model, with \inlineColorbox{DrawioOrange}{\textit{Random}} context, with \inlineColorbox{DrawioBlue}{\textit{Pseudo-labeling}} examples, and using our \inlineColorbox{DrawioPurple}{\ours}.
    We use Qwen2-VL~\cite{wang2024qwen2} 7B as the LMM.
    }
    \label{supp:fig:cars}
\end{figure*}

\begin{figure*}[p!]
    \centering
    \begin{subfigure}[b]{0.3\linewidth}
        \centering
        \includegraphics[width=\linewidth]{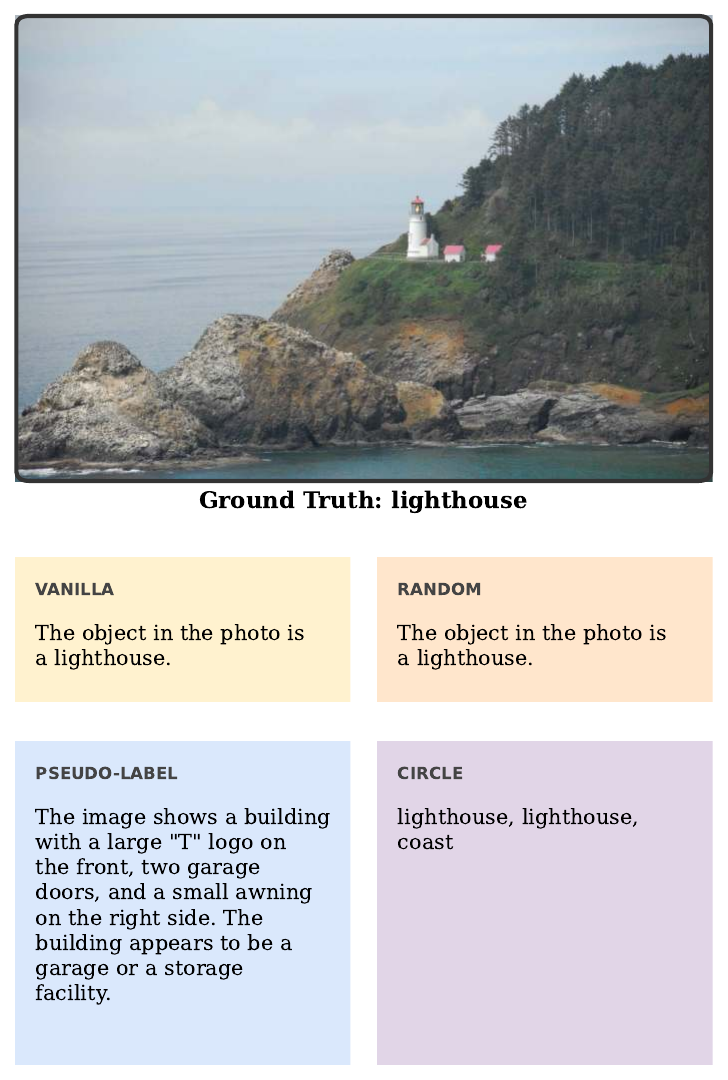}
        \label{supp:fig:sun:a}
    \end{subfigure}
    \hfill
    \begin{subfigure}[b]{0.3\linewidth}
        \centering
        \includegraphics[width=\linewidth]{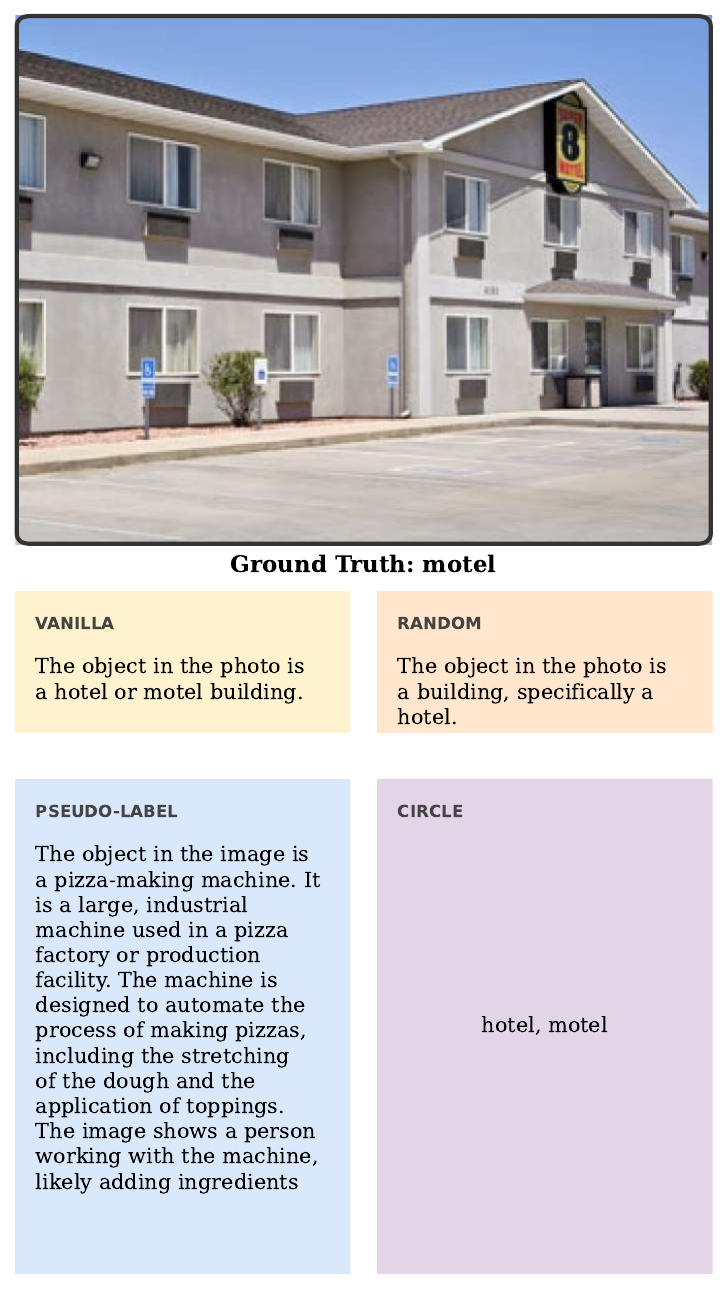}
        \label{supp:fig:sun:b}
    \end{subfigure}
    \hfill
    \begin{subfigure}[b]{0.3\linewidth}
        \centering
        \includegraphics[width=\linewidth]{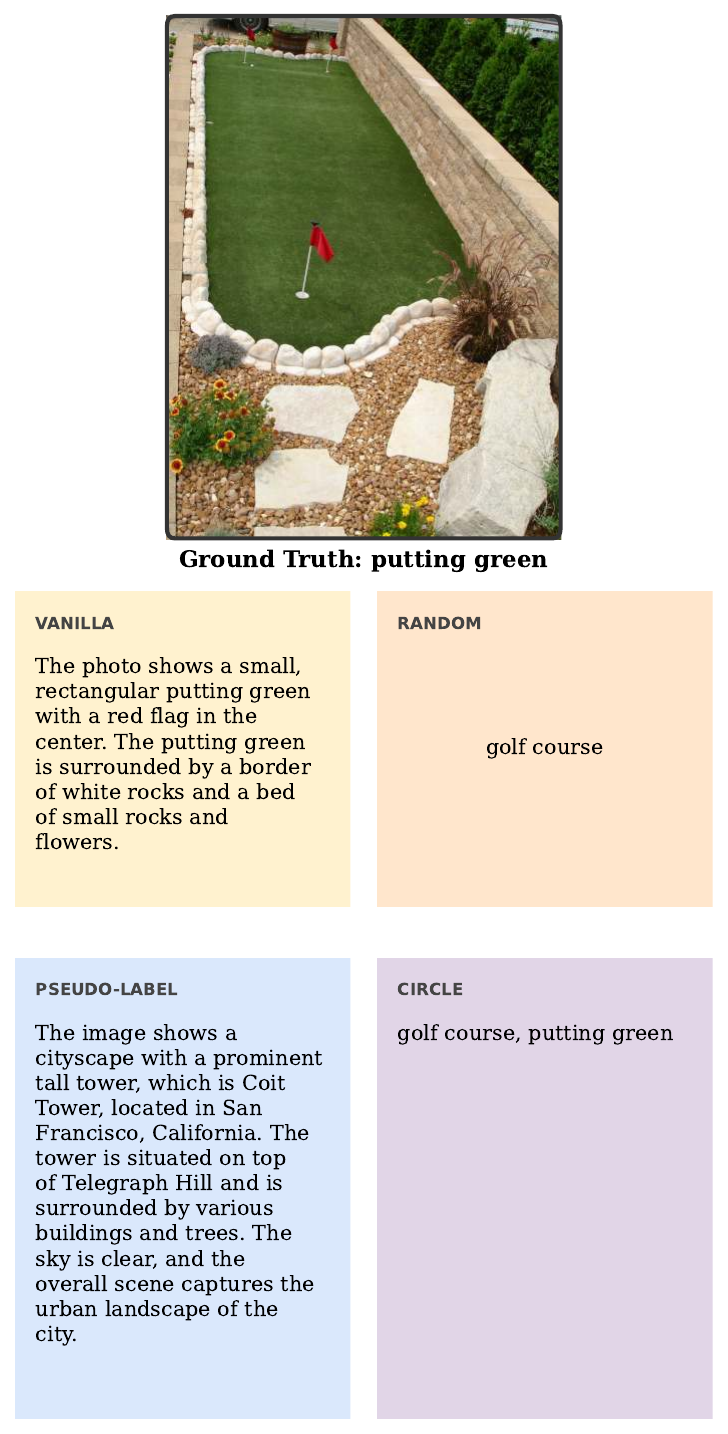}
        \label{supp:fig:sun:c}
    \end{subfigure}
    
    \caption{
    \textbf{Qualitative results from SUN397~\cite{xiao2010sun}}
    We visualize three distinct samples, showing the response given by the \inlineColorbox{DrawioYellow}{\textit{Vanilla}} model, with \inlineColorbox{DrawioOrange}{\textit{Random}} context, with \inlineColorbox{DrawioBlue}{\textit{Pseudo-labeling}} examples, and using our \inlineColorbox{DrawioPurple}{\ours}.
    We use Qwen2-VL~\cite{wang2024qwen2} 7B as the LMM.
    }
    \label{supp:fig:sun}
\end{figure*}

We provide additional qualitative examples to complement the visual analysis presented in \cref{fig:qualitative} (\textit{Main}).
\cref{supp:fig:caltech101,supp:fig:dtd,supp:fig:fgvc,supp:fig:flwr,supp:fig:pets,supp:fig:cars,supp:fig:sun} display a selection of samples across a subset of the ten evaluated datasets, providing three distinct examples for each to illustrate the model's behavior in diverse scenarios.

In these examples, we observe that while baseline methods exhibit inconsistent performance, occasionally identifying the correct concept but frequently misclassifying or hallucinating unrelated categories (especially in the \textit{Random} context), our \ours predicts the correct label more consistently.
This underscores the robustness of \ours, which effectively leverages the context it constructs during the iterative refinement steps to guide the underlying LMM to the correct specificity and format.

\clearpage

\end{document}